\documentclass[10pt,twocolumn,letterpaper]{article}

\usepackage[pagenumbers]{cvpr} 

\usepackage{graphicx}
\usepackage{amsmath}
\usepackage{amssymb}
\usepackage{booktabs}
\usepackage{amsmath}
\usepackage{algorithm}
\usepackage[noend]{algpseudocode}
\usepackage{multirow}
\usepackage{enumerate}
\usepackage{comment}
\usepackage{stackengine}
\usepackage{gensymb}
\usepackage{pifont}
\usepackage[accsupp]{axessibility}
\usepackage{tabularx}
\usepackage{multirow}
\newcommand*\samethanks[1][\value{footnote}]{\footnotemark[#1]}
\usepackage{subcaption}
\usepackage{xcolor} 

\usepackage{booktabs}
\usepackage{tabularx}
\usepackage[pagebackref,breaklinks,colorlinks,citecolor=blue]{hyperref}

\usepackage[capitalize]{cleveref}
\crefname{section}{Sec.}{Secs.}
\Crefname{section}{Section}{Sections}
\Crefname{table}{Table}{Tables}
\crefname{table}{Tab.}{Tabs.}

\def\cvprPaperID{6218} 
\def\confName{CVPR}
\def\confYear{2023}

\begin{document}

\title{ACL-SPC: Adaptive Closed-Loop system \textit{for} \\ Self-Supervised Point Cloud Completion}

\author{Sangmin Hong$^{1}$\thanks{equal contribution} \qquad Mohsen Yavartanoo$^{2}$\samethanks \qquad Reyhaneh Neshatavar$^{2}$ \qquad Kyoung Mu Lee$^{1,2}$ \\$^{1}$IPAI,  $^{2}$Dept. of ECE \& ASRI, Seoul National University, Seoul, Korea\\
{\tt\small \{mchiash2,myavartanoo,reyhanehneshat,kyoungmu\}@snu.ac.kr}}

\maketitle

\begin{abstract}
  Point cloud completion addresses filling in the missing parts of a partial point cloud obtained from depth sensors and generating a complete point cloud.
  Although there has been steep progress in the supervised methods on the synthetic point cloud completion task, it is hardly applicable in real-world scenarios due to the domain gap between the synthetic and real-world datasets or the requirement of prior information.
  To overcome these limitations, we propose a novel self-supervised framework ACL-SPC for point cloud completion to train and test on the same data.
  ACL-SPC takes a single partial input and attempts to output the complete point cloud using an adaptive closed-loop~(ACL) system that enforces the output same for the variation of an input.
  We evaluate our ACL-SPC on various datasets to prove that it can successfully learn to complete a partial point cloud as the first self-supervised scheme.
  Results show that our method is comparable with unsupervised methods and achieves superior performance on the real-world dataset compared to the supervised methods trained on the synthetic dataset.
  Extensive experiments justify the necessity of self-supervised learning and the effectiveness of our proposed method for the real-world point cloud completion task.  
  The code is publicly available from this \href{https://github.com/Sangminhong/ACL-SPC_PyTorch}{link}.
 \end{abstract}

\section{Introduction}
\label{sec:Introduction}
\begin{figure}[ht]
\centering
    \includegraphics[trim={0 0 0 0cm},clip, width=\linewidth]{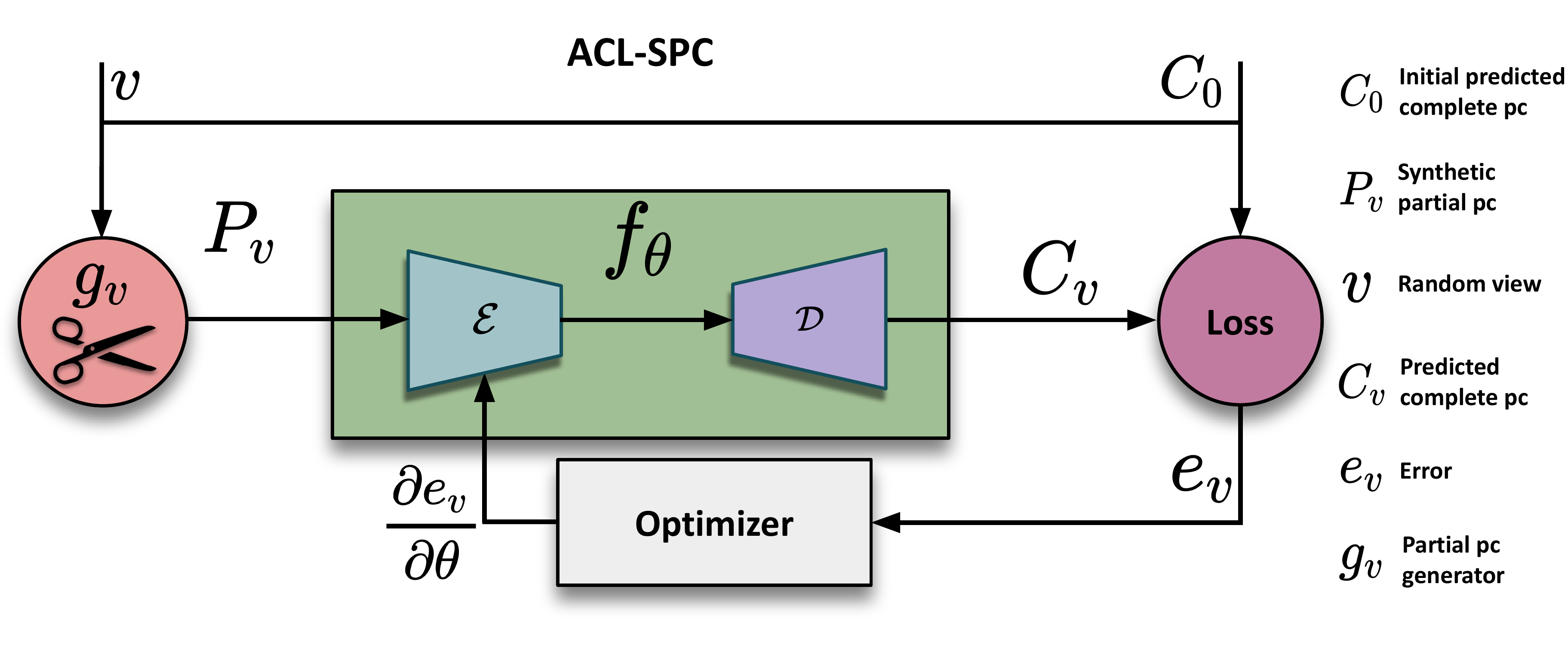}
    \caption{
        \textbf{Overview of our proposed pipeline.}
        We first generate $C_0$ using the initial partial point cloud. 
        Then, multiple synthetic point clouds $P_v$ are generated from the random views of $C_0$.
        We input the generated $P_v$ to the network and make predicted complete point clouds.
        We take the loss between $C_0$ and $C_v$ to optimize the parameters of the network $f_{\theta}$.
    }
    \label{fig:pipeline}
    \vspace{-5mm}
\end{figure}
Along with the development of autonomous driving cars and robotics, the usage of depth sensors such as LiDARs has increased.
These sensors can collect numerous points in the 3D space, and the combination of these points forms a 3D representation called a point cloud.
Point cloud representation has been widely used in many applications as it is highly convertible to other 3D data representations, \eg, voxel and mesh, and accessible for obtaining information from the real world.
However, point clouds obtained from a real-world sensor, \eg, a LiDAR, are often incomplete and sparse due to occlusion, limitations of sensor resolution, and viewing angle~\cite{pcn} leading to loss of some geometric information and difficulty in proceeding with further applications \eg, object detection~\cite{pang2019libra} and object segmentation~\cite{chen2021moving}.
We define such point clouds as partial point clouds.
Therefore, point cloud completion is a crucial task that infers completing geometric 3D shapes by using such partial point cloud observations.
%
%

With the advent of deep learning, previous data-driven works~\cite{pcn, wang2020cascaded, GRNET} have been able to solve this task using complete point cloud ground-truths. 
%
Even though such methods have achieved decent performance, they are not applicable in real-world scenarios where the ground-truth point clouds are not easy to obtain.  
For these reasons, researchers have recently attempted to overcome the lack of high-quality and large-scale paired training data using multiple views of the point cloud in unsupervised and weakly-supervised manners.
Especially, recent methods~\cite{WeakPCN, PointPnCNet} leverage multi-view consistency of the desired object, which shows effectiveness in supervising 3D shape prediction.
PointPnCNet~\cite{PointPnCNet} claims that its method is based on self-supervised learning. 
However, combining multi-view consistency enables reconstructing a complete 3D point cloud and can be weak supervision.
Moreover, collecting multiple partial views of an object in real-world scenarios is difficult as gathering ground-truth point clouds.
Therefore, the necessity for multi-view consistency prevents this method from being fully self-supervised.
Meanwhile, other methods~\cite{pcl2pcl, ShapeInversion, Cycle4Completion, Optde} exploit unpaired partial and complete point clouds~\cite{pcl2pcl, Cycle4Completion} or pre-trained models~\cite{ShapeInversion, Optde} on synthetic data to overcome the difficulty of collecting ground-truth.
However, the need for unpaired data limits the methods' applicability to a few categories. 

To overcome the challenges mentioned above, we propose a novel and the first self-supervised method called ACL-SPC for point cloud completion using only a single partial point cloud.
We develop an adaptive closed-loop~(ACL)~\cite{aastrom2013adaptive} system as shown in Figure~\ref{fig:pipeline} to design our self-supervised point cloud completion framework ACL-SPC. 
In ACL-SPC, an encoder adaptively reacts to the variance in the input by adjusting its parameters to generate the same output.
Using our developed ACL, our method tries to generate a complete point cloud from a single partial input captured from an unknown viewpoint without any prior information or multi-view consistency and also simulates several synthetic partial point clouds from the reconstructed point cloud.
Under our defined novel loss function, our ACL-SPC can learn to generate the same complete point cloud from all such synthetic point clouds and the initial partial point cloud without any supervision.
In the experiments, we demonstrate the ability of our method to restore a complete point cloud and the effect of our designed loss functions on saving fine details and improving quantitative performance.
We also evaluate our method with various datasets, including real-world scenarios, and verify that our method can be applied in practice.
Evaluation results show that our method is comparable to other unsupervised methods and performs better than the supervised method trained on a synthetic dataset.

Our main contributions can be summarized as follows: 
\begin{itemize}
\item We propose ACL-SPC by developing an adaptive control-loop ACL framework to solve the point cloud completion problem in a self-supervised manner.
\item We also design an effective self-supervised loss function to train our method without requiring any other information and using only a single partial point cloud taken from an unknown viewpoint.
\item Our method achieves superior performance in real-world scenarios compared to methods trained on synthetic datasets and comparative performance among other unsupervised methods.
\end{itemize}
\section{Related Works}
\label{related_works}
\subsection{Supervised point cloud completion}
Point cloud completion is the task of reconstructing a complete geometry of a shape from partial point clouds.
Before the advancement of the deep neural network, some traditional geometric-based methods~\cite{Nealen_2006, Sarkar_2017, Dai_2017_CVPR} have been attempted to complete shapes using the geometric priors from a partial input without any external data.
%
%
Other methods~\cite{Mitra_ACM_2006, Pauly_ACM_2008, Podolak_ACM_2006, Sung_ACM_2015, Thrun_CVPR_2005} have been proposed to handle the point cloud completion task by utilizing the symmetry property of the object to complete the incomplete parts. 

With the development of deep learning, some learning-based methods~\cite{Dai_2017_CVPR, Nguyen_2016_CVPR, sharma2016vconv} have shown up to solve complete the partial point cloud using a large amount of data.
However, these methods have converted a partial point cloud into voxels to apply convolutional neural networks~(CNNs), which leads to computational complexity and losing some geometrical information of point clouds.
PCN~\cite{pcn} as the first data-driven approach learned a completion network directly from point clouds rather than converting to other representations. 
Further, various works have proposed developed architectures using a novel rooted tree structure~\cite{TopNet}, 3D grids as the intermediate representation~\cite{GRNET}, feedback refinement module~\cite{FBNet}, and transformers~\cite{PointTr, seedformer} to improve the performance.
However, these methods deviate from real-world scenarios, as gathering ground truth point clouds is cost-inefficient and not practical.
%
\begin{figure*}[t]
\centering
    \includegraphics[width=\linewidth, page=1]{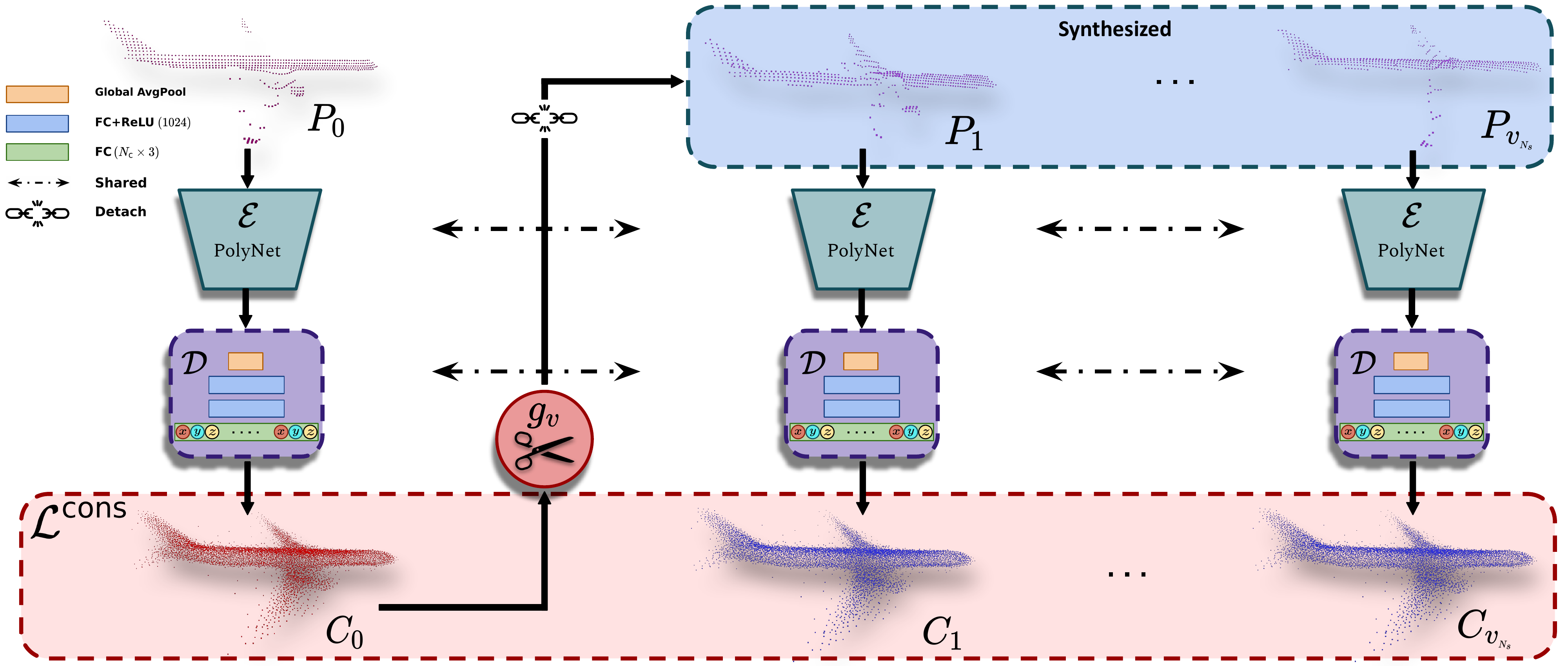}
    \caption{
        \textbf{The framework of ACL-SPC.}
        Our framework consists of an encoder-decoder style network where the parameters are shared between the objects.
        The network adopts the PolyNet~\cite{polynet} as the encoder and three fully connected~(FC) layers as the decoder.
        Our network first takes the input partial point cloud and generates an estimated complete point cloud.
        Using this point cloud, we make multiple synthesized partial point clouds as new inputs.
        Again, the network outputs estimated complete point clouds from the synthesized partial point clouds.
        We apply consistency loss between multiple estimated complete point clouds and optimize the parameters of the network.
    }
    \label{fig:framework}
    \vspace{-4mm}
\end{figure*}
\subsection{Unsupervised point cloud completion}
Due to the aforementioned limitations of supervised methods for point cloud completion, existing unsupervised approaches~\cite{WeakPCN, PointPnCNet, DPC} have been proposed to handle point cloud completion tasks where ground truth data are unavailable. 
Meanwhile, weakly supervised methods~\cite{WeakPCN, DPC} have attempted to predict the complete point cloud using multiple partial views, which are not always available in real-world scenarios.
%
%
Later, PointPnCNet~\cite{PointPnCNet} introduced an inpainting framework with geometric consistency to overcome the above issue and claim that the method is the first self-supervised work for this task. 
Nevertheless, this method has exploited geometric consistency between multi-views of an object and showed that without this supervision, they could not complete the partial point cloud.
%
%
Moreover, there have been attempts to utilize unpaired complete point clouds from synthetic datasets to solve the difficulty of accessing ground-truths~\cite{pcl2pcl, ShapeInversion, Cycle4Completion, Optde}. 
As a domain gap exists between unpaired complete point clouds and partial point clouds, these methods design architectures that can transform from one domain to the other and eventually solve the issue.
%
However, these methods can only be suitable for categories available in synthetic datasets.
\subsection{Self-supervised Learning}
Self-supervised learning has attracted increasing attention in computer vision due to its practicality and ability to avoid the need for expensive annotated datasets.
Following the advances in CNNs, recent self-supervised learning methods have incorporated generative~\cite{goodfellow2014generative, zhu2017unpaired, karras2019style, van2016pixel, reed2016generative, kim2017learning} and contrastive approaches~\cite{xie2020pointcontrast, afham2022crosspoint, du2021self}, to learn the features from unlabeled data where an input itself provides supervision.
%
%
Furthermore, researchers~\cite{NEURIPS2019_wallach, NEURIPS2020_Sharma, Eckart_2021_CVPR, https://doi.org/10.48550/arxiv.2204.08196, arXiv_2022_Zhou, arXiv_2022_Yan, PointMAE, MaskPoint} have started to apply self-supervised methods on point clouds to overcome the cumbersome task of annotating. 
These works have successfully shown great performance on feature learning to handle tasks such as classification~\cite{NEURIPS2019_wallach, NEURIPS2020_Sharma, Eckart_2021_CVPR, arXiv_2022_Zhou, arXiv_2022_Yan, PointMAE, MaskPoint}, segmentation~\cite{NEURIPS2019_wallach, NEURIPS2020_Sharma, Eckart_2021_CVPR, arXiv_2022_Zhou, arXiv_2022_Yan, PointMAE, MaskPoint} or upsampling~\cite{https://doi.org/10.48550/arxiv.2204.08196}. 
In this way, we propose the first self-supervised method for point cloud completion using only a partial point cloud as input without any prior information.
\section{Method}
\label{sec:Method}
In control theory~\cite{Nise, ogata2010modern}, closed-loop systems have many applications in various areas such as aerospace, electronics, and biomedical.
Especially adaptive closed-loop~(ACL) is a system where a controller automatically gives a compensated signal for the variation in the system so that the overall result remains the same~\cite{aastrom2013adaptive, Nise, ogata2010modern}.
In an ACL system, a controller outputs an appropriate signal after receiving feedback from the error between the desired output and the generated one.
Meanwhile, obtaining a complete point cloud generator invariant to the view of captured partial point clouds is essential for point cloud completion tasks.
We believe that the aforementioned attribute of the ACL system can be utilized for this task because it is appropriate for constructing the same complete point cloud, whatever a partial point cloud of an object comes in as an input. 
Therefore, we develop the concept of ACL for point cloud completion and introduce a novel self-supervised partial point-cloud completion framework~(ACL-SPC).
\begin{table*}[t]
    \small
    \centering
    \setlength\tabcolsep{0.005pt}
    \begin{tabularx}{\linewidth}{l 
    >{\centering\arraybackslash}X 
    >{\centering\arraybackslash}X 
    >{\centering\arraybackslash}X  
    >{\centering\arraybackslash}X
    >{\centering\arraybackslash}X
    >{\centering\arraybackslash}X
    >{\centering\arraybackslash}X
    >{\centering\arraybackslash}X
    >{\centering\arraybackslash}X
    >{\centering\arraybackslash}X  
    >{\centering\arraybackslash}X
    >{\centering\arraybackslash}X  
    >{\centering\arraybackslash}X
    >{\centering\arraybackslash}X   
    >{\centering\arraybackslash}X  
    >{\centering\arraybackslash}X
    >{\centering\arraybackslash}X   
    >{\centering\arraybackslash}X 
    >{\centering\arraybackslash}X 
    }

    \toprule
    \multirow{2}{*}{\textbf{Supervision}}&
    \multicolumn{2}{c}{\multirow{2}{*}{\textbf{Method}}}&&
    \multicolumn{3}{c}{\textbf{Airplane}} &&
    \multicolumn{3}{c}{\textbf{Car}}&&
    \multicolumn{3}{c}{\textbf{Chair}}&&
    \multicolumn{3}{c}{\textbf{Average}} &
    \\
    && && 
    P$\downarrow$& C$\downarrow$ & CD$\downarrow$ && P$\downarrow$ & C$\downarrow$ & CD$\downarrow$ && P$\downarrow$ & C$\downarrow$ & CD$\downarrow$ && P$\downarrow$ & C$\downarrow$ & CD$\downarrow$ &
    \\
    \midrule
    & \multicolumn{2}{c}{DPC~\cite{DPC}}& 
    & {-} & {-} & {3.91} &  
    & {-} & {-} & {3.47} &
    & {-} & {-} & {4.30} &
    & {-} & {-} & {3.89} &\\
    Unsupervised&\multicolumn{2}{c}{Gu~\etal~\cite{WeakPCN}}& 
    & {0.91} & {1.05} & \textbf{1.95} & 
    & {1.27} & {1.41} & \textbf{2.68} &
    & {1.69} & {1.64} & \textbf{3.33} &
    & {1.29} & {1.36} & \textbf{2.65} &\\
    &\multicolumn{2}{c}{PointPnCNet~\cite{PointPnCNet}}& 
    & {1.58} & {1.74} & {3.32} & 
    & {1.98} & {2.98} & {4.96} &
    & {2.72} & {2.68} & {5.40} &
    & {1.75} & {2.46} & {4.56} &\\
    \midrule
    Self-supervised &\multicolumn{2}{c}{\textbf{Ours}}& 
    & {1.20} & \textbf{0.80} & {2.01} & 
    & {1.65} & \textbf{1.28} & {2.93} &
    & {2.25} & \textbf{1.46} & {3.71} &
    & {1.70} & \textbf{1.18} & {2.88} &\\
    \bottomrule
    \end{tabularx}

    \caption{
        \textbf{Quantitative results on three categories airplane, car and chair.} 
        We also calculate average values among the categories.
        P, C, and CD refers to precision, coverage, and Chamfer distance, respectively.
        All the values are multiplied by $100$.
        %
        }
    \label{tab:shapenet}
    \vspace{-4mm}
\end{table*}

\begin{figure*}
     \centering
     \begin{subfigure}[b]{0.14\textwidth}
         \centering
         \includegraphics[width=\textwidth, page=1]{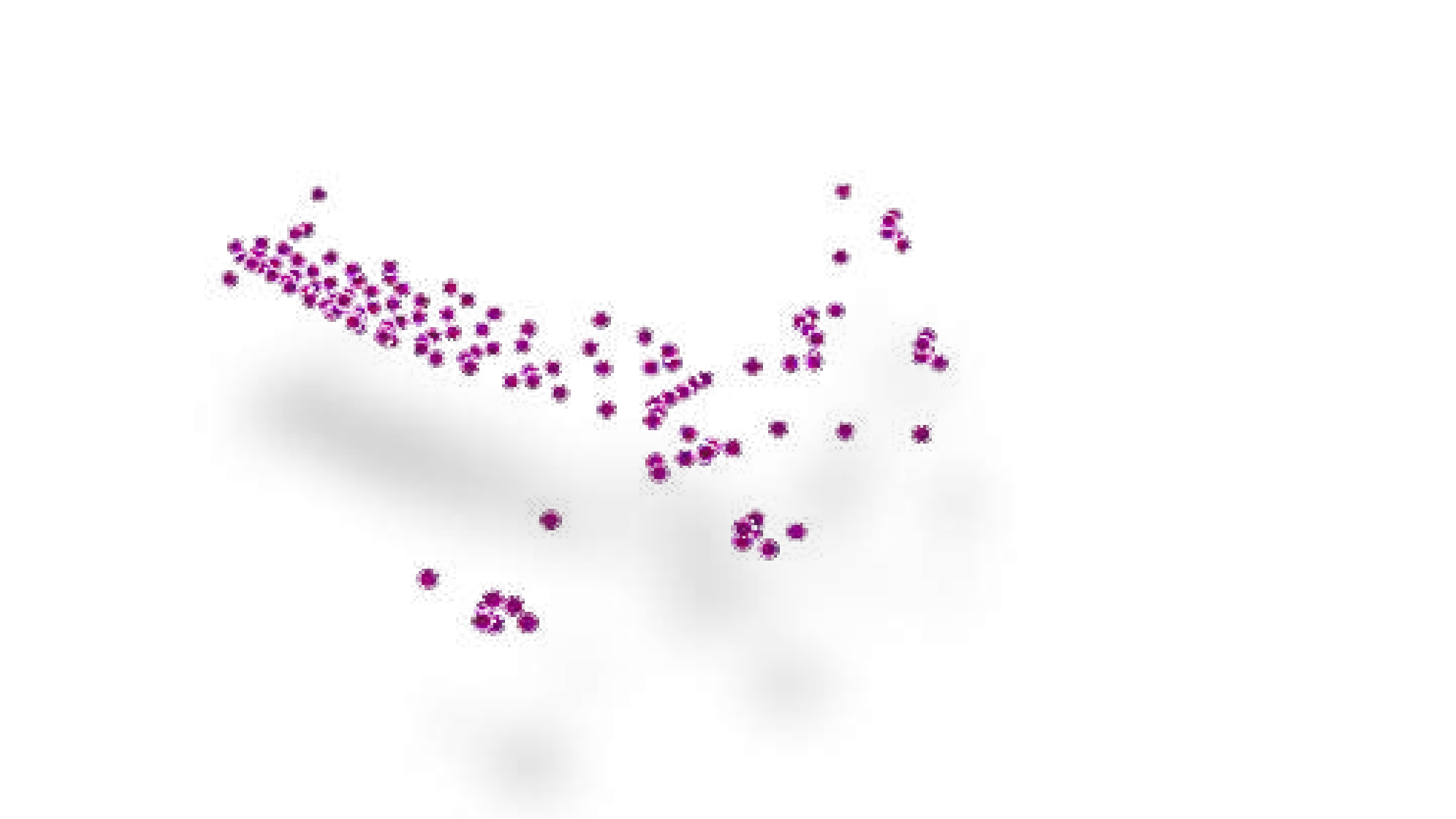}
         \label{fig:loss_a}
     \end{subfigure}
     \hfill
     \begin{subfigure}[b]{0.14\textwidth}
         \centering
         \includegraphics[width=\textwidth, page=2]{figures/qualitative_results_compare.pdf}
         \label{fig:loss_b}
     \end{subfigure}
     \hfill
     \begin{subfigure}[b]{0.14\textwidth}
         \centering
         \includegraphics[width=\textwidth, page=3]{figures/qualitative_results_compare.pdf}
         \label{fig:loss_b}
     \end{subfigure}
     \hfill
     \begin{subfigure}[b]{0.14\textwidth}
         \centering
         \includegraphics[width=\textwidth, page=4]{figures/qualitative_results_compare.pdf}
         \label{fig:loss_b}
     \end{subfigure}
     \hfill
     \begin{subfigure}[b]{0.14\textwidth}
         \centering
         \includegraphics[width=\textwidth, page=5]{figures/qualitative_results_compare.pdf}
         \label{fig:loss_a}
     \end{subfigure}
     \hfill
     \begin{subfigure}[b]{0.14\textwidth}
         \centering
         \includegraphics[width=\textwidth, page=6]{figures/qualitative_results_compare.pdf}
         \label{fig:loss_b}
     \end{subfigure}
     \hfill
     \begin{subfigure}[b]{0.14\textwidth}
         \centering
         \includegraphics[width=\textwidth, page=7]{figures/qualitative_results_compare.pdf}
         \label{fig:loss_b}
     \end{subfigure}
     \hfill
     \begin{subfigure}[b]{0.14\textwidth}
         \centering
         \includegraphics[width=\textwidth, page=8]{figures/qualitative_results_compare.pdf}
         \label{fig:loss_b}
     \end{subfigure}
     \hfill   
     \begin{subfigure}[b]{0.14\textwidth}
         \centering
         \includegraphics[width=\textwidth, page=9]{figures/qualitative_results_compare.pdf}
         \label{fig:loss_b}
     \end{subfigure}
     \hfill
     \begin{subfigure}[b]{0.14\textwidth}
         \centering
         \includegraphics[width=\textwidth, page=10]{figures/qualitative_results_compare.pdf}
         \label{fig:loss_b}
     \end{subfigure}
     \hfill 
     \begin{subfigure}[b]{0.14\textwidth}
         \centering
         \includegraphics[width=\textwidth, page=11]{figures/qualitative_results_compare.pdf}
         \label{fig:loss_b}
     \end{subfigure}
     \hfill
     \begin{subfigure}[b]{0.14\textwidth}
         \centering
         \includegraphics[width=\textwidth, page=12]{figures/qualitative_results_compare.pdf}
         \label{fig:loss_b}
     \end{subfigure}
     \hfill     
     \begin{subfigure}[b]{0.14\textwidth}
         \centering
         \includegraphics[width=\textwidth, page=13]{figures/qualitative_results_compare.pdf}
         \caption{Input}
         \label{subfig:input}
     \end{subfigure}
     \hfill
     \begin{subfigure}[b]{0.14\textwidth}
         \centering
         \includegraphics[width=\textwidth, page=14]{figures/qualitative_results_compare.pdf}
         \caption{GT}
         \label{subfig:GT}
     \end{subfigure}
     \hfill
     \begin{subfigure}[b]{0.14\textwidth}
         \centering
         \includegraphics[width=\textwidth, page=15]{figures/qualitative_results_compare.pdf}
         \caption{Multi-View}
         \label{subfig:multiview}
     \end{subfigure}
     \hfill
     \begin{subfigure}[b]{0.14\textwidth}
         \centering
         \includegraphics[width=\textwidth, page=16]{figures/qualitative_results_compare.pdf}
         \caption{GRNet\cite{GRNET}}
         \label{subfig:GRNet}
     \end{subfigure}
     \hfill
     \begin{subfigure}[b]{0.14\textwidth}
         \centering
         \includegraphics[width=\textwidth, page=17]{figures/qualitative_results_compare.pdf}
         \caption{Gu~\etal\cite{WeakPCN}}
         \label{subfig:Gu}
         
     \end{subfigure}   
     \hfill
     \begin{subfigure}[b]{0.14\textwidth}
         \centering
         \includegraphics[width=\textwidth, page=18]{figures/qualitative_results_compare.pdf}
         \caption{\textbf{Ours}}
         \label{subfig:ours}
     \end{subfigure}
     \hfill

        \caption{
        \textbf{Qualitative comparison on the ShapeNet dataset.} 
        We visualize a) the partial input, b) completed ground truth point cloud, c) multi-view point cloud, results on d) GRNet, e) Gu~\etal, and f) ours. 
        The multi-view point cloud is the concatenation of five random partial views of an object.
        Our result show that our method can recover most of the missing parts from the partial input. 
        }
        \label{fig:qual_figure}
        \vspace{-3mm}
\end{figure*}     

\begin{figure*}[t]
	\captionsetup[]{labelformat=empty}
		\newcommand{\rowArg}{1.53cm}
		\newcommand{\fullSize}{4.35cm}
		\newcommand{\patchSize}{1.23cm}
		\setlength\tabcolsep{0.05cm}
    \subfloat[Quantitative results.]{
    \footnotesize
    \begin{tabularx}{0.5\linewidth}{l 
    >{\centering\arraybackslash}X 
    >{\centering\arraybackslash}X 
    >{\centering\arraybackslash}X 
    >{\centering\arraybackslash}X
    >{\centering\arraybackslash}X }
    \toprule
    \textbf{Supervision}&
    \multicolumn{2}{l}{\textbf{Method}}& 
    \textbf{P}  & 
    \textbf{C} &
    \textbf{CD} \\
    \midrule
    \multirow{3}{*}{Supervised} 
    & \multicolumn{2}{l}{GRNet~\cite{GRNET}} & \textbf{4.63} & {6.90} & \textbf{11.53} \\
    & \multicolumn{2}{l}{SFNet~\cite{SnowflakeNet}} &  {14.12} & {12.64} & {26.76} \\
    & \multicolumn{2}{l}{pcn~\cite{pcn}} & {9.83} & {17.96} & {27.79} \\
    \midrule
    \multirow{2}{*}{Unsupervised} & \multicolumn{2}{l}{Gu~\cite{WeakPCN}} & {8.70} & {10.70} & {19.40}  \\
    & \multicolumn{2}{l}{PointPnCNet~\cite{PointPnCNet}} & {9.00} & {10.00} & {19.00} \\ 
    \midrule
    Self-supervised& \multicolumn{2}{l}{\textbf{Ours}} & {11.67} & \textbf{5.63} & {17.30} \\
    \bottomrule
    \label{fig:QuantKITTI}
    \end{tabularx}}
    \subfloat[Qualitative results. \label{tab:QualiKITTI}]{
        \footnotesize
		\begin{tabular}[b]{c c c}
				\includegraphics[width = .075\textwidth, page=15]
				{figures/KITTI_Result2.pdf}
				\includegraphics[width = .075\textwidth, page=16]
				{figures/KITTI_Result2.pdf} 
				\includegraphics[width =.075\textwidth, page=18]
				{figures/KITTI_Result2.pdf} 
				\includegraphics[width =.075\textwidth, page=19]
				{figures/KITTI_Result2.pdf} 				
				\includegraphics[width = .075\textwidth, page=21]
				{figures/KITTI_Result2.pdf}	
				\includegraphics[width = .075\textwidth, page=17]
				{figures/KITTI_Result2.pdf}
				\vspace{1mm}
				\\
				\includegraphics[width = .075\textwidth, page=1]
				{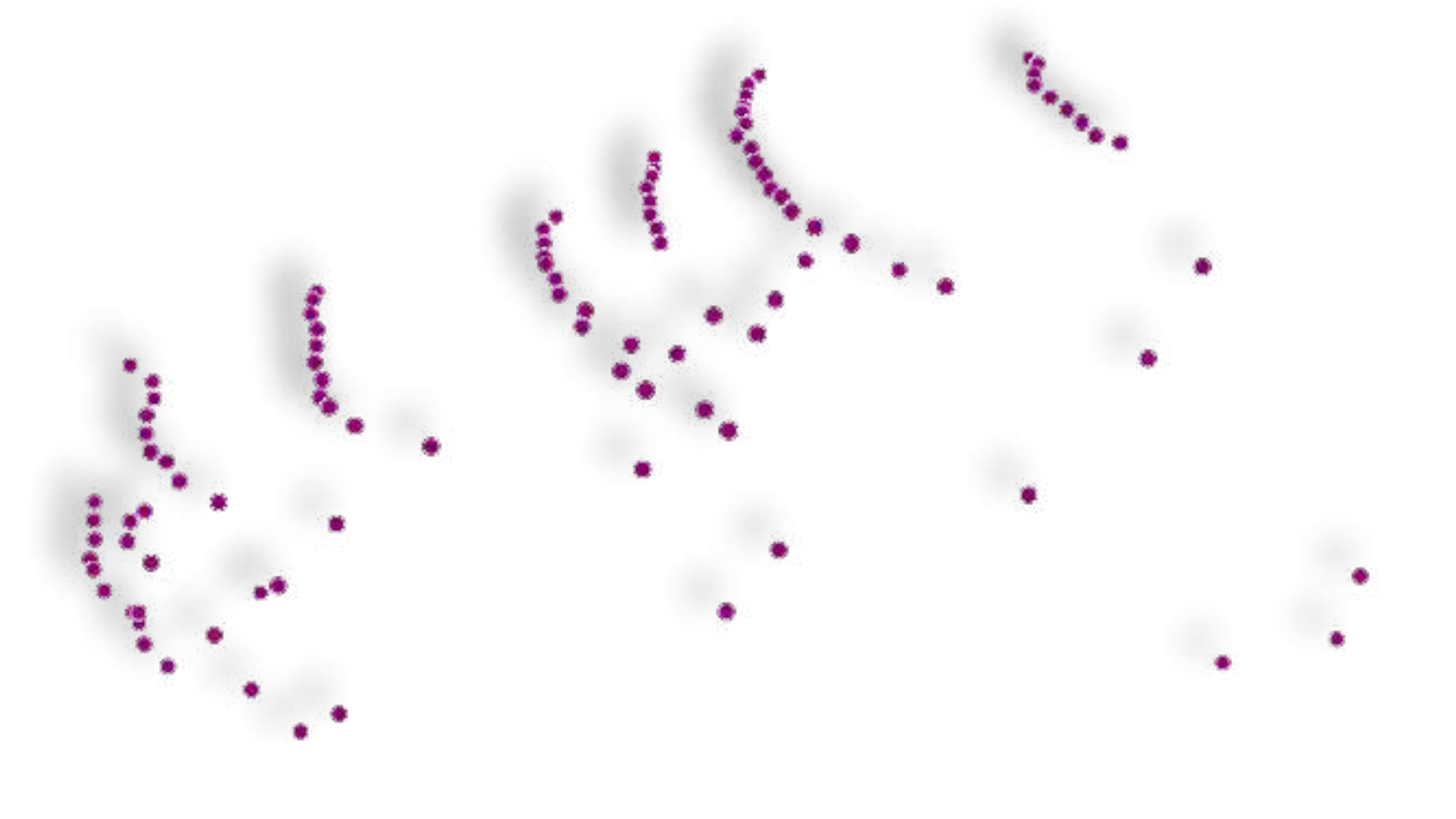}
				\includegraphics[width = .075\textwidth, page=2]
				{figures/KITTI_Result2.pdf}
				\includegraphics[width =.075\textwidth, page=4]
				{figures/KITTI_Result2.pdf}
				\includegraphics[width =.075\textwidth, page=5]				
				{figures/KITTI_Result2.pdf} 
				\includegraphics[width = .075\textwidth, page=7]
				{figures/KITTI_Result2.pdf}	
				\includegraphics[width = .075\textwidth, page=3]
				{figures/KITTI_Result2.pdf}	
				\vspace{1mm}
				\\  
			\setcounter{subfigure}{0}
			\begin{minipage}{.075\textwidth}
            \centering
            \includegraphics[width=1\linewidth, page=8]{figures/KITTI_Result2.pdf}
            \vspace{\abovecaptionskip}%
            \tiny Input
            \end{minipage}
            
            \begin{minipage}{.075\textwidth}
            \centering
            \includegraphics[width=1\linewidth, page=9]{figures/KITTI_Result2.pdf}
            \vspace{\abovecaptionskip}%
            \tiny GT
            \end{minipage}
            
            \begin{minipage}{.075\textwidth}
            \centering
            \includegraphics[width=1\linewidth, page=11]{figures/KITTI_Result2.pdf}
            \vspace{\abovecaptionskip}%
            \tiny GRNet\cite{GRNET}
            \end{minipage}
            
            \begin{minipage}{.075\textwidth}
            \centering
            \includegraphics[width=1\linewidth, page=12]{figures/KITTI_Result2.pdf}
            \vspace{\abovecaptionskip}%
            \tiny SFNet~\cite{SnowflakeNet}
            
            \end{minipage}
            \begin{minipage}{.075\textwidth}
            \centering
            \includegraphics[width=1\linewidth, page=14]{figures/KITTI_Result2.pdf}
            \vspace{\abovecaptionskip}%
            \tiny pcn  ~\cite{pcn}
            \end{minipage}            
            
            \begin{minipage}{.075\textwidth}
            \centering
            \includegraphics[width=1\linewidth, page=10]{figures/KITTI_Result2.pdf}
            \vspace{\abovecaptionskip}%
            \tiny Ours
            \end{minipage}			
            \setcounter{subfigure}{1}
            
	    \end{tabular}}
	\caption{\textbf{Evaluation on the SemanticKITTI~\cite{SemanticKITTI} dataset.}
    We compare our results with various supervised and unsupervised methods by a) evaluating our quantitative results in terms of precision~(P), coverage~(C), and Chamfer distance~(CD) and b) visualizing their outputs.
    For supervised methods, we use their pretrained models on the synthetic PCN~\cite{pcn} dataset to evaluate on the SemanticKITTI.
	}
	\label{fig:kitti3}
	\vspace{-4mm}
\end{figure*}
\subsection{ACL-SPC}
Using a conventional ACL as a point cloud completion system requires the target complete the point cloud and several partial point cloud observations to optimize the system.
However, accessing the target complete point cloud and several partial observations in real-world scenarios is not always possible.
Therefore, we develop the ACL system such that it generates the complete point cloud using only a single partial observation and without requiring the target complete point cloud for optimizing the system, as shown in Figure~\ref{fig:framework}.
To achieve this goal, we employ a learnable model $f_{\theta}$ on an input partial point cloud observation $P_0\in\mathbb{R}^{N_{\text{p}}\times3}$ as follows:
\begin{equation}\label{eq:CVF-SSPC1}
C_0=f_{\theta}(P_0),
\end{equation}
where $C_0\in\mathbb{R}^{N_{\text{c}}\times3}$ is the generated complete point cloud and $N_{\text{p}}$ and $N_{\text{c}}$ refers to the number of points in the input and output point cloud, respectively.
Then we apply a partial point cloud generator $g_v$ to generate a set of partial point clouds $P_v$ from the generated point cloud $C_{0}$ as follows:
\begin{equation}\label{eq:CVF-SSPC2}
\forall v\in\{v_i\}_{i=1}^{N_{\text{s}}}, \; P_v=g_v(C_{0}), 
\end{equation}
where $v_i$ is a random parameter to generate $N_{\text{s}}$ number of different partial point clouds.
We again employ the same model $f_{\theta}$ on the generated synthetic partial point clouds $P_v$ to generate the same point cloud $C_0$ as follows:
\begin{equation}\label{eq:CVF-SSPC3}
\forall v\in\{v_i\}_{i=1}^{N_{\text{s}}}, \; C_v=f_{\theta}(P_v)=f_{\theta}(g_v(C_{0})),
\end{equation}
where $C_v$ is the predicted complete point cloud for the generated point cloud $C_0$. 
Then, we optimize the system by a loss function between the predicted complete point clouds $C_{v}$ and the generated initial complete point cloud $C_0$.
Accordingly, our ACL-SPC learns to generate the same complete point cloud for different partial point cloud observations $P_v$ synthesized from the generated $C_0$.
Since the learnable model $f_{\theta}$ is optimized to map any partial point cloud $P_v$ to its corresponding target complete point cloud $C_0$, the generated point cloud $C_0$ as the output of $f_{\theta}$ on the input partial point cloud $P_0$ must predict the target complete point cloud. 
\subsection{Loss functions}
To train our network $f_{\theta}$, we use two self-supervised loss functions. 
First, to optimize the ACL-SPC and guarantee to generate the same predicted complete point clouds, we design the consistency loss function $\mathcal{L}^{\text{cons}}$ between the predicted complete point clouds $C_v$ and $C_0$ as follows:
\begin{equation}\label{eq:loss_cons}
\mathcal{L}^{\text{cons}} = \frac{1}{N_{\text{c}}\times N_{\text{s}}}\sum_{v\in\{v_i\}_{i=1}^{N_{s}}}||C_v-C_0||^2_2,
\end{equation}
where $||.||_2$ represents the $L_2$ norm. 
We further utilize the weighted Chamfer distance~\cite{PointPnCNet} loss $\mathcal{L}^{\text{wcd}}$ between the predicted complete point cloud $C_0$ and the input partial point cloud $P_0$.
The weighted Chamfer distance is invariant to the permutation of the order of points which is composed of two parts with the corresponding weights as follows:
\begin{equation}\label{eq:loss_chamfer}
\small
\mathcal{L}^{\text{wcd}} = \frac{\alpha}{N_{\text{c}}}\sum_{p\in C_0} \min_{q\in P_0}||p-q||_2 + \frac{\beta}{{N_\text{p}}}\sum_{q\in P_0} \min_{p\in C_0}||q-p||_2.
\end{equation}
The first term measures the mean distance for each point in the source point cloud $C_0$ to the closest point in the target point cloud $P_0$, while the second term measures the mean distance from each point in the target point cloud $P_0$ to its nearest point in the source point cloud $C_0$.  
Therefore, the second term leads the predicted point cloud $C_0$ to cover the points in the target point cloud $P_0$ while the first term performs as a regularizer.
We set $\alpha=0.1$ and $\beta=0.9$ to enforce the points to cover the non-missing parts of the point cloud and let the remaining points be flexible to fill in the missing parts.  
The total loss $\mathcal{L}^{\text{total}}$ is the weighted summation of two aforementioned loss functions as follows:
\begin{equation}\label{eq:loss_total}
\mathcal{L}^{\text{total}} 
=\lambda_{\text{cons}}\mathcal{L}^{\text{cons}}
+\mathcal{L}^{\text{wcd}}.
\end{equation}
%
%
%
%
\subsection{Training details}
\label{subsec:TrainingDetails}
Our model $f_{\theta}$ includes an encoder $\mathcal{E}$ which learns the local and global features from the partial input point clouds and a decoder $\mathcal{D}$ to generate the points of the complete point clouds as shown in Figure~\ref{fig:framework}.
We use PolyNet~\cite{polynet}, a powerful spatial graph CNN, as the encoder $\mathcal{E}$ which consists of four squeezed PolyConv layers with the sizes of $64$, $128$, $256$, and $512$, respectively.
We apply a random down-sampling followed by max-pooling after the first three PolyConv layers to reduce the point size to $512$, $128$, and $32$, respectively.
We employ a global average pooling after the last PolyConv layer to eliminate the spatial dependency and obtain $512$ invariant features to various partial observations and point permutations. 
We use three fully connected~(FC) layers as the decoder $\mathcal{D}$ with the sizes of $1024$, $1024$, and ${N_{\text{c}}}\times3$, respectively, where the ReLU non-linear activation function is applied on the outputs of the first and the second FC layers.
In fact, $g_v$ is a non-learnable function that generates the synthetic partial point clouds by projecting the generated complete point cloud to a depth map at a random view $v$ from azimuth [$0\degree$, $360\degree$] and elevation [$-20\degree$, $40\degree$].
We then back-project the depth map into 3D.
To avoid double backpropagation and optimize $f_{\theta}$ only once, we use the detach operator~\cite{Pytorch} as shown in Figure~\ref{fig:framework}.
We exploit the Adam optimizer and optimize the model for every $32$ different input partial point clouds and their $N_{\text{s}}$ synthetic partial point clouds and set $N_{s}=8$ and $\lambda_{\text{cons}}=10$ in our baseline.
For the inference, we feed the input partial point cloud to the trained model $f_{\theta}$ to directly generate the corresponding complete point cloud, which takes $12ms$ on average for each sample with a NVIDIA RTX 2080Ti.
%

\section{Experiments}
\label{Experiments}
\subsection{Datasets and metrics}
\label{subsec:Datasets}
In this section, we discuss the training and evaluation datasets and the metrics used to compare our proposed method ACL-SPC with the related methods.
\begin{table*}[t]
    \small
    \centering
    \setlength\tabcolsep{0.0001pt}
    \begin{tabularx}{\linewidth}{l 
    >{\centering\arraybackslash}X
    >{\arraybackslash}X
    >{\centering\arraybackslash}X
    >{\centering\arraybackslash}X 
    >{\centering\arraybackslash}X 
    >{\centering\arraybackslash}X 
    >{\centering\arraybackslash}X
    >{\centering\arraybackslash}X
    >{\centering\arraybackslash}X
    >{\centering\arraybackslash}X
    >{\centering\arraybackslash}X
    >{\centering\arraybackslash}X
    >{\centering\arraybackslash}X
    >{\centering\arraybackslash}X 
    >{\centering\arraybackslash}X 
    >{\centering\arraybackslash}X 
    >{\centering\arraybackslash}X 
    >{\centering\arraybackslash}X
    }

    \toprule
    \multirow{3}{*}{\textbf{Supervision}}& &
    \multirow{3}{*}{\textbf{Method}}&&& 
    \multicolumn{4}{c}{\textbf{ScanNet}} && \multicolumn{4}{c}{\textbf{MatterPort3D}} &&
    \multicolumn{2}{c}{\textbf{KITTI}}
    \\
    & &&&& \multicolumn{2}{c}{\textbf{Chair}} & \multicolumn{2}{c}{\textbf{Table}} && \multicolumn{2}{c}{\textbf{Chair}} & \multicolumn{2}{c}{\textbf{Table}}  && \multicolumn{2}{c}{\textbf{Car}}
    \\
    & &&&& {UCD$\downarrow$} & {UHD$\downarrow$} & {UCD$\downarrow$} & {UHD$\downarrow$} && {UCD$\downarrow$} & {UHD$\downarrow$} & {UCD$\downarrow$} & {UHD$\downarrow$} && {UCD$\downarrow$} & {UHD$\downarrow$} \\
    \midrule
    \multirow{6}{*}{Unsupervised} && pcl2pcl~\cite{pcl2pcl} &&& 17.3 & 10.1 & 9.1 & 11.8 && 15.9 & 10.5 & 6.0 &  11.8 && 9.2 &  14.1 
    \\
    && ShapeInversion~\cite{ShapeInversion} &&& 3.2 & 10.1 & 3.3 & 11.9 && 3.6 & 10.0 & 3.1 & 11.8 && 2.9 & 13.8
    \\
    && $+$UHD~\cite{ShapeInversion} &&& 4.0 &9.3 & 6.6 & 11.0 && 4.5 & 9.5 & 5.7 & 10.7	&& 5.3 & 12.5
    \\
    && Cycle4Comp.~\cite{Cycle4Completion} &&& 5.1 & 6.4	& 3.6 & 5.9	&& 8.0 & 8.4 & 4.2 & 6.8 && 3.3 & 5.8
    \\
    && DE~\cite{Optde} &&& 2.8 & 5.4 & 2.5 & 5.2 && 3.8 & 6.1 & 2.5 & 5.4 && 1.8 & {3.5}
    \\
    && OptDE~\cite{Optde} &&& 2.6 & 5.5 & 1.9 & \textbf{4.6} && 3.0 & 5.5 & \textbf{1.9} & 5.3 && \textbf{1.6} & \textbf{3.5}
    \\

    \midrule
    Self-supervised && \textbf{Ours} &&& \textbf{1.4} & \textbf{4.7} & \textbf{1.8} & {5.1} && \textbf{1.8} & \textbf{4.8} & {2.1} & \textbf{4.9} && {2.0} & {4.9}    
    \\
    \bottomrule
    \end{tabularx}

    \caption{
        \textbf{Quantitative results on the real-world datasets~\cite{ScanNet, Matterport3D, KITTI} in the categories of chair, table, and car.} We evaluate the method in terms of UCD and UHD where the values are multiplied by $10^2$ and $10^4$ respectively. 
        }
    \label{tab:scanmatterkitti}
    \vspace{-3mm}
\end{table*}

\begin{figure*}
     \centering
     \begin{subfigure}[b]{0.12\textwidth}
         \centering
         \includegraphics[trim={0cm 3cm 3cm 0cm},clip, width=\textwidth, page=5]{figures/Scannet_chair.pdf}
     \end{subfigure}
     \hfill
     \begin{subfigure}[b]{0.11\textwidth}
         \centering
         \includegraphics[trim={3cm 3cm 3cm 0cm},clip, width=\textwidth, page=6]{figures/Scannet_chair.pdf}
     \end{subfigure}
     \hfill
     \begin{subfigure}[b]{0.11\textwidth}
         \centering
         \includegraphics[trim={3cm 3cm 3cm 0cm},clip, width=\textwidth, page=7]{figures/Scannet_chair.pdf}
     \end{subfigure}
     \hfill
     \begin{subfigure}[b]{0.11\textwidth}
         \centering
         \includegraphics[trim={3cm 3cm 3cm 0cm},clip, width=\textwidth, page=8]{figures/Scannet_chair.pdf}
     \end{subfigure}
     \hfill
     \hspace{0.5cm}
     \begin{subfigure}[b]{0.11\textwidth}
         \centering
         \includegraphics[trim={2cm 2cm 2cm 0cm},clip, width=\textwidth, page=1]{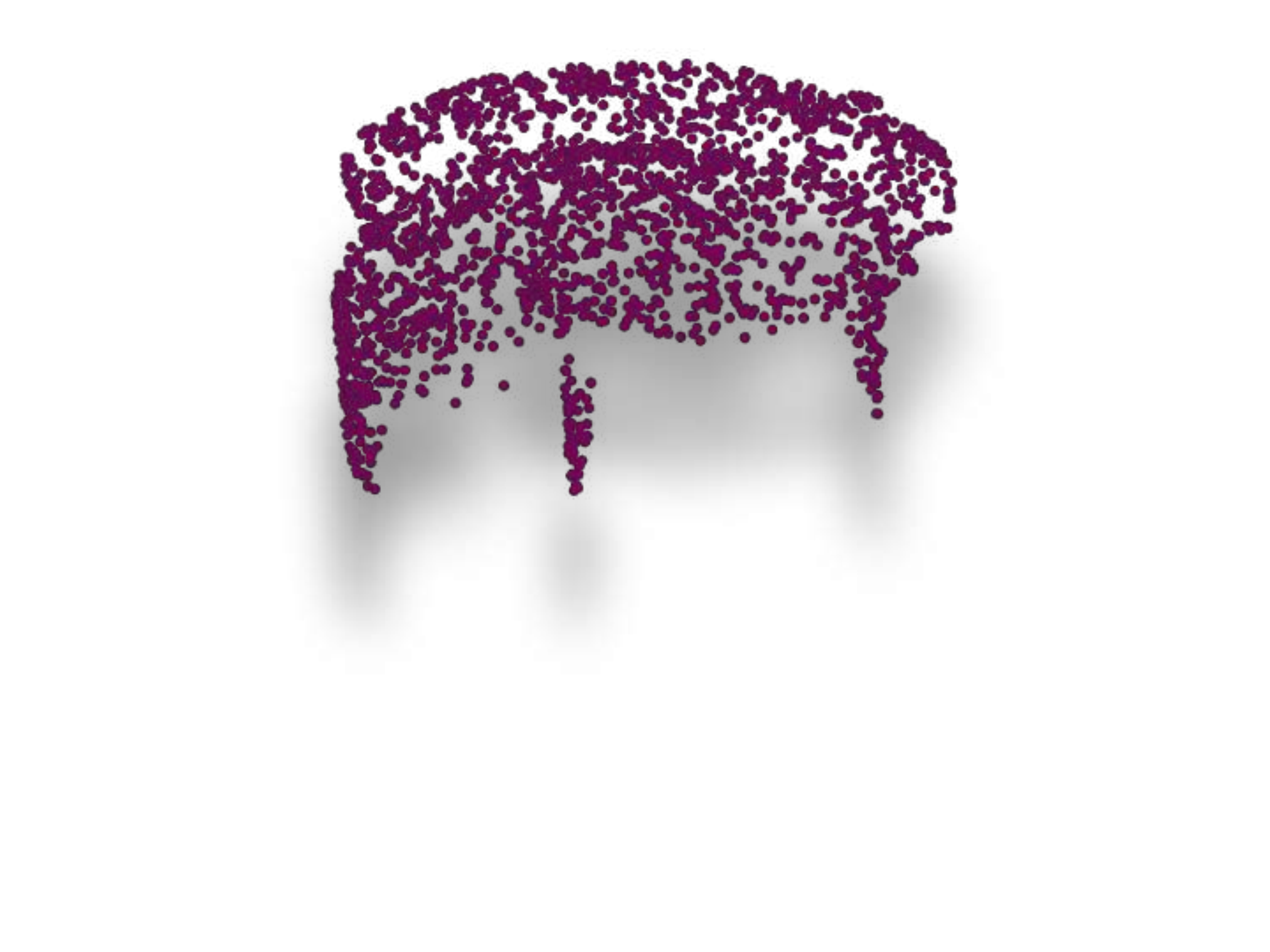}
     \end{subfigure}    
     \hfill
     \begin{subfigure}[b]{0.11\textwidth}
         \centering
         \includegraphics[trim={2cm 2cm 2cm 0cm},clip, width=\textwidth, page=2]{figures/Scannet_table.pdf}
     \end{subfigure}
     \hfill
     \begin{subfigure}[b]{0.11\textwidth}
         \centering
         \includegraphics[trim={2cm 2cm 2cm 0cm},clip, width=\textwidth, page=3]{figures/Scannet_table.pdf}
     \end{subfigure}
     \hfill
     \begin{subfigure}[b]{0.11\textwidth}
         \centering
         \includegraphics[trim={2cm 2cm 2cm 0cm},clip, width=\textwidth, page=4]{figures/Scannet_table.pdf}
     \end{subfigure}      
     \hfill
     \begin{subfigure}[b]{0.12\textwidth}
         \centering
         \includegraphics[trim={0cm 2cm 2cm 2cm},clip, width=\textwidth, page=1]{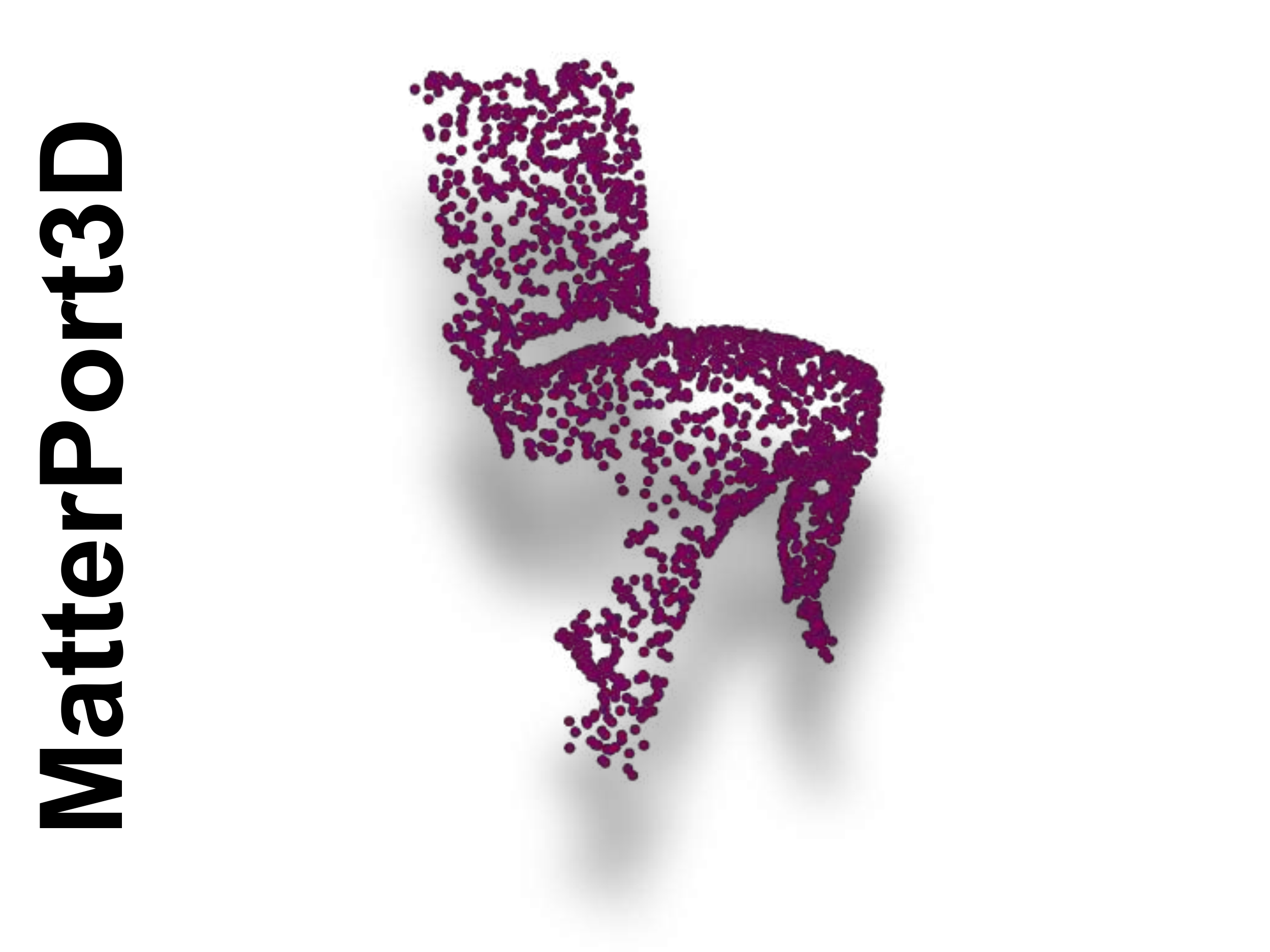}
     \end{subfigure}  
     \hfill
     \begin{subfigure}[b]{0.11\textwidth}
         \centering
         \includegraphics[trim={2cm 2cm 2cm 2cm},clip, width=\textwidth, page=2]{figures/MatterPort_chair.pdf}
     \end{subfigure}  
     \hfill
     \begin{subfigure}[b]{0.11\textwidth}
         \centering
         \includegraphics[trim={2cm 2cm 2cm 2cm},clip, width=\textwidth, page=3]{figures/MatterPort_chair.pdf}
     \end{subfigure} 
     \hfill
     \begin{subfigure}[b]{0.11\textwidth}
         \centering
         \includegraphics[trim={2cm 2cm 2cm 2cm},clip, width=\textwidth, page=4]{figures/MatterPort_chair.pdf}
     \end{subfigure} 
     \hfill
     \hspace{0.5cm}
     \begin{subfigure}[b]{0.11\textwidth}
         \centering
         \includegraphics[trim={2cm 2cm 2cm 2cm},clip, width=\textwidth, page=1]{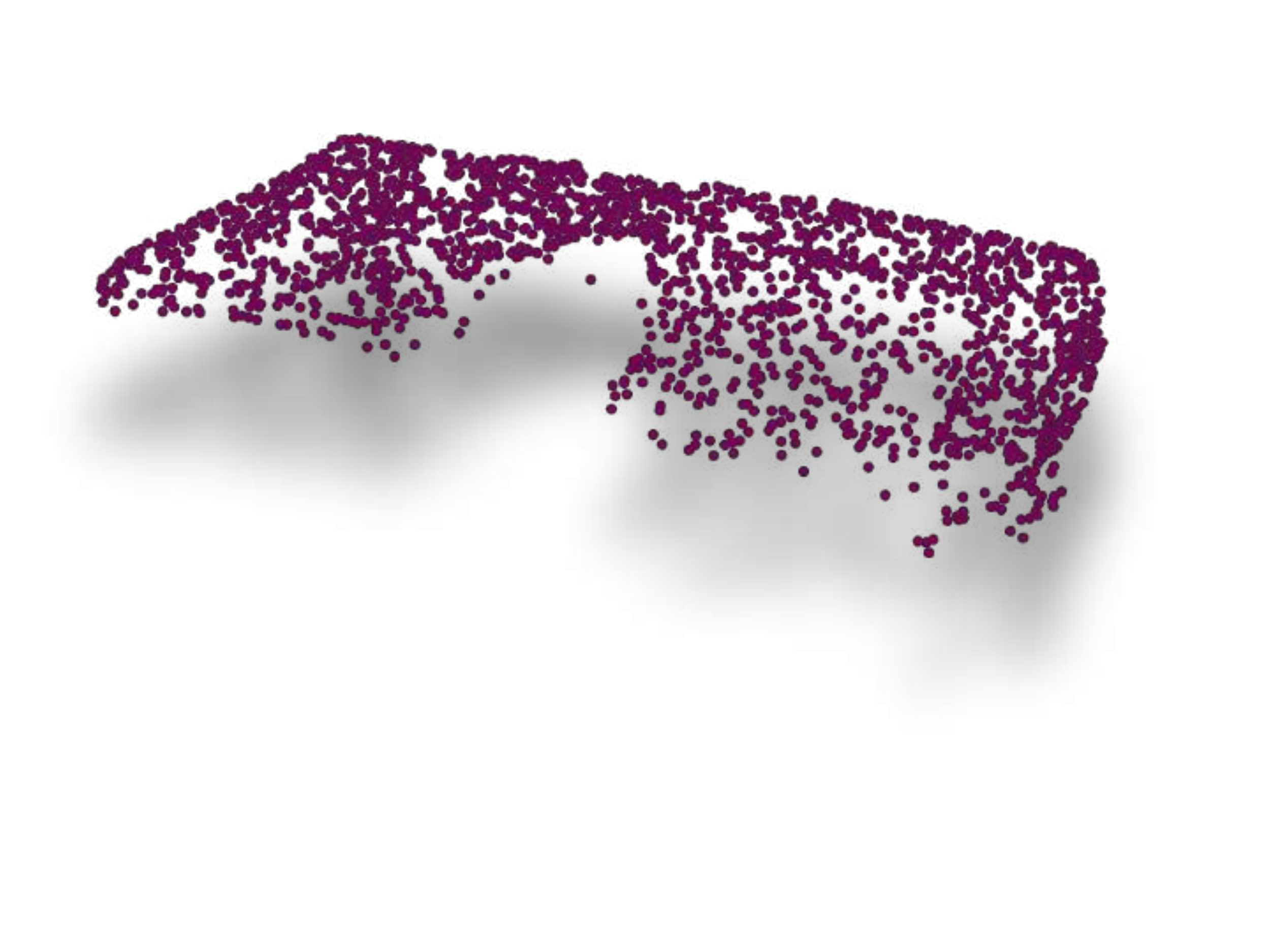}
     \end{subfigure} 
     \hfill
     \begin{subfigure}[b]{0.11\textwidth}
         \centering
         \includegraphics[trim={2cm 2cm 2cm 2cm},clip, width=\textwidth, page=2]{figures/MatterPort_table.pdf}
     \end{subfigure} 
     \hfill
     \begin{subfigure}[b]{0.11\textwidth}
         \centering
         \includegraphics[trim={2cm 2cm 2cm 2cm},clip, width=\textwidth, page=3]{figures/MatterPort_table.pdf}
     \end{subfigure} 
     \hfill
     \begin{subfigure}[b]{0.11\textwidth}
         \centering
         \includegraphics[trim={2cm 2cm 2cm 2cm},clip, width=\textwidth, page=4]{figures/MatterPort_table.pdf}
     \end{subfigure} 
     \hfill
     \begin{subfigure}[b]{0.12\textwidth}
         \centering
         \includegraphics[trim={0cm 2cm 2cm 2cm},clip, width=\textwidth, page=1]{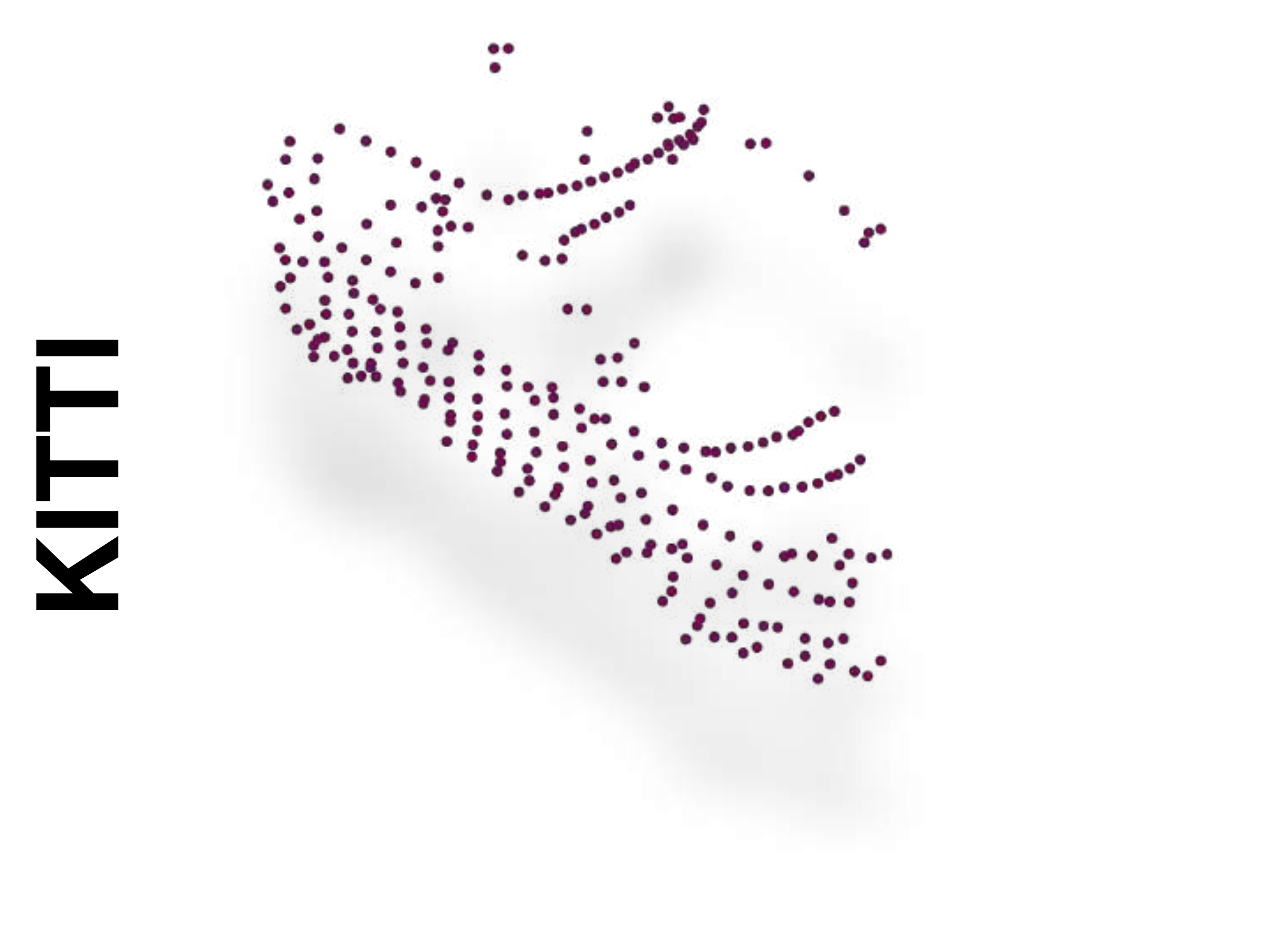}
         \caption*{Input}
     \end{subfigure} 
     \hfill
     \begin{subfigure}[b]{0.11\textwidth}
         \centering
         \includegraphics[trim={2cm 2cm 2cm 2cm},clip, width=\textwidth, page=2]{figures/KITTI_car.pdf}
         \caption*{ShapeInv.~\cite{ShapeInversion}}
     \end{subfigure} 
     \hfill
     \begin{subfigure}[b]{0.11\textwidth}
         \centering
         \includegraphics[trim={2cm 2cm 2cm 2cm},clip, width=\textwidth, page=3]{figures/KITTI_car.pdf}
         \caption*{OptDE~\cite{Optde}}
     \end{subfigure} 
     \hfill
     \begin{subfigure}[b]{0.11\textwidth}
         \centering
         \includegraphics[trim={2cm 2cm 2cm 2cm},clip, width=\textwidth, page=4]{figures/KITTI_car.pdf}
         \caption*{Ours}
     \end{subfigure} 
     \hfill
     \hspace{0.5cm}
     \begin{subfigure}[b]{0.11\textwidth}
         \centering
         \includegraphics[trim={2cm 2cm 2cm 1cm},clip, width=\textwidth, page=5]{figures/KITTI_car.pdf}
         \caption*{Input}
     \end{subfigure} 
     \hfill
     \begin{subfigure}[b]{0.11\textwidth}
         \centering
         \includegraphics[trim={2cm 2cm 2cm 1cm},clip, width=\textwidth, page=6]{figures/KITTI_car.pdf}
         \caption*{ShapeInv.~\cite{ShapeInversion}}
     \end{subfigure} 
     \hfill
     \begin{subfigure}[b]{0.11\textwidth}
         \centering
         \includegraphics[trim={2cm 2cm 2cm 1cm},clip, width=\textwidth, page=7]{figures/KITTI_car.pdf}
         \caption*{OptDE~\cite{Optde}}
     \end{subfigure} 
     \hfill
     \begin{subfigure}[b]{0.11\textwidth}
         \centering
         \includegraphics[trim={2cm 2cm 2cm 1cm},clip, width=\textwidth, page=8]{figures/KITTI_car.pdf}
         \caption*{Ours}
     \end{subfigure} 
        \caption{
        \textbf{Qualitative results on the real-world datasets~\cite{ScanNet, Matterport3D, KITTI} in the categories of chair, table, and car.}} 
        \label{fig:real_world_qual}
        \vspace{-4mm}
\end{figure*}     

\noindent
\textbf{Synthetic Datasets:} 
  ShapeNet~\cite{ShapeNet} is a large-scaled dataset including curated 3D shapes represented by CAD models, which consists of $55$ categories. 
  Among them, we focus on three categories, airplanes, cars, and chairs, to maleaaintain the same setup as the previous works~\cite{WeakPCN, PointPnCNet, DPC}.
  Followed by these works, we capture RGB-D data from five random views for each object, transfer them to 3D, and resample $3096$ points of them to generate a set of partial point clouds.
  The ground truth complete point clouds with the fixed $8192$ number of points are used for evaluation.
  %
  %
  %
  
\noindent
\textbf{Real-World Datasets:} 
  Similar to previous works~\cite{pcl2pcl, ShapeInversion, Optde}, we evaluate our method using three sources of real scans: ScanNet~\cite{ScanNet} (chairs and tables), MatterPort3D~\cite{Matterport3D} (chairs and tables), and KITTI~\cite{KITTI} (car). 
  The ScanNet and MatterPort3D datasets are richly annotated 3D reconstructions of indoor environments, whereas the KITTI dataset is of outdoor scenes.
  We resample the points to 2048 points to match the previous work's settings~\cite{pcl2pcl, ShapeInversion, Optde}. 
  %
  %
  SemanticKITTI dataset~\cite{SemanticKITTI} is derived from the KITTI dataset~\cite{KITTI}, including only the car objects which are captured in multi-views from sequence $00$ to $10$ when parked.
  We take out the sequence $08$ for testing and exploit the other sequences for training.
  Note that the input points are resampled to $1024$ for convenience.
  As the SemanticKITTI dataset has no ground truth complete point cloud, we follow the same steps as in the previous work~\cite{WeakPCN} to generate them by aggregating partial point clouds.\\
\noindent
\textbf{Metrics.} 
We utilize the Chamfer distance~(CD) between the reconstructed point cloud and the ground truth to evaluate the performance of our ACL-SPC method. 
Chamfer distance is the average distance between each point in a point cloud and the nearest point in the other as follows:
\begin{equation}\label{eq:loss_chamfer}
\begin{split}
\mathcal{CD}(C_0, \text{GT})= {1\over {N_c}}\sum_{p\in C_0} \min_{q\in GT}||p-q||_2  
\\ + {1\over {N_g}}\sum_{q\in GT} \min_{p\in C_0}||q-p||_2,
\end{split}
\end{equation}
where GT is defined as the ground truth complete point cloud with $N_g$ points.
The first and second term refers to precision and coverage, respectively.
Precision infers how much the generated points are distributed well compared to the ground-truth data,
while coverage refers to how much the missing parts of the partial point cloud are filled in.
Accordingly, the coverage is an important metric for point cloud completion tasks, reflecting the effectiveness of the methods to fill the missing parts.
Additionally, we utilize two metrics called Unidirectional Chamfer Distance~(UCD) and Unidirectional Hasudorff Distance~(UHD) for the real-world datasets~\cite{ScanNet, Matterport3D, KITTI} in the same way as previous works~\cite{pcl2pcl, ShapeInversion, Optde}.
To calculate UCD, we obtain the first term of $\mathcal{CD}(P_0, C_0)$ between the partial input point cloud $P_0$ and the predicted complete point cloud $C_0$ using equation~\ref{eq:loss_chamfer}.
Similarly, we measure the UHD with the single side of Hausdorff distance as follows:
\begin{equation}\label{eq:UHD}
\mathcal{UHD}(P_0, C_0)= \max_{p\in P_0}\min_{q\in C_0}||p-q||_2
\end{equation}
Although the two metrics do not reflect the completeness of shape, they can give a fair comparison where ground-truth data is unavailable.
%
%
\subsection{Evaluation on synthetic dataset}
\label{subsec:Eval_shapeNet}
In this section, we qualitatively and quantitatively evaluate our ACL-SPC method on the ShapeNet dataset and compare the results with the related methods~\cite{WeakPCN,PointPnCNet,DPC}.
We train our network for each category separately with $1000$ epochs by a learning rate of $0.001$, which is decayed by $0.5$ for every $200$ epochs to generate $N_c=8192$ points.
We visualize and compare the results of our method with the supervised~\cite{GRNET} and unsupervised~\cite{WeakPCN} methods in Figure~\ref{fig:qual_figure}.
We note that GRNet~\cite{GRNET} and Gu~\etal~\cite{WeakPCN} utilize the GT and the multi-view information as their supervision, respectively. 
Using the multi-view information leads to achieving a high-quality appearance for Gu~\etal~\cite{WeakPCN} because the concatenated point cloud from five random partial point clouds is almost as the GT as shown in Figure~\ref{subfig:multiview}. 
Even without this information, our ACL-SPC method shows comparable results in completing the missing parts of the input in a fully self-supervised manner.
Moreover, our quantitative results in Table~\ref{tab:shapenet} show that our method can outperform the unsupervised methods DPC~\cite{DPC} and PointPnCNet~\cite{PointPnCNet} with a large gap $1.01$ and $1.68$ with the CD on average while performing only $0.23$ lower performance compared to Gu~\etal method.
%
%
Therefore, our method can learn even better without any prior information compared to some of the unsupervised methods that have leveraged multiple partial views.
Moreover, our method outperforms all the unsupervised methods by the coverage metric, which shows its superiority in covering the missing parts.
%
%
%
%
\subsection{Evaluation on real-world dataset}
\label{subsec:SemanticKITTI}
We evaluate our self-supervised method ACL-SPC on SemanticKITTI~\cite{SemanticKITTI} dataset and compare the results with both supervised and unsupervised methods.
We train our network for $500$ epochs with a learning rate of $0.001$, and it is decayed by $0.5$ for every $200$ epochs to output $N_c=8192$ number of points.
We exploit pretrained models of the supervised methods~\cite{GRNET, SnowflakeNet} on the synthetic PCN~\cite{pcn} dataset to test on the real-world SemanticKITTI dataset.
%
%
As shown in Figure~\ref{fig:QuantKITTI}, our method outperforms the unsupervised methods Gu~\etal~\cite{WeakPCN} and PointPnCNet~\cite{PointPnCNet} in terms of coverage and CD. 
It also achieves better coverage compared to the supervised method GRNet~\cite{GRNET}, while it shows superior performance in all metrics compared to the supervised method SFNet~\cite{SnowflakeNet}.
Moreover, we can see through Figure~\ref{tab:QualiKITTI} that the supervised methods perform poorly on the real-world dataset compared to our method due to the domain gap with the synthetic dataset, which emphasizes the generalizability of our self-supervised method in real-world scenarios.
Additionally, we can validate that the coverage is more important than other metrics as our method shows better qualitative results than GRNet~\cite{GRNET} even though our method shows larger precision and CD values.

Furthermore, we quantitatively and qualitatively evaluate our method on ScanNet~\cite{ScanNet}, MatterPort3d~\cite{Matterport3D}, and KITTI~\cite{KITTI} dataset as shown in Table~\ref{tab:scanmatterkitti} and Figure~\ref{fig:real_world_qual}, respectively.
In contrast to the unsupervised methods~\cite{pcl2pcl, ShapeInversion, Cycle4Completion, Optde} that require synthetic datasets in addition to the real-world datasets for training because they either need unpaired ground truth~\cite{pcl2pcl, Cycle4Completion} or pretrained model~\cite{Cycle4Completion, Optde}, our method is trained only on real-world datasets.
Except for some metrics in the table categories, our method generally performs better than the state-of-the-art~\cite{Optde} in ScanNet and MatterPort3D datasets as shown in Table~\ref{tab:scanmatterkitti}. 
However, in the KITTI dataset, our method is slightly behind the state-of-the-art by $0.4$ and $1.4$ on UCD and UHD, respectively.
We also qualitatively compare the results with unsupervised methods~\cite{ShapeInversion, Optde} as shown in Figure~\ref{fig:real_world_qual}.
The ShapeInversion~\cite{ShapeInversion} generally fails to generate the missing parts in some cases, such as the chair class of MatterPort3D datasets. 
Meanwhile, OptDE~\cite{Optde} generates lots of noise in most samples, especially in the categories of the ScanNet~\cite{ScanNet} dataset.
In contrast, our method generates plausible points in the missing parts of the input in all samples.
One drawback of our results is that the output point cloud is not uniformly distributed, being sparser in the input regions where points were absent. 
Thus, even without requiring any synthetic datasets, we illustrate that our method is competitive compared to other unsupervised methods. 
\subsection{Ablation study}
In this section, we further analyze our novel ACL-SPC method by extensive ablations studies on test-time adaptation, the effect of defined loss functions, the number of synthesized data, training on the multi-class dataset, and training on a dataset including only one view of objects.
\subsubsection{Test-time adaptation}
\begin{table}
    \small
    \centering
    \setlength\tabcolsep{0.0001pt}
    \begin{tabularx}{0.8\linewidth}{l 
    >{\centering\arraybackslash}X 
    >{\centering\arraybackslash}X 
    >{\centering\arraybackslash}X
    >{\centering\arraybackslash}X }
    \toprule
    \textbf{Supervision}&
    \textbf{P}$\downarrow$ & 
    \textbf{C}$\downarrow$ &
    \textbf{CD}$\downarrow$\\
    \midrule
    Supervised & {17.29} & {8.57} & {25.86}  \\
    Self-supervised & {11.67} & \textbf{5.63} & {17.30}   \\
    Test-time adapt. & \textbf{9.62} & {7.09} & \textbf{16.71} \\
    \bottomrule
    \end{tabularx}
    \caption{
        \textbf{Evaluation on Test-time adaptation.} We train the network in three modes: supervised, self-supervised, and test-time adaptation.
        The values are multiplied by 100. }
    \label{tab:test-adaptation}   
    \vspace{-4mm}
\end{table}
\begin{figure*}
	\captionsetup[]{labelformat=empty}
		\newcommand{\rowArg}{1.53cm}
		\newcommand{\fullSize}{4.35cm}
		\newcommand{\patchSize}{1.5cm}
		\setlength\tabcolsep{0.05cm}
    \subfloat[Quantitative results.]{
    \footnotesize
    \begin{tabularx}{0.5\linewidth}{l >{\centering\arraybackslash}X >{\centering\arraybackslash}X >{\centering\arraybackslash}X>{\centering\arraybackslash}X>{\centering\arraybackslash}X>{\centering\arraybackslash}X>{\centering\arraybackslash}X>{\centering\arraybackslash}X>{\centering\arraybackslash}X>
    {\centering\arraybackslash}X>
    {\centering\arraybackslash}X}
    \toprule
    \multirow{2}{*}{\textbf{Loss}}&
    \multicolumn{3}{c}{\textbf{Airplane}}  && 
    \multicolumn{3}{c}{\textbf{Car}} &&
    \multicolumn{3}{c}{\textbf{Chair}} 
    \\
    & P$\downarrow$ & C$\downarrow$ & CD$\downarrow$ && P$\downarrow$ & C$\downarrow$ & CD$\downarrow$ && P$\downarrow$ & C$\downarrow$ & CD$\downarrow$
    \\
    \midrule
    $-\mathcal{L}^{wcd}$ 
    & {1.15} & {18.76} & {19.91} && {4.66} & {26.51} & {31.17} &&  {3.91} & {26.41} & {30.32} \\
    $-\mathcal{L}^{cons}$
    & \textbf{1.10} & {1.22} & {2.32} && \textbf{1.45} & {2.11} & {3.56} && \textbf{1.99} & {1.74} & {3.73}\\
    \midrule
    \hspace{2mm}$\mathcal{L}^{total}$
    & {1.20} & \textbf{0.81} & \textbf{2.01} && {1.65} & \textbf{1.28} & \textbf{2.93} && {2.25} & \textbf{1.46} & \textbf{3.71}\\
    \bottomrule
    \end{tabularx}\label{tab:loss}
    \vspace{4mm}
    }
    \hspace{1cm}
     \subfloat[Qualitative results. \label{tab:qual_effect_loss}]{
     \footnotesize
		\begin{tabular}[b]{c c c c}
				\includegraphics[page=1, width = \patchSize]{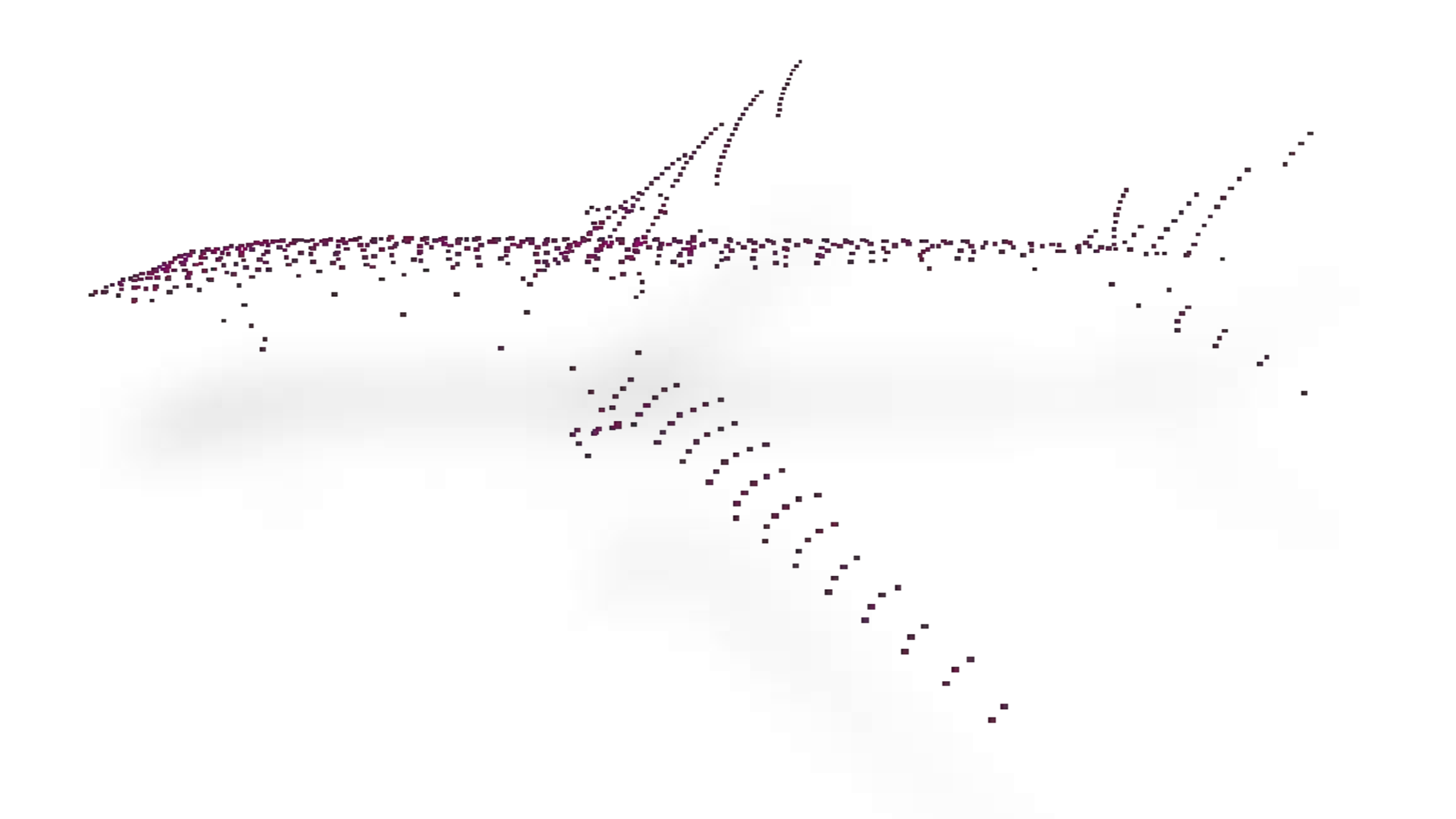}
				\hfill
				\includegraphics[page=2, width = \patchSize]{figures/exp_airplane_V13.pdf}
				\hfill
				\includegraphics[page=4, width = \patchSize]{figures/exp_airplane_V13.pdf}
				\hfill
				\includegraphics[page=3, width = \patchSize]{figures/exp_airplane_V13.pdf} \\  
				\includegraphics[page=1, width = \patchSize]{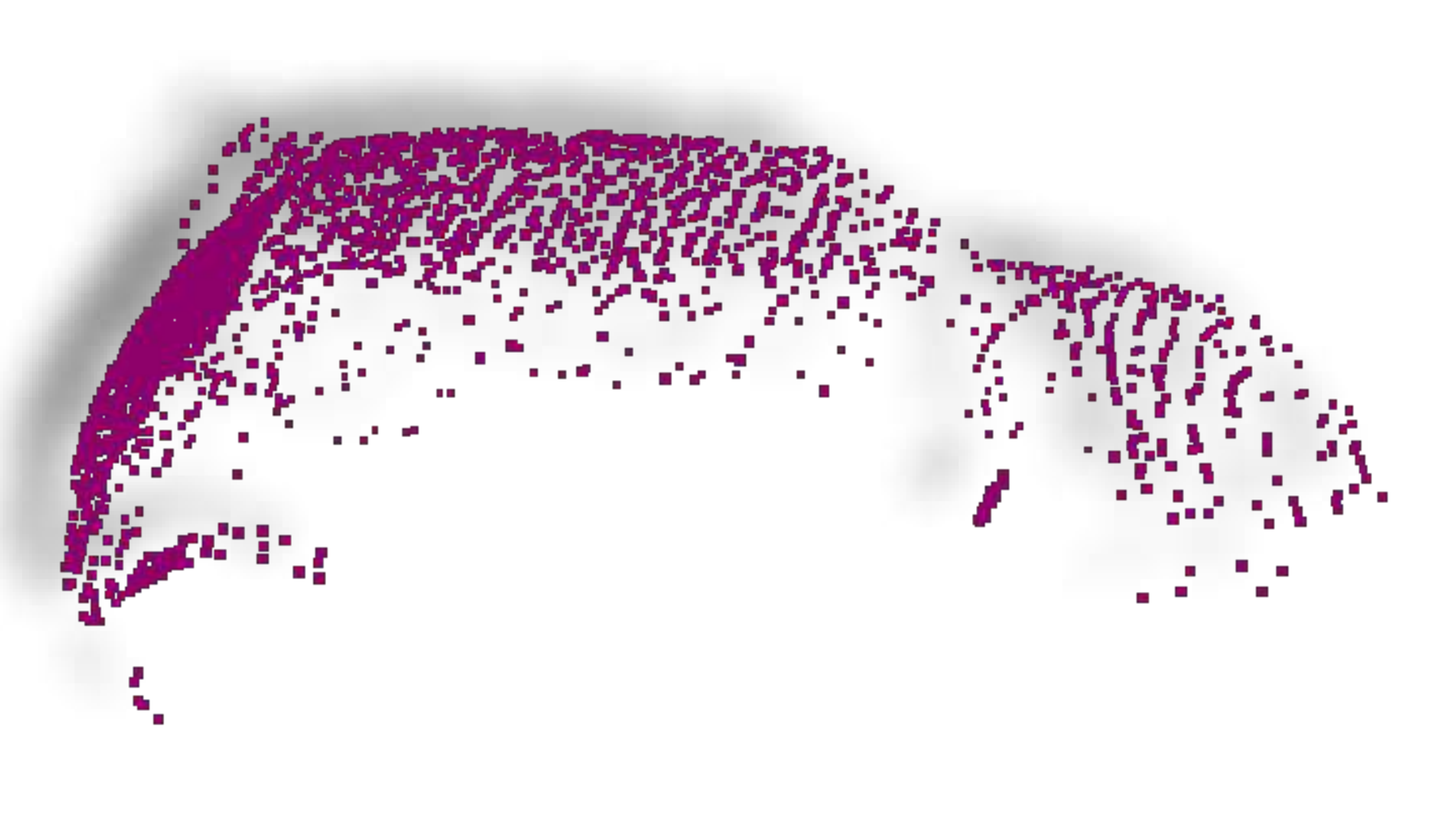}
				\hfill
				\includegraphics[page=2, width = \patchSize]{figures/exp_car_v8.pdf}
				\hfill
				\includegraphics[page=4, width = \patchSize]{figures/exp_car_v8.pdf}		
				\hfill
				\includegraphics[page=3, width = \patchSize]{figures/exp_car_v8.pdf} \\  
			\setcounter{subfigure}{0}
            \begin{minipage}{\patchSize}
            \centering
			\includegraphics[page=1, width=\patchSize]{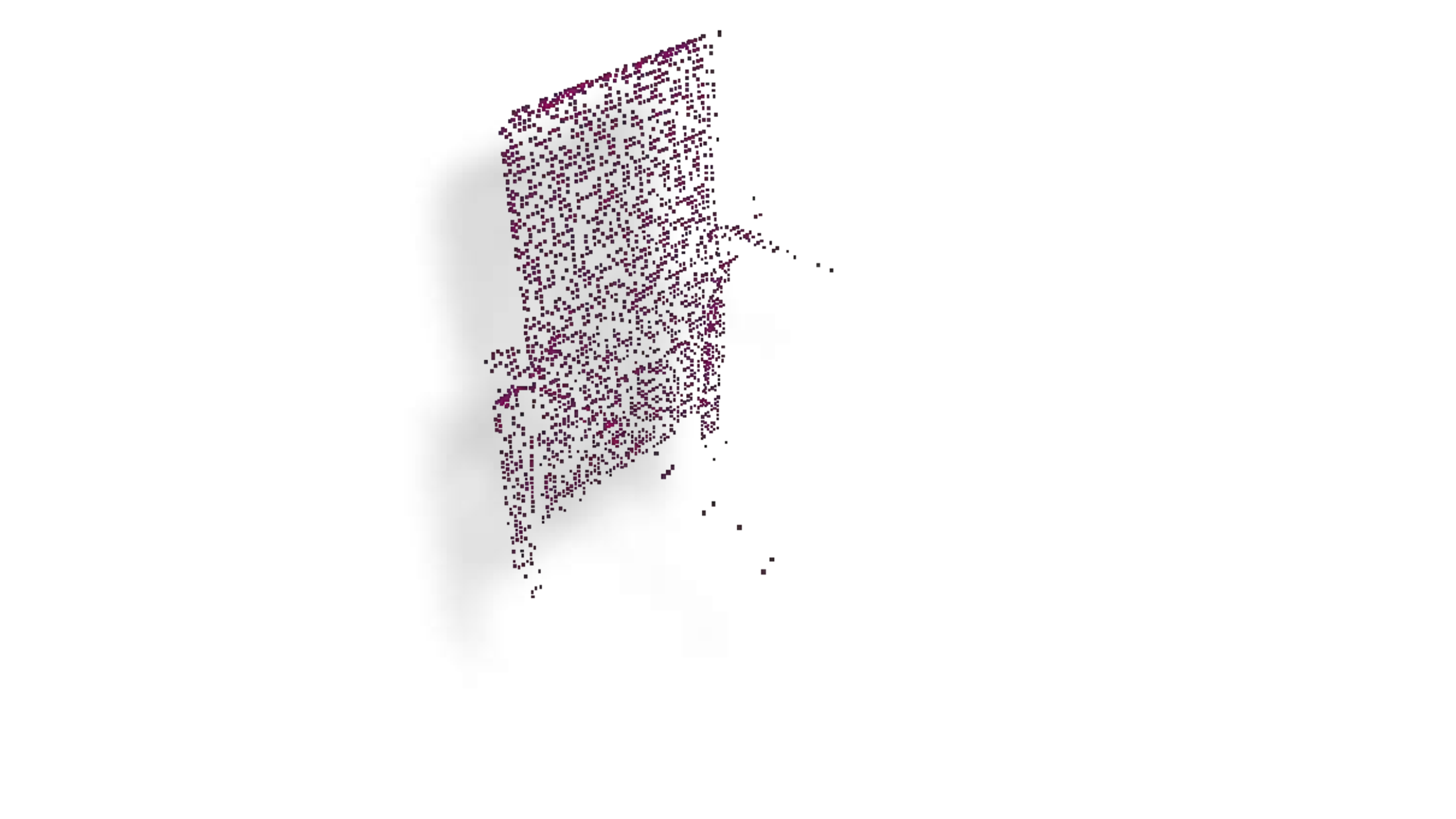}
            \vspace{\abovecaptionskip}%
            \tiny Input
            \end{minipage}
            \begin{minipage}{\patchSize}
            \centering
			\includegraphics[page=2,
			width =\patchSize]{figures/exp_chair_v17.pdf}
            \vspace{\abovecaptionskip}%
            \tiny GT
            \end{minipage} 
            \begin{minipage}{\patchSize}
            \centering
			\includegraphics[page=4, width = \patchSize]{figures/exp_chair_v17.pdf}
            \vspace{\abovecaptionskip}%
            \tiny $-\mathcal{L}^{\text{cons}}$
            \end{minipage} 
            \begin{minipage}{\patchSize}
            \centering
			\includegraphics[page=3, width = \patchSize]{figures/exp_chair_v17.pdf}
            \vspace{\abovecaptionskip}%
            \tiny $\mathcal{L}^{\text{total}}$ 
            \end{minipage}            
            
            
    \setcounter{subfigure}{1}
	\end{tabular}}
     \hfill
        \caption{\textbf{Evaluation on the effects of loss functions.} We show a) precision~(P) and coverage~(C) values multiplied by 100 for each experiment with different losses, and b) qualitative results.}
        \vspace{-3mm}
        \label{fig:effect_loss}
\end{figure*}     

\begin{table*}[t]
    \small
    \centering
    \setlength\tabcolsep{0.005pt}
    \begin{tabularx}{\linewidth}{l 
    >{\centering\arraybackslash}X 
    >{\centering\arraybackslash}X 
    >{\centering\arraybackslash}X  
    >{\centering\arraybackslash}X
    >{\centering\arraybackslash}X
    >{\centering\arraybackslash}X
    >{\centering\arraybackslash}X
    >{\centering\arraybackslash}X
    >{\centering\arraybackslash}X
    >{\centering\arraybackslash}X  
    >{\centering\arraybackslash}X
    >{\centering\arraybackslash}X  
    >{\centering\arraybackslash}X
    >{\centering\arraybackslash}X   
    >{\centering\arraybackslash}X  
    >{\centering\arraybackslash}X
    >{\centering\arraybackslash}X   
    >{\centering\arraybackslash}X 
    }
    
    \toprule
    \multirow{2}{*}{\textbf{Ablation}}&
    \multicolumn{2}{c}{\multirow{2}{*}{\textbf{Setup}}}&&
    \multicolumn{3}{c}{\textbf{Airplane}} &&
    \multicolumn{3}{c}{\textbf{Car}}&&
    \multicolumn{3}{c}{\textbf{Chair}}&&
    \multicolumn{3}{c}{\textbf{Average}}\\
    &&&& P$\downarrow$& C$\downarrow$ & CD$\downarrow$ && P$\downarrow$ & C$\downarrow$ & CD$\downarrow$ && P$\downarrow$ & C$\downarrow$ & CD$\downarrow$ && P$\downarrow$ & C$\downarrow$ & CD$\downarrow$
    \\
    \midrule
    & 1 & &
    & \textbf{1.19} & {0.85} & {2.04} & 
    & {1.63} & \textbf{1.27} & \textbf{2.90} & 
    & {2.37} & {1.45} & {3.82} & 
    & {1.73} & {1.19} & {2.92} \\
    Num Syns & 4 &  &
    & {1.23} & {0.82} & {2.05} & 
    & \textbf{1.62} & {1.30} & {2.92} & 
    & {2.55} & \textbf{1.43} & {3.97} & 
    & {1.80} & \textbf{1.18} & {2.98} \\ 
    &8&  &
    & {1.20} & \textbf{0.81} & \textbf{2.01} & 
    & {1.65} & {1.28} & {2.93} & 
    & \textbf{2.25} & {1.46} & \textbf{3.71} & 
    & \textbf{1.70} &\textbf {1.18} & \textbf{2.89} \\ 
    \midrule
    \multirow{2}{*}{Class} & Single &  &
    & \textbf{1.20} & {0.81} & \textbf{2.01} & 
    & \textbf{1.65} & {1.28} & {2.93} & 
    & \textbf{2.25} & {1.46} & \textbf{3.71} & 
    & \textbf{1.70} & {1.18} & \textbf{2.89} \\ 
    & Multi&  &
    & {1.40} & \textbf{0.79} & {2.19} & 
    & {1.66} & \textbf{1.25} & \textbf{2.91} & 
    & {2.35} & \textbf{1.42} & {3.76} & 
    & {1.80} & \textbf{1.15} & {2.96} \\ 
    \midrule
    \multirow{2}{*}{Views} & 1&  &
    & {1.23} & {0.89} & {2.12} & 
    & \textbf{1.63} & \textbf{1.27} & \textbf{2.90} & 
    & \textbf{2.15} & {1.54} & \textbf{3.69} & 
    & \textbf{1.67} & {1.23} & {2.90} \\ 
     & 5 &  &
    & \textbf{1.20} & \textbf{0.81} & \textbf{2.01} & 
    & {1.65} & {1.28} & {2.93} & 
    & {2.25} & \textbf{1.46} & {3.71} & 
    & {1.70} & \textbf{1.18} & \textbf{2.89}\\ 
    \bottomrule
    \end{tabularx}
    \caption{
        \textbf{Quantitative effects of the number of synthetic partial views, single-/multi-class training, and single-/multi-view training.} 
        We present the values of precision~(P), coverage~(C), and Chamfer distance~(CD) multiplied by 100.
        %
        %
        %
        }
    \label{tab:ablation}
    \vspace{-4.5mm}
\end{table*}
Similar to previous works~\cite{pcl2pcl, ShapeInversion, Cycle4Completion, Optde}, we show that our method can also be effective for test-time adaptation.
We train and test our network in three different schemes, as shown in Table~\ref{tab:test-adaptation}. 
First, we train the network in the supervised setting on a synthetic dataset and then evaluate it on a real-world dataset.
Second, we train our network in a self-supervised manner, without any pretraining, and train and test it on the real-world dataset. 
Finally, on the test-adaptation setting, it goes through the pretraining stage and then moves on to the adaptation stage with our ACL-SPC framework.
Table~\ref{tab:test-adaptation} displays the precision, coverage, and CD of each experiment setting.
The results show the suitability of our ACL-SPC method not only for self-supervised learning but also for test-time adaption. 
\subsubsection{Effect of each loss}
We evaluate the effects of each loss by taking out each loss at a time for each experiment.
We report the quantitative results for the experiments without $\mathcal{L}^{\text{wcd}}$, without $\mathcal{L}^{\text{cons}}$, and with the total loss $\mathcal{L}^{\text{total}}$ in Figure~\ref{tab:loss}.
Taking out $\mathcal{L}^{wcd}$ affects critically in the results as there is no guarantee to cover points in the input. 
On the other hand, excluding $\mathcal{L}^{\text{cons}}$ results in a worse coverage value which proves the importance of our proposed ACL-SPC to fill the missing parts of partial input point clouds.
To visualize this effect, we qualitatively compare our results with and without $\mathcal{L}^{\text{cons}}$ in Figure~\ref{tab:qual_effect_loss}. 
Without $\mathcal{L}^{\text{cons}}$, only the input is covered while the missing parts are still uncovered.
We note that since without $\mathcal{L}^{\text{wcd}}$ there is no constraint to generate points in the location of the input, 
the method produce all points in the same position.
%
\subsubsection{Number of synthesized data}
\label{subsubsec:synthesized_data}
We analyze how the number of synthesized data can influence point cloud completion results.
Table~\ref{tab:ablation} presents the precision, coverage, and CD values when $N_s$ is set to $1$, $4$, and $8$. 
According to the results, we achieve the best performance by $N_s=8$ on average among other setups, while there is no dramatic difference as it shows only $0.03$, $0.01$, and $0.03$ difference compared to having one number of synthesized data.
Consequently, having more synthesized data slightly enhances the performance. %
\subsubsection{Training on multi-class}
\label{subsubsec:multi-class}
In our main experiment, we train our model on a specific class and present the results in section~\ref{subsec:Eval_shapeNet}.
However, in real-world scenarios where the classes of objects are not identified, it is necessary to train the network on multi-class objects.
Table~\ref{tab:ablation} shows the quantitative results of the ShapeNet dataset when trained with multi-class objects.
The results demonstrate not much difference in the performance as the precision and CD values were $0.10$ and $0.07$ worse, while the coverage was $0.03$ better in the multi-class training. 
Thus, we believe that our network can learn to understand the appropriate features of a particular object even when there are various classes in the training set.  
\subsubsection{Single-view training}
\label{subsubsec:single-view}
As mentioned in section~\ref{related_works}, recent works~\cite{WeakPCN, PointPnCNet, DPC} leverage multi-partial views for supervision.
Even though our method does not take the advantage of multi-view supervision, we validate the power of our method to be trained on only a single view per object.
As the training set includes multiple partial views for the same object, we take out these views and leave only one partial view to prove that it does not significantly affect our method's performance.
According to Table~\ref{tab:ablation}, our method shows only $0.01$ CD difference with the model trained with only the single partial view included in the training set. 
Through the results, we confirm that our method can still perform as expected even without multi-views of an object in the training set.

\section{Conclusion}
\label{sec:Conclusion}
In this paper, we propose ACL-SPC, the first self-supervised point cloud completion method from only a single input partial point cloud.
Our method learns to complete partial point clouds by adaptively controlling the output in a closed-loop system.
We also introduce a consistency loss to generate the same complete point cloud and learn the geometric features of the object.
Our extensive experiments demonstrate that our method can be more useful in real-world scenarios without performance degradation than other methods.
In most cases, our method shows better performance in the coverage than precision showing the excellent performance of filling in the missing parts.

\textbf{Limitations and future works.}
One remaining limitation of our method is that there is no constraint to not generate redundant points, which results in high precision values.
To improve the precision and reduce noise, we will apply more constraints for future works. 
We will further find applications of our self-supervised framework in other point cloud restoration tasks such as denoising and upsampling. 

\noindent
\textbf{Acknowledgement.}
This work was supported in part by the IITP grants [No.2021-0-01343, Artificial Intelligence Graduate School Program (Seoul National University), No.2022-0-00156, No. 2021-0-02068, and No.2022-0-00156], and the NRF grant [No. 2021M3A9E4080782] funded by the Korea government (MSIT).
\clearpage
\small
\bibliographystyle{ieee_fullname}
\bibliography{egbib}


\end{document}


\title{Supplementary Material \textit{for} \\ ACL-SPC: Adaptive Closed-Loop system \textit{for} \\ Self-Supervised Point Cloud Completion}

\author{Sangmin Hong$^{1}$\thanks{equal contribution} \qquad Mohsen Yavartanoo$^{2}$\samethanks \qquad Reyhaneh Neshatavar$^{2}$ \qquad Kyoung Mu Lee$^{1,2}$ \\$^{1}$IPAI,  $^{2}$Dept. of ECE \& ASRI, Seoul National University, Seoul, Korea\\
{\tt\small \{mchiash2,myavartanoo,reyhanehneshat,kyoungmu\}@snu.ac.kr}}

\maketitle

\renewcommand{\thesection}{S1}
\section{Ablation on network architecture}
We use PolyNet~[\textcolor{blue}{47}] as the encoder is that it is an advanced GraphCNN that can learn local features efficiently.
To justify the design choice of our proposed architecture, we have tested our framework by changing both the encoder and decoder modules. 
We compare the chamfer distance, per epoch training time, and per sample testing time when converting the encoder, decoder, or both to PCN~[\textcolor{blue}{49}] or TopNet~[\textcolor{blue}{37}]. 
The results are shown in Table~\ref{tab:Network_Arch}.
The results demonstrate that our method can leverage PolyNet to improve performance.
We further show that using our shallower decoder $\mathcal{D}$ can save the test time while obtaining slightly better performance.
We note that even with a much simpler encoder~(PCN) our method is still able to achieve relatively satisfactory performance.
\begin{table}[!h]
    \scriptsize
    \centering
    \setlength\tabcolsep{0.00001pt}
    \begin{tabularx}{\linewidth}{l 
    >{\centering\arraybackslash}X 
    >{\centering\arraybackslash}X 
    >{\centering\arraybackslash}X 
    >{\centering\arraybackslash}X 
    >{\centering\arraybackslash}X 
    }
    \toprule
    \textbf{Network Architecture} &
    \textbf{CD$\downarrow$} & 
    \multicolumn{2}{c}{\textbf{Train time (s)$\downarrow$}}~~ &
    \multicolumn{2}{c}{\textbf{Test time (ms)$\downarrow$}} \\
    \midrule
    TopNet~[\textcolor{blue}{37}] & {3.21} & \multicolumn{2}{c}{229.6} & \multicolumn{2}{c}{61.5} \\
    PCN~[\textcolor{blue}{49}] & {2.99} & \multicolumn{2}{c}{245.1} & \multicolumn{2}{c}{40.2} \\
    PCN encoder~[\textcolor{blue}{49}] + $\mathcal{D}$ & {3.07} & \multicolumn{2}{c}{\textbf{118.1}} & \multicolumn{2}{c}{39.9}    \\ 
    $\mathcal{E}$(PolyNet[\textcolor{blue}{47}]) + PCN decoder~[\textcolor{blue}{49}] & {2.93} & \multicolumn{2}{c}{314.3} & \multicolumn{2}{c}{42.3} \\
    \midrule
    \textbf{Ours} ($\mathcal{E}$(PolyNet[\textcolor{blue}{47}]) + $\mathcal{D}$)  & \textbf{2.88} & \multicolumn{2}{c}{179.8} & \multicolumn{2}{c}{\textbf{38.5}} \\
    \bottomrule
    \end{tabularx}
    \caption{\textbf{Ablation studies for the choice of the network architecture.} The values of CD are multiplied by 100. The train time is measured per epoch and the test time is measured per sample.}  
    \label{tab:Network_Arch}  
    \vspace{-2mm}
\end{table}
\renewcommand{\thesection}{S2}
\section{Analysis of losses}
We use a global average pooling after the last PolyConv layer to eliminate the spatial dependency and obtain permutation-invariant representative features of an object.
Therefore, for different partial observations of the same object, the encoder must generate the same representative features.
Accordingly, the decoder can generate the complete point clouds $C_v$ with the same point orders for different observations. 
Moreover, in our $\mathcal{L}^{\text{cons}}$ loss, instead of using the original or GT point clouds which have different orders from the generated complete point clouds $C_v$, we reduce the MSE between the points of the generated complete point clouds $C_0$ and $C_v$ which have the same orders.
We further justify our claim by replacing our loss function $\mathcal{L}^{\text{cons}}$ with the Chamfer loss $\mathcal{L}^{\text{chamf}}$
between $C_0$ and $C_v$ and showing the effectiveness of $\mathcal{L}^{\text{cons}}$ as shown in Table~\ref{tab:Loss_analysis}.
\begin{table}[!h]
    \centering
    \small
    \setlength\tabcolsep{0.0001pt}
    \begin{tabularx}{\linewidth}{l 
    >{\centering\arraybackslash}X 
    >{\centering\arraybackslash}X 
    >{\centering\arraybackslash}X
    >{\centering\arraybackslash}X }
    \toprule
    \textbf{Loss function}&
    \textbf{P}$\downarrow$ & 
    \textbf{C}$\downarrow$ &
    \textbf{CD}$\downarrow$\\
    \midrule
    $\mathcal{L}^{\text{chamf}}$ & \textbf{1.69} & {2.53} & {4.22}\\
    $\mathcal{L}^{\text{cons}}$  & {1.70} & \textbf{1.18} & \textbf{2.88}\\
    \bottomrule
    \end{tabularx}
    \caption{
    \textbf{Evaluation on loss function.}
    P, C, and CD refers to precision, coverage, and chamfer distance respectively. All the avalues are multiplied by 100.}
    \label{tab:Loss_analysis}   
    \vspace{-6mm}
\end{table}

\renewcommand{\thesection}{S3}
\section{Qualitative results on Asymmetric Data}
\begin{figure}[!h]
     \centering
     \begin{subfigure}[b]{0.2\linewidth}
         \centering
         \includegraphics[page=1, width=\textwidth]{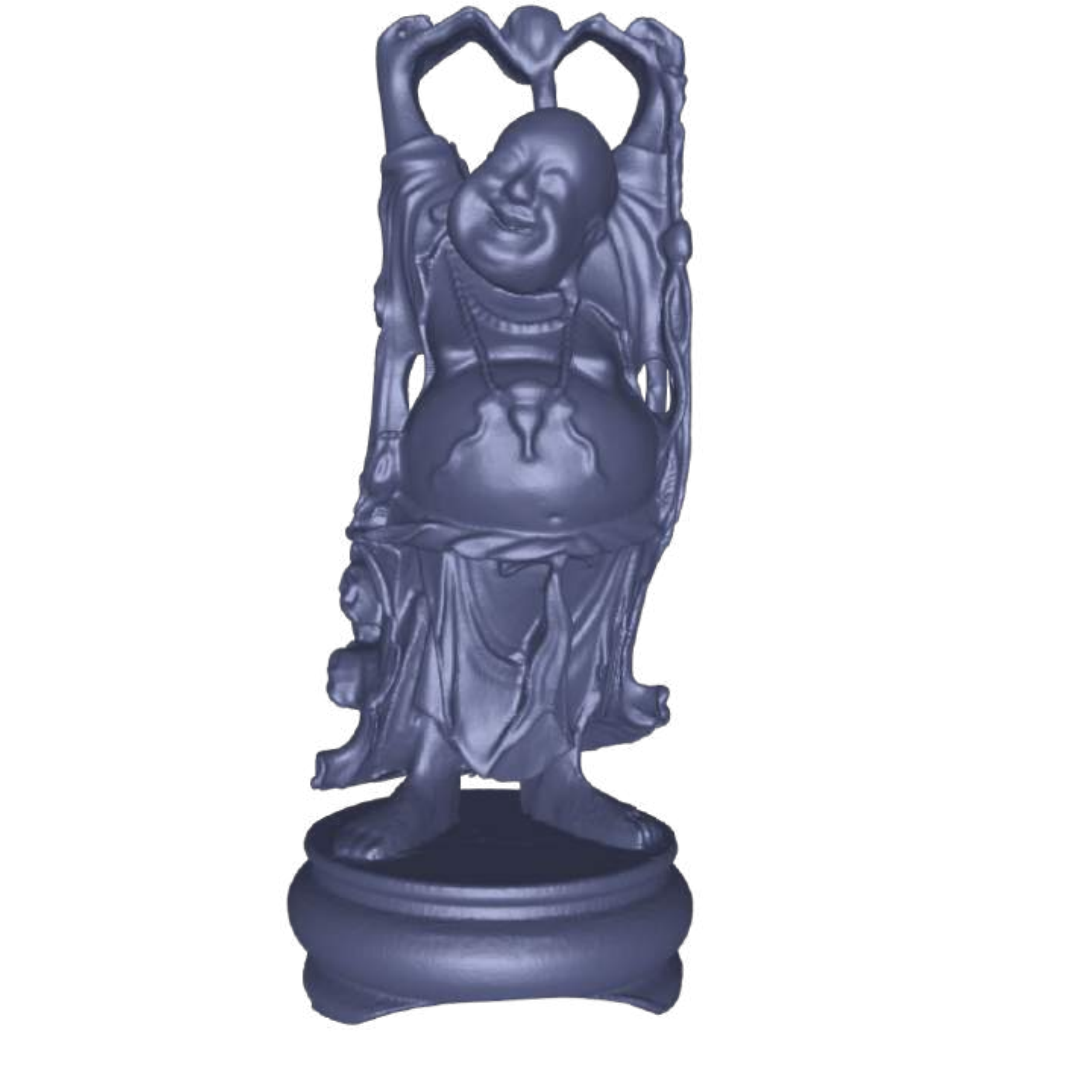}
         \caption*{Object}
     \end{subfigure}
     \hfill
     \begin{subfigure}[b]{0.2\linewidth}
         \centering
         \includegraphics[page=2, width=\textwidth]{figures/rebuttal/rebuttal_assymetric.pdf}
         \caption*{Input}
     \end{subfigure}
     \hfill
     \begin{subfigure}[b]{0.2\linewidth}
         \centering
         \includegraphics[page=3, width=\textwidth]{figures/rebuttal/rebuttal_assymetric.pdf}
         \caption*{Output}
     \end{subfigure}
     \hfill
     \begin{subfigure}[b]{0.2\linewidth}
         \centering
         \includegraphics[page=4,width=\textwidth]{figures/rebuttal/rebuttal_assymetric.pdf}
         \caption*{GT}
     \end{subfigure}
     \vspace{2mm}
     \\
     \begin{subfigure}[b]{0.2\linewidth}
         \centering
         \includegraphics[page=5, width=\textwidth]{figures/rebuttal/rebuttal_assymetric.pdf}
         \caption*{Object}
     \end{subfigure}
     \hfill
     \begin{subfigure}[b]{0.2\linewidth}
         \centering
         \includegraphics[page=6, width=\textwidth]{figures/rebuttal/rebuttal_assymetric.pdf}
         \caption*{Input}
     \end{subfigure}
     \hfill   
     \begin{subfigure}[b]{0.2\linewidth}
         \centering
         \includegraphics[page=7, width=\textwidth]{figures/rebuttal/rebuttal_assymetric.pdf}
         \caption*{Output}
     \end{subfigure}
     \hfill
     \begin{subfigure}[b]{0.2\linewidth}
         \centering
         \includegraphics[page=8, width=\textwidth]{figures/rebuttal/rebuttal_assymetric.pdf}
         \caption*{GT}
     \end{subfigure}
    \caption{
    \textbf{Qualitative results on Asymmetric Data.} 
    }        
\label{fig:asymmetric}    
\end{figure}

We extract different partial views from asymmetric objects and used them to train and evaluate our method. According to Figure~\ref{fig:asymmetric}, the results show that our method reconstructs the asymmetric shapes reliably as well as symmetric shapes. \\

\begin{table*}[t]
    \small
    \centering
    \setlength\tabcolsep{0.0001pt}
    \begin{tabularx}{\linewidth}{l >{\centering\arraybackslash}X >{\centering\arraybackslash}X >{\centering\arraybackslash}X  >{\centering\arraybackslash}X
    >{\centering\arraybackslash}X
    >{\centering\arraybackslash}X
    >{\centering\arraybackslash}X
    >{\centering\arraybackslash}X
    >{\centering\arraybackslash}X
    >{\centering\arraybackslash}X
    >{\centering\arraybackslash}X 
    >{\centering\arraybackslash}X 
    >{\centering\arraybackslash}X 
    >{\centering\arraybackslash}X 
    >{\centering\arraybackslash}X 
    >{\centering\arraybackslash}X 
    >{\centering\arraybackslash}X 
    >{\centering\arraybackslash}X 
    >{\centering\arraybackslash}X 
    >{\centering\arraybackslash}X 
    }

    \toprule
    \multirow{2}{*}{\textbf{Supervision}}&
    \multicolumn{2}{c}{\multirow{2}{*}{\textbf{Method}}}& 
    \multicolumn{2}{c}{\textbf{Plane}} & \multicolumn{2}{c}{\textbf{Cabinet}} &
    \multicolumn{2}{c}{\textbf{Car}} & \multicolumn{2}{c}{\textbf{Chair}} &
    \multicolumn{2}{c}{\textbf{Lamp}} & \multicolumn{2}{c}{\textbf{Sofa}} &
    \multicolumn{2}{c}{\textbf{Table}} & \multicolumn{2}{c}{\textbf{Vessel}} &
    \multicolumn{2}{c}{\textbf{Average}}\\
    \\
    & & & \textbf{P}$\downarrow$ & \textbf{C}$\downarrow$ & \textbf{P}$\downarrow$ & \textbf{C}$\downarrow$ & \textbf{P}$\downarrow$ & \textbf{C}$\downarrow$ & \textbf{P}$\downarrow$ & \textbf{C}$\downarrow$ & \textbf{P}$\downarrow$ & \textbf{C}$\downarrow$ & \textbf{P}$\downarrow$ & \textbf{C}$\downarrow$ & \textbf{P}$\downarrow$ & \textbf{C}$\downarrow$ &
    \textbf{P}$\downarrow$ & \textbf{C}$\downarrow$ & \textbf{P}$\downarrow$ & \textbf{C}$\downarrow$\\
    \midrule
    \multirow{3}{*}{Supervised}
    & \multicolumn{2}{c}{PCN}[\textcolor{blue}{49}] & {0.58} & {0.52} & {1.01} & {1.12} & {0.90} & \textbf{0.83} & {1.17} & {1.03} & {1.33} & {0.94} & {1.23} & {1.10} & {0.88} & {0.84} & {1.07} & {0.86} & {1.02} & {0.91}  \\
    & \multicolumn{2}{c}{GRNet}[\textcolor{blue}{43}] & {0.68} & {0.61} & {1.08} & \textbf{1.00} & {0.98} & {0.91} & {1.01} & {0.87} & {0.89} & {0.71} & {1.12} & {0.99} & {0.89} & {0.80} & {0.88} & {0.73} & {0.94} & {0.83}\\
    & \multicolumn{2}{c}{SFNet}[\textcolor{blue}{42}] & \textbf{0.46} & \textbf{0.39} & \textbf{0.85} & {1.01} & \textbf{0.80} & {0.84} & \textbf{0.72} & \textbf{0.82} & \textbf{0.55} & \textbf{0.64} & \textbf{0.90} & \textbf{0.96} & \textbf{0.61} & \textbf{0.68} & \textbf{0.60} & \textbf{0.67} & \textbf{0.69} & \textbf{0.75} \\
    \midrule
    Self-supervised & \multicolumn{2}{c}{\textbf{Ours}} & {1.81} & {0.58} & {2.80} & {1.28} & {1.96} & {0.98} & {3.43} & {1.10} & {5.31} & {0.98} & {3.08} & {1.26} & {3.17} & {1.05} & {3.40} & {0.82} & {3.12} & {1.01} \\
    \bottomrule
    \end{tabularx}

    \caption{
        \textbf{Evaluation and comparison on eight categories of PCN~[\textcolor{blue}{49}] dataset.}
        P and C refers to precision and coverage, respectively. 
        All the values are multiplied by 100.
        Compared to supervised methods, our result shows competitive results on coverage.
        }
    \label{tab:pcn}
\end{table*}

\renewcommand{\thesection}{S4}
\section{Evaluation on PCN}
PCN~[\textcolor{blue}{49}] dataset originates from ShapeNet~[\textcolor{blue}{5}] dataset where a set of $16,384$ points are uniformly sampled from the surface of each shape.
Partial point clouds are obtained by projecting the 3D point cloud to 2.5D data from eight random views and back projecting to 3D for each object.
Note that the points are resampled to a fixed number of $3,096$ points for each partial point cloud.  

As shown in Table~\ref{tab:pcn}, we train and evaluate our method on the PCN~[\textcolor{blue}{49}] dataset and compare it with the supervised methods.
%
%
Even though our self-supervised method ACL-SPC does not achieve the best performance among the supervised methods, it shows decent performance in the coverage, which demonstrates the ability of our ACL-SPC method to cover the missing parts without using the ground-truth.
However, since there is no such strict supervision as a complete point cloud, our method has higher precision, implying that it is noisier than the supervised methods. 

\begin{figure*}[h]
\vspace{4mm}
     \centering
     \begin{subfigure}[b]{0.14\textwidth}
         \centering
         \includegraphics[page=1, width=\textwidth]{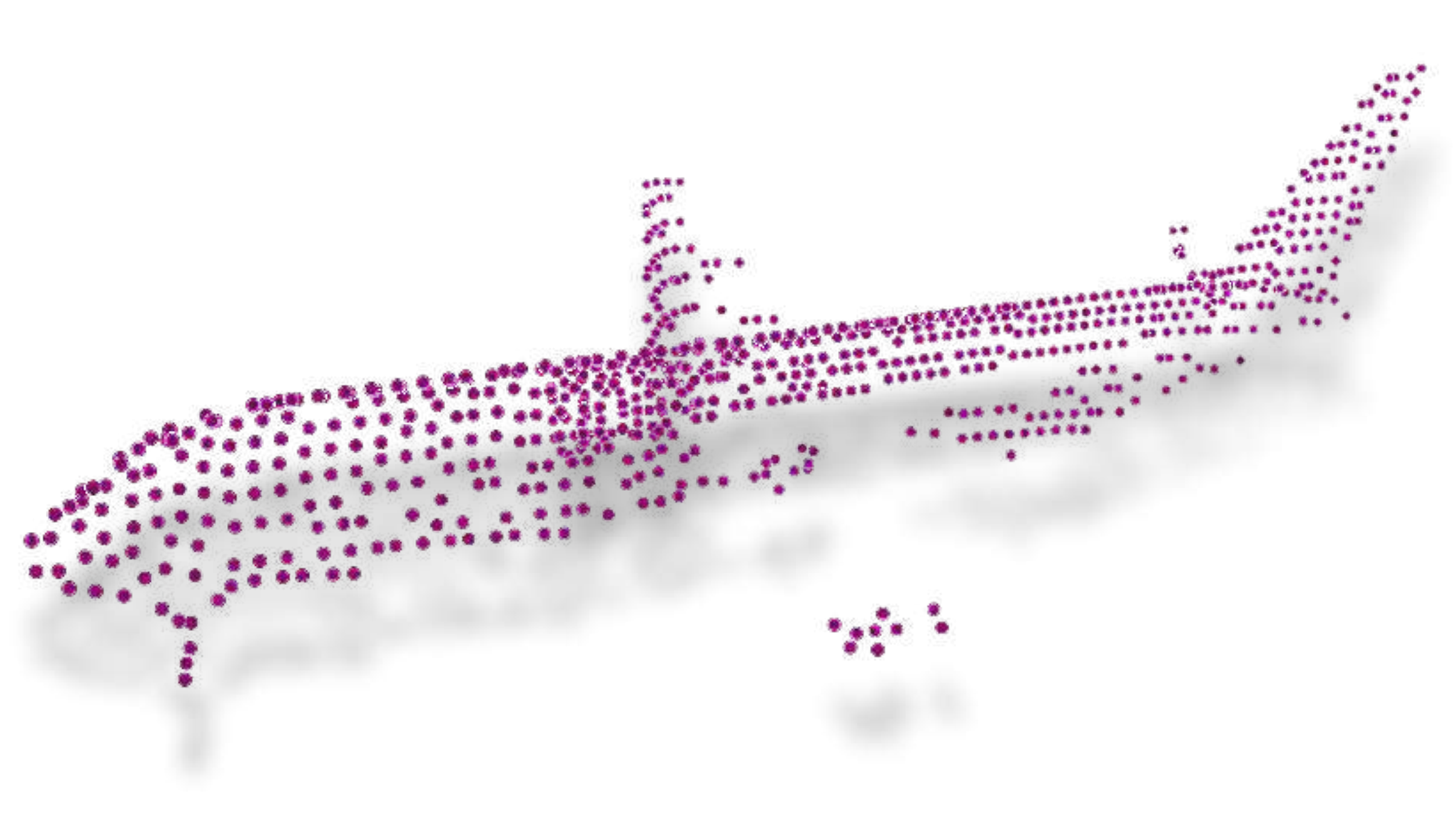}
     \end{subfigure}
     \hfill
     \begin{subfigure}[b]{0.14\textwidth}
         \centering
         \includegraphics[page=2, width=\textwidth]{figures/pcn_qual.pdf}
     \end{subfigure}
     \hfill
     \begin{subfigure}[b]{0.14\textwidth}
         \centering
         \includegraphics[page=3, width=\textwidth]{figures/pcn_qual.pdf}
     \end{subfigure}
     \hfill
     \begin{subfigure}[b]{0.14\textwidth}
     \caption*{\tiny \textbf{Supevised methods}}
         \centering
         \includegraphics[page=4,width=\textwidth]{figures/pcn_qual.pdf}
     \end{subfigure}
     \hfill
     \begin{subfigure}[b]{0.14\textwidth}
         \centering
         \includegraphics[page=5, width=\textwidth]{figures/pcn_qual.pdf}
     \end{subfigure}
     \hfill
     \begin{subfigure}[b]{0.14\textwidth}
     \caption*{\tiny \textbf{Self-supervised method}}
         \centering
         \includegraphics[page=6,width=\textwidth]{figures/pcn_qual.pdf}
     \end{subfigure}
     \vspace{4mm}
     \\
     \begin{subfigure}[b]{0.14\textwidth}
         \centering
         \includegraphics[page=7, width=\textwidth]{figures/pcn_qual.pdf}
     \end{subfigure}
     \hfill
     \begin{subfigure}[b]{0.14\textwidth}
         \centering
         \includegraphics[page=8, width=\textwidth]{figures/pcn_qual.pdf}
     \end{subfigure}
     \hfill   
     \begin{subfigure}[b]{0.14\textwidth}
         \centering
         \includegraphics[page=9, width=\textwidth]{figures/pcn_qual.pdf}
     \end{subfigure}
     \hfill
     \begin{subfigure}[b]{0.14\textwidth}
         \centering
         \includegraphics[page=10, width=\textwidth]{figures/pcn_qual.pdf}
     \end{subfigure}
     \hfill 
     \begin{subfigure}[b]{0.14\textwidth}
         \centering
         \includegraphics[page=11, width=\textwidth]{figures/pcn_qual.pdf}
     \end{subfigure}
     \hfill
     \begin{subfigure}[b]{0.14\textwidth}
         \centering
         \includegraphics[page=12, width=\textwidth]{figures/pcn_qual.pdf}
     \end{subfigure}
     \vspace{4mm}
     \\ 
     \begin{subfigure}[b]{0.14\textwidth}
         \centering
         \includegraphics[page=13, width=\textwidth]{figures/pcn_qual.pdf}
     \end{subfigure}
     \hfill
     \begin{subfigure}[b]{0.14\textwidth}
         \centering
         \includegraphics[page=14, width=\textwidth]{figures/pcn_qual.pdf}
     \end{subfigure}
     \hfill
     \begin{subfigure}[b]{0.14\textwidth}
         \centering
         \includegraphics[page=15, width=\textwidth]{figures/pcn_qual.pdf}
     \end{subfigure}
     \hfill
     \begin{subfigure}[b]{0.14\textwidth}
         \centering
         \includegraphics[page=16, width=\textwidth]{figures/pcn_qual.pdf}
     \end{subfigure}
     \hfill
     \begin{subfigure}[b]{0.14\textwidth}
         \centering
         \includegraphics[page=17, width=\textwidth]{figures/pcn_qual.pdf}
     \end{subfigure}   
     \hfill
     \begin{subfigure}[b]{0.14\textwidth}
         \centering
         \includegraphics[page=18, width=\textwidth]{figures/pcn_qual.pdf}
     \end{subfigure}
     \vspace{4mm}
     \\
     \begin{subfigure}[b]{0.14\textwidth}
         \centering
         \includegraphics[page=19, width=\textwidth]{figures/pcn_qual.pdf}
     \end{subfigure}
     \hfill
     \begin{subfigure}[b]{0.14\textwidth}
         \centering
         \includegraphics[page=20, width=\textwidth]{figures/pcn_qual.pdf}
     \end{subfigure}
     \hfill
     \begin{subfigure}[b]{0.14\textwidth}
         \centering
         \includegraphics[page=21, width=\textwidth]{figures/pcn_qual.pdf}
     \end{subfigure}
     \hfill
     \begin{subfigure}[b]{0.14\textwidth}
         \centering
         \includegraphics[page=22, width=\textwidth]{figures/pcn_qual.pdf}
     \end{subfigure}
     \hfill
     \begin{subfigure}[b]{0.14\textwidth}
         \centering
         \includegraphics[page=23, width=\textwidth]{figures/pcn_qual.pdf}
     \end{subfigure}   
     \hfill
     \begin{subfigure}[b]{0.14\textwidth}
         \centering
         \includegraphics[page=24, width=\textwidth]{figures/pcn_qual.pdf}
     \end{subfigure}
     \vspace{4mm}
     \\
     \begin{subfigure}[b]{0.14\textwidth}
         \centering
         \includegraphics[page=25, width=\textwidth]{figures/pcn_qual.pdf}
     \end{subfigure}
     \hfill
     \begin{subfigure}[b]{0.14\textwidth}
         \centering
         \includegraphics[page=26, width=\textwidth]{figures/pcn_qual.pdf}
     \end{subfigure}
     \hfill
     \begin{subfigure}[b]{0.14\textwidth}
         \centering
         \includegraphics[page=27, width=\textwidth]{figures/pcn_qual.pdf}
     \end{subfigure}
     \hfill
     \begin{subfigure}[b]{0.14\textwidth}
         \centering
         \includegraphics[page=28, width=\textwidth]{figures/pcn_qual.pdf}
     \end{subfigure}
     \hfill
     \begin{subfigure}[b]{0.14\textwidth}
         \centering
         \includegraphics[page=29, width=\textwidth]{figures/pcn_qual.pdf}
     \end{subfigure}   
     \hfill
     \begin{subfigure}[b]{0.14\textwidth}
         \centering
         \includegraphics[page=30, width=\textwidth]{figures/pcn_qual.pdf}
     \end{subfigure}
     \vspace{4mm}
     \\
     \begin{subfigure}[b]{0.14\textwidth}
         \centering
         \includegraphics[page=31, width=\textwidth]{figures/pcn_qual.pdf}
     \end{subfigure}
     \hfill
     \begin{subfigure}[b]{0.14\textwidth}
         \centering
         \includegraphics[page=32, width=\textwidth]{figures/pcn_qual.pdf}
     \end{subfigure}
     \hfill
     \begin{subfigure}[b]{0.14\textwidth}
         \centering
         \includegraphics[page=33, width=\textwidth]{figures/pcn_qual.pdf}
     \end{subfigure}
     \hfill
     \begin{subfigure}[b]{0.14\textwidth}
         \centering
         \includegraphics[page=34, width=\textwidth]{figures/pcn_qual.pdf}
     \end{subfigure}
     \hfill
     \begin{subfigure}[b]{0.14\textwidth}
         \centering
         \includegraphics[page=35, width=\textwidth]{figures/pcn_qual.pdf}
     \end{subfigure}   
     \hfill
     \begin{subfigure}[b]{0.14\textwidth}
         \centering
         \includegraphics[page=36, width=\textwidth]{figures/pcn_qual.pdf}
     \end{subfigure}
     \vspace{4mm}
     \\
     \begin{subfigure}[b]{0.14\textwidth}
         \centering
         \includegraphics[page=37, width=\textwidth]{figures/pcn_qual.pdf}
     \end{subfigure}
     \hfill
     \begin{subfigure}[b]{0.14\textwidth}
         \centering
         \includegraphics[page=38, width=\textwidth]{figures/pcn_qual.pdf}
     \end{subfigure}
     \hfill
     \begin{subfigure}[b]{0.14\textwidth}
         \centering
         \includegraphics[page=39, width=\textwidth]{figures/pcn_qual.pdf}
     \end{subfigure}
     \hfill
     \begin{subfigure}[b]{0.14\textwidth}
         \centering
         \includegraphics[page=40, width=\textwidth]{figures/pcn_qual.pdf}
     \end{subfigure}
     \hfill
     \begin{subfigure}[b]{0.14\textwidth}
         \centering
         \includegraphics[page=41, width=\textwidth]{figures/pcn_qual.pdf}
     \end{subfigure}   
     \hfill
     \begin{subfigure}[b]{0.14\textwidth}
         \centering
         \includegraphics[page=42, width=\textwidth]{figures/pcn_qual.pdf}
     \end{subfigure}
     \vspace{4mm}
     \\
     \begin{subfigure}[b]{0.14\textwidth}
         \centering
         \includegraphics[page=43, width=\textwidth]{figures/pcn_qual.pdf}
         \caption*{Input}
     \end{subfigure}
     \hfill
     \begin{subfigure}[b]{0.14\textwidth}
         \centering
         \includegraphics[page=44, width=\textwidth]{figures/pcn_qual.pdf}
         \caption*{GT}
     \end{subfigure}
     \hfill
     \begin{subfigure}[b]{0.14\textwidth}
         \centering
         \includegraphics[page=45, width=\textwidth]{figures/pcn_qual.pdf}
         \caption*{PCN~[\textcolor{blue}{49}]}
     \end{subfigure}
     \hfill
     \begin{subfigure}[b]{0.14\textwidth}
         \centering
         \includegraphics[page=46, width=\textwidth]{figures/pcn_qual.pdf}
         \caption*{GRNet~[\textcolor{blue}{43}]}
     \end{subfigure}
     \hfill
     \begin{subfigure}[b]{0.14\textwidth}
         \centering
         \includegraphics[page=47, width=\textwidth]{figures/pcn_qual.pdf}
         \caption*{SFNet~[\textcolor{blue}{42}]}
     \end{subfigure}   
     \hfill
     \begin{subfigure}[b]{0.14\textwidth}
         \centering
         \includegraphics[page=48, width=\textwidth]{figures/pcn_qual.pdf}
         \caption*{\textbf{Ours}}
     \end{subfigure}
        \caption{
        \textbf{Qualitative comparison on the PCN~[\textcolor{blue}{49}] dataset.} 
        PCN~[\textcolor{blue}{49}], GRNet~[\textcolor{blue}{43}], and SFNet~[\textcolor{blue}{42}] are supervised methods, whereas our method is self-supervised and is trained without the use of ground-truth. 
        }
        \label{fig:supp_pcn}
\end{figure*}     

\begin{table*}[t]
    \small
    \centering
    \setlength\tabcolsep{0.0001pt}
    \begin{tabularx}{\linewidth}{l >{\centering\arraybackslash}X >{\centering\arraybackslash}X >{\centering\arraybackslash}X  >{\centering\arraybackslash}X
    >{\centering\arraybackslash}X
    >{\centering\arraybackslash}X
    >{\centering\arraybackslash}X
    >{\centering\arraybackslash}X
    >{\centering\arraybackslash}X
    >{\centering\arraybackslash}X
    >{\centering\arraybackslash}X 
    >{\centering\arraybackslash}X 
    >{\centering\arraybackslash}X 
    >{\centering\arraybackslash}X 
    >{\centering\arraybackslash}X 
    >{\centering\arraybackslash}X 
    >{\centering\arraybackslash}X 
    >{\centering\arraybackslash}X 
    >{\centering\arraybackslash}X 
    >{\centering\arraybackslash}X 
    >{\centering\arraybackslash}X 
    }
      
    \toprule
    \multirow{2}{*}{\textbf{Supervision}}&
    \multicolumn{2}{c}{\multirow{2}{*}{\textbf{Method}}}& 
    \multicolumn{2}{c}{\textbf{Bus}} & \multicolumn{2}{c}{\textbf{Bed}} &
    \multicolumn{2}{c}{\textbf{Bookshelf}} & \multicolumn{2}{c}{\textbf{Bench}} &
    \multicolumn{2}{c}{\textbf{Guitar}} & \multicolumn{2}{c}{\textbf{Motorbike}} &
    \multicolumn{2}{c}{\textbf{Skateboard}} & \multicolumn{2}{c}{\textbf{Pistol}} &
    \multicolumn{2}{c}{\textbf{Average}}
    \\
    & & & \textbf{P} & \textbf{C} & \textbf{P} & \textbf{C} & \textbf{P} & \textbf{C} & \textbf{P} & \textbf{C} & \textbf{P} & \textbf{C} & \textbf{P} & \textbf{C} & \textbf{P} & \textbf{C} &  \textbf{P} & \textbf{C} & \textbf{P} & \textbf{C}\\
    \midrule
    \multirow{3}{*}{Supervised}
    & \multicolumn{2}{c}{PCN~[\textcolor{blue}{49}]} & \textbf{0.96} & \textbf{0.93} & {2.68} & {1.65} & {1.75} & {1.21} & \textbf{1.32} & \textbf{0.88} & {1.20} & {0.88} & {1.74} & {1.21} & {1.57} & {0.83} & {1.50} & {1.34} & {1.59} & {1.12}  \\
    & \multicolumn{2}{c}{GRNet~[\textcolor{blue}{43}]} & {1.13} & {1.18} & \textbf{2.38} & {2.10} & \textbf{1.63} & {1.34} & {1.38} & {0.92} & {1.03} & {0.75} & {1.14} & {1.23} & {1.43} & {0.83} & {1.12} & {1.53} & \textbf{1.41} & {1.24} \\
    & \multicolumn{2}{c}{SFNet~[\textcolor{blue}{42}]} & {1.13} & {1.25} & {2.90} & {2.80} & {2.39} & {2.01} & {1.60} & {1.21} & \textbf{0.49} & \textbf{0.62} & \textbf{1.13} & {1.12} & \textbf{0.74} & {0.82} & \textbf{1.10} & {1.36} & {1.44} & {1.40} \\
    \midrule
    Self-supervised & \multicolumn{2}{c}{\textbf{Ours}} & {2.25} & {0.96} & {4.90} & \textbf{1.43} & {2.98} & \textbf{1.17} & {3.32} & {0.99} & {4.57} & {0.81} & {3.14} & \textbf{0.87} & {2.90} & \textbf{0.70} & {5.47} & \textbf{0.92} & {3.69} & \textbf{0.98} \\
    \bottomrule
    \end{tabularx}

    \caption{
        \textbf{Evaluation and comparison on eight unseen categories of pcn~[\textcolor{blue}{49}] dataset.}
        P and C refers to precision and coverage, respectively. 
        All the values are multiplied by 100.
        Compared to supervised methods, our result shows competitive results on coverage.
        }
    \label{tab:pcn_unseen}
    \vspace{2mm}
\end{table*}

We also qualitatively compare our results to the supervised methods PCN~[\textcolor{blue}{49}], GRNet~[\textcolor{blue}{43}], and SFNet~[\textcolor{blue}{42}] on the PCN~[\textcolor{blue}{49}] dataset in Figure~\ref{fig:supp_pcn}.
The results demonstrate that our method can fill the missing parts of the input  almost par with the supervised methods.
Furthermore, we quantitatively evaluate our model on unseen categories of PCN~[\textcolor{blue}{49}] dataset.
The results in Table~\ref{tab:pcn_unseen} illustrate that our method can achieve superior performance on unseen categories compared with the supervised methods in terms of coverage in average. 
We can conclude from the results that our method is applicable in real-world situations where there is no information about the object's category.  


\renewcommand{\thesection}{S5}
\section{Learning progress}
\begin{figure*}
\vspace{4mm}
     \centering
     \begin{subfigure}[b]{0.13\textwidth}
         \centering
         \includegraphics[page=1,width=\textwidth]{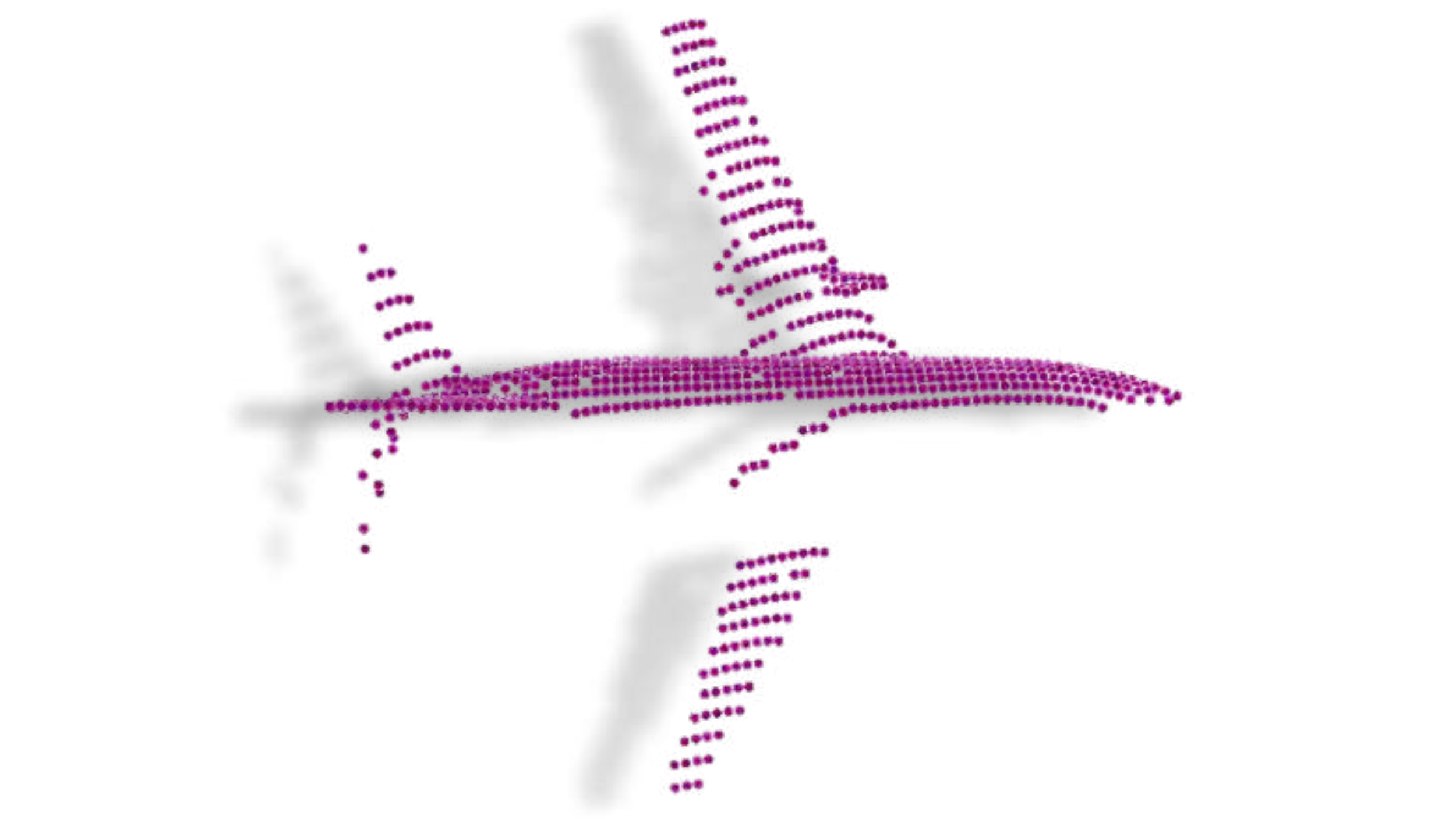}
     \end{subfigure}
     \hfill
     \begin{subfigure}[b]{0.13\textwidth}
         \centering
         \includegraphics[page=2,width=\textwidth]{figures/iterative.pdf}
     \end{subfigure}
     \hfill
     \begin{subfigure}[b]{0.13\textwidth}
         \centering
         \includegraphics[page=3,width=\textwidth]{figures/iterative.pdf}
     \end{subfigure}
     \hfill
     \begin{subfigure}[b]{0.13\textwidth}
         \centering
         \includegraphics[page=4,width=\textwidth]{figures/iterative.pdf}
     \end{subfigure}
     \hfill
     \begin{subfigure}[b]{0.13\textwidth}
         \centering
         \includegraphics[page=5,width=\textwidth]{figures/iterative.pdf}
     \end{subfigure}
     \hfill
     \begin{subfigure}[b]{0.13\textwidth}
         \centering
         \includegraphics[page=6,width=\textwidth]{figures/iterative.pdf}
     \end{subfigure}
     \hfill
     \begin{subfigure}[b]{0.13\textwidth}
         \centering
         \includegraphics[page=7,width=\textwidth]{figures/iterative.pdf}
     \end{subfigure}
     \vspace{4mm}
     \\
     \begin{subfigure}[b]{0.13\textwidth}
         \centering
         \includegraphics[page=8,width=\textwidth]{figures/iterative.pdf}
     \end{subfigure}
     \hfill
     \begin{subfigure}[b]{0.13\textwidth}
         \centering
         \includegraphics[page=9,width=\textwidth]{figures/iterative.pdf}
     \end{subfigure}
     \hfill   
     \begin{subfigure}[b]{0.13\textwidth}
         \centering
         \includegraphics[page=10,width=\textwidth]{figures/iterative.pdf}
     \end{subfigure}
     \hfill
     \begin{subfigure}[b]{0.13\textwidth}
         \centering
         \includegraphics[page=11,width=\textwidth]{figures/iterative.pdf}
     \end{subfigure}
     \hfill 
     \begin{subfigure}[b]{0.13\textwidth}
         \centering
         \includegraphics[page=12,width=\textwidth]{figures/iterative.pdf}
     \end{subfigure}
     \hfill
     \begin{subfigure}[b]{0.13\textwidth}
         \centering
         \includegraphics[page=13,width=\textwidth]{figures/iterative.pdf}
     \end{subfigure}
     \hfill
     \begin{subfigure}[b]{0.13\textwidth}
         \centering
         \includegraphics[page=14,width=\textwidth]{figures/iterative.pdf}
     \end{subfigure}
     \vspace{4mm}     
     \\
     \begin{subfigure}[b]{0.13\textwidth}
         \centering
         \includegraphics[page=15,width=\textwidth]{figures/iterative.pdf}
         \caption*{Input}
     \end{subfigure}
     \hfill
     \begin{subfigure}[b]{0.13\textwidth}
         \centering
         \includegraphics[page=16,width=\textwidth]{figures/iterative.pdf}
         \caption*{Epoch 1}
     \end{subfigure}
     \hfill
     \begin{subfigure}[b]{0.13\textwidth}
         \centering
         \includegraphics[page=17,width=\textwidth]{figures/iterative.pdf}
         \caption*{Epoch 10}
     \end{subfigure}
     \hfill
     \begin{subfigure}[b]{0.13\textwidth}
         \centering
         \includegraphics[page=18,width=\textwidth]{figures/iterative.pdf}
         \caption*{Epoch 100}
     \end{subfigure}
     \hfill
     \begin{subfigure}[b]{0.13\textwidth}
         \centering
         \includegraphics[page=19,width=\textwidth]{figures/iterative.pdf}
         \caption*{Epoch 300}
     \end{subfigure}   
     \hfill
     \begin{subfigure}[b]{0.13\textwidth}
         \centering
         \includegraphics[page=20,width=\textwidth]{figures/iterative.pdf}
         \caption*{Epoch best}
     \end{subfigure}
     \hfill
     \begin{subfigure}[b]{0.13\textwidth}
         \centering
         \includegraphics[page=21,width=\textwidth]{figures/iterative.pdf}
         \caption*{GT}
     \end{subfigure}
        \caption{
        \textbf{Qualitative improvement by iteration.} 
        }
        \label{fig:supp_iterative}
\end{figure*}     

We visualize the learning progress of our ACL-SPC framework by the output results from different epochs in Figure~\ref{fig:supp_iterative}.
We visualize the results at the epoch of 1, 10, 100, 300, and that when it has the best chamfer distance result.
The results demonstrate how our method can gradually learn to fill (epoch best) a given partial point cloud starting from a point cloud with naive shape (epoch 1).

\renewcommand{\thesection}{S6}
\section{Robustness to Viewpoint}
\begin{figure*}[t]
	\captionsetup[subfloat]{labelformat=empty}
		\newcommand{\rowArg}{0.6cm}
		\newcommand{\fullSize}{2.cm}
		\newcommand{\patchSize}{2.6cm}
		\setlength{\tabcolsep}{0.05cm}		
            \begin{tabular}[b]{c c c}
			\multicolumn{2}{c}{\multirow{2}{*}[\rowArg]{
					\subfloat[GT]
					{\includegraphics[page=1, height=\fullSize]
					{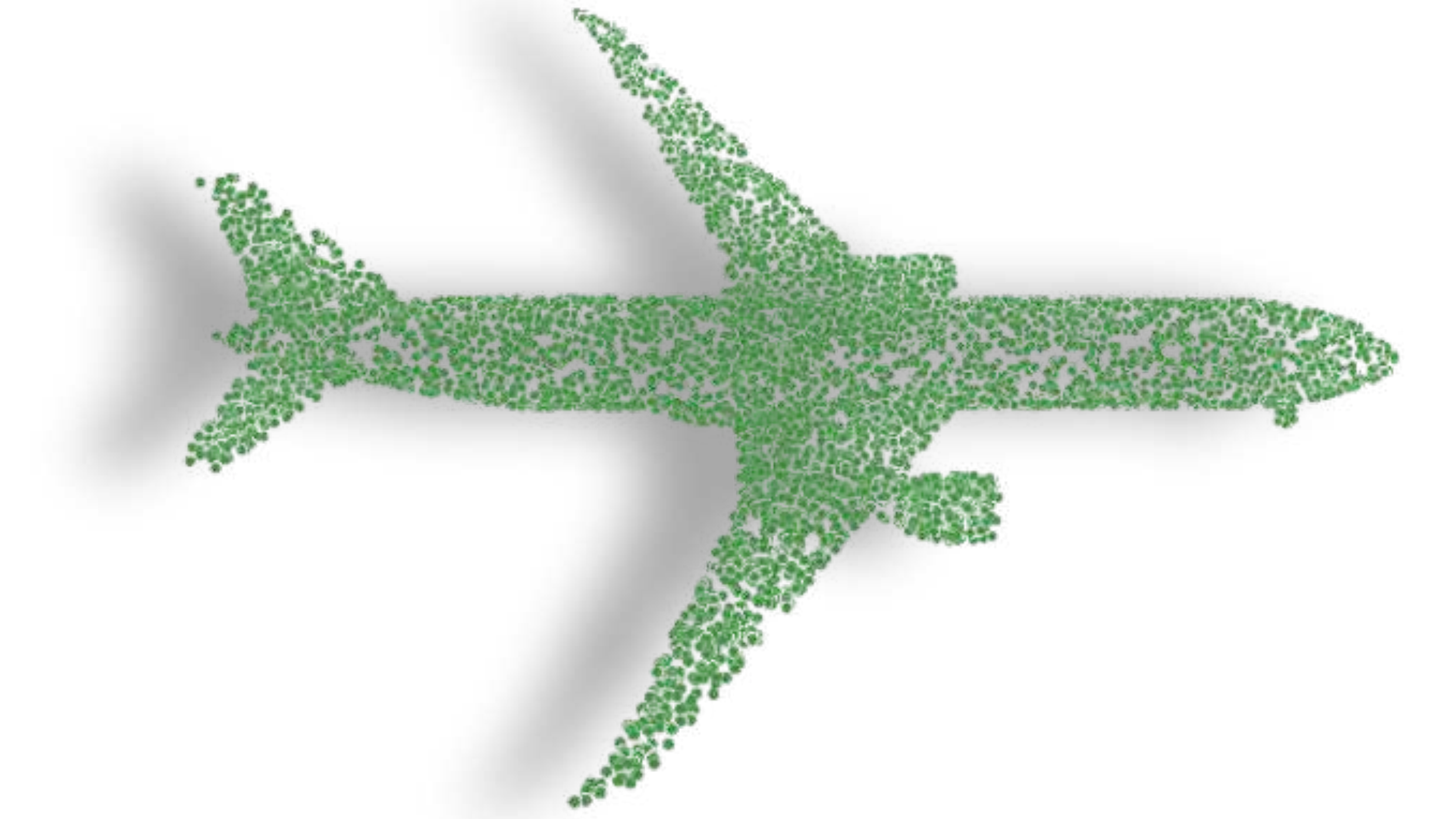}}}} &
			\subfloat[\centering \tiny Partial View 1 ]{
				\includegraphics[page=2, width = \patchSize]
				{figures/supp_view.pdf}} 
			\subfloat[\centering \tiny Partial View 2  ]{
				\includegraphics[page=3, width = \patchSize]
				{figures/supp_view.pdf}} 
			\subfloat[\centering \tiny Partial View 3  ]{
				\includegraphics[page=4, width = \patchSize]
				{figures/supp_view.pdf}} 
			\subfloat[\centering \tiny Partial View 4  ]{
				\includegraphics[page=5, width = \patchSize]
				{figures/supp_view.pdf}}
			\subfloat[\centering \tiny Partial View 5  ]{
				\includegraphics[page=6, width = \patchSize]
				{figures/supp_view.pdf}} \\  & &

			\subfloat[\centering  \tiny Complete 1  ]{
				\includegraphics[page=7, width = \patchSize]
				{figures/supp_view.pdf}}
			\subfloat[\centering \tiny Complete 2 ]{
				\includegraphics[page=8, width = \patchSize]
				{figures/supp_view.pdf}}
			\subfloat[\centering \tiny Complete 3 ]{
				\includegraphics[page=9, width = \patchSize]
				{figures/supp_view.pdf}}
			\subfloat[\centering \tiny Complete 4]{
				\includegraphics[page=10, width = \patchSize]
				{figures/supp_view.pdf}}
			\subfloat[\centering \tiny Complete 5]{
				\includegraphics[page=11, width = \patchSize]
				{figures/supp_view.pdf}}
	    \end{tabular}
	    
\vspace{4mm}	    
\end{figure*}
\begin{figure*}[ht]
	\captionsetup[subfloat]{labelformat=empty}
		\newcommand{\rowArg}{0.6cm}
		\newcommand{\fullSize}{2.cm}
		\newcommand{\patchSize}{2.6cm}
		\setlength{\tabcolsep}{0.05cm}		
            \begin{tabular}[b]{c c c}
			\multicolumn{2}{c}{\multirow{2}{*}[\rowArg]{
					\subfloat[GT]
					{\includegraphics[page=12, height=\fullSize]
					{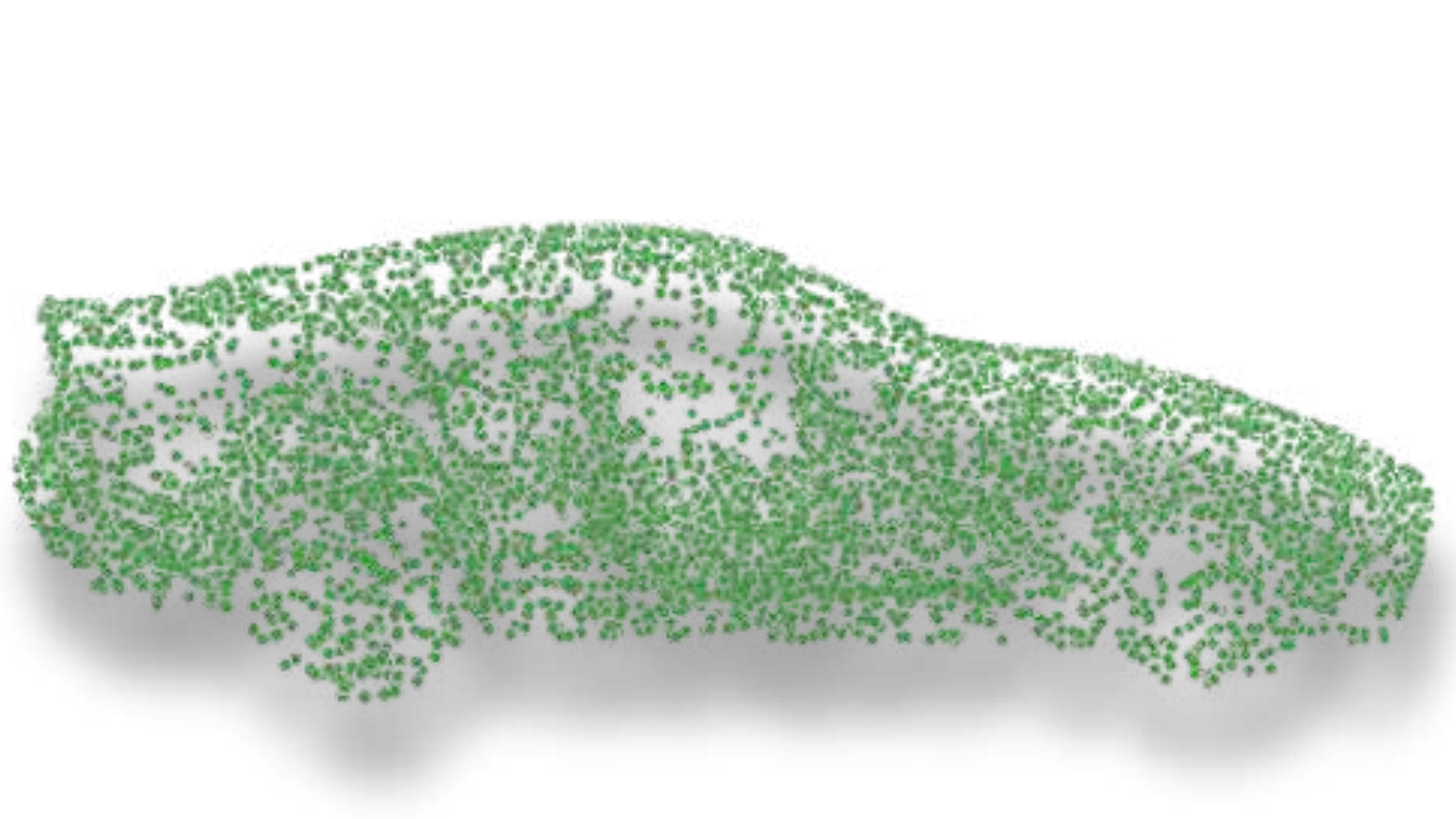}}}} &
			\subfloat[\centering \tiny Partial View 1 ]{
				\includegraphics[page=13, width = \patchSize]
				{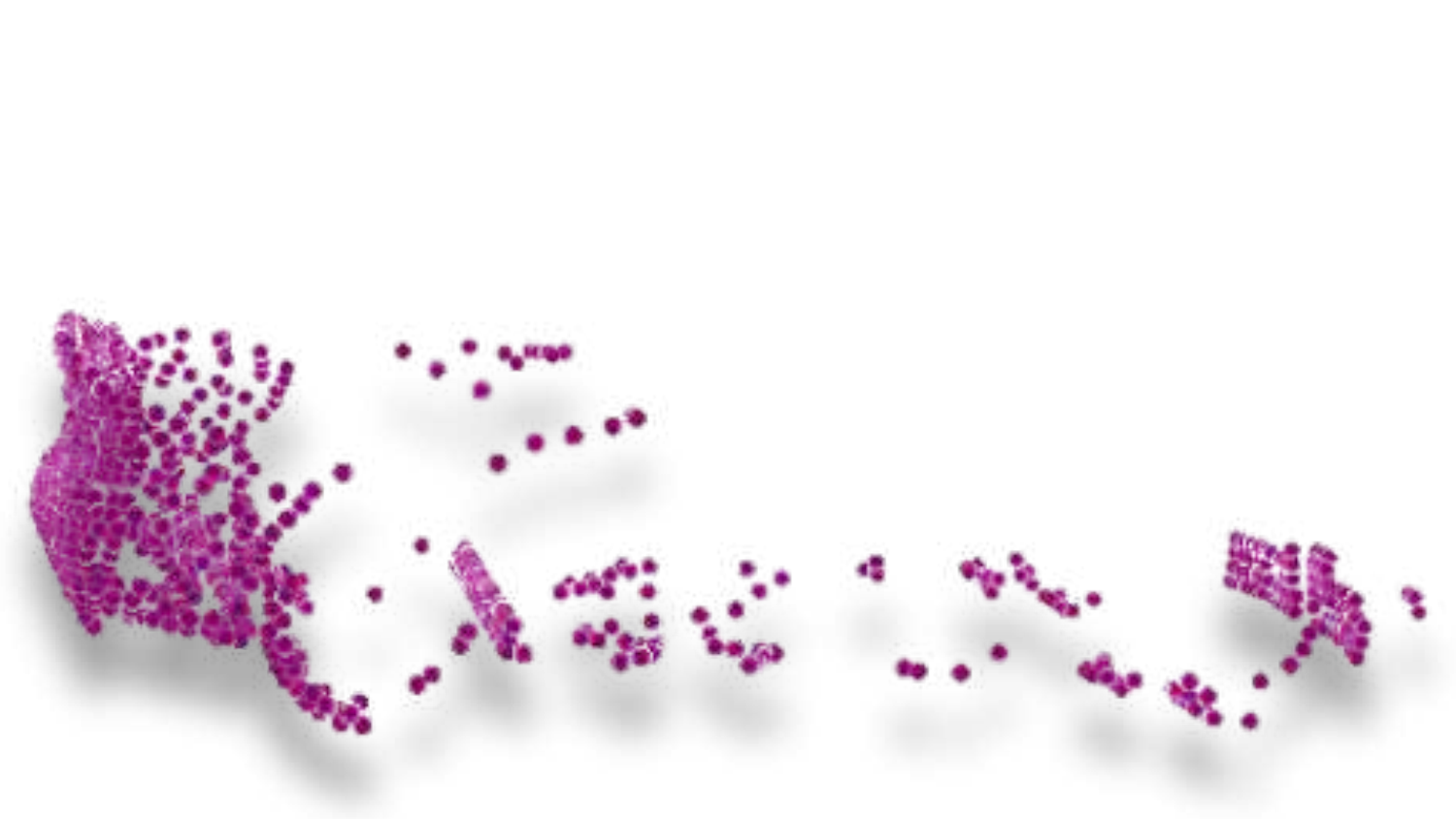}} 
			\subfloat[\centering \tiny Partial View 2  ]{
				\includegraphics[page=14, width = \patchSize]
				{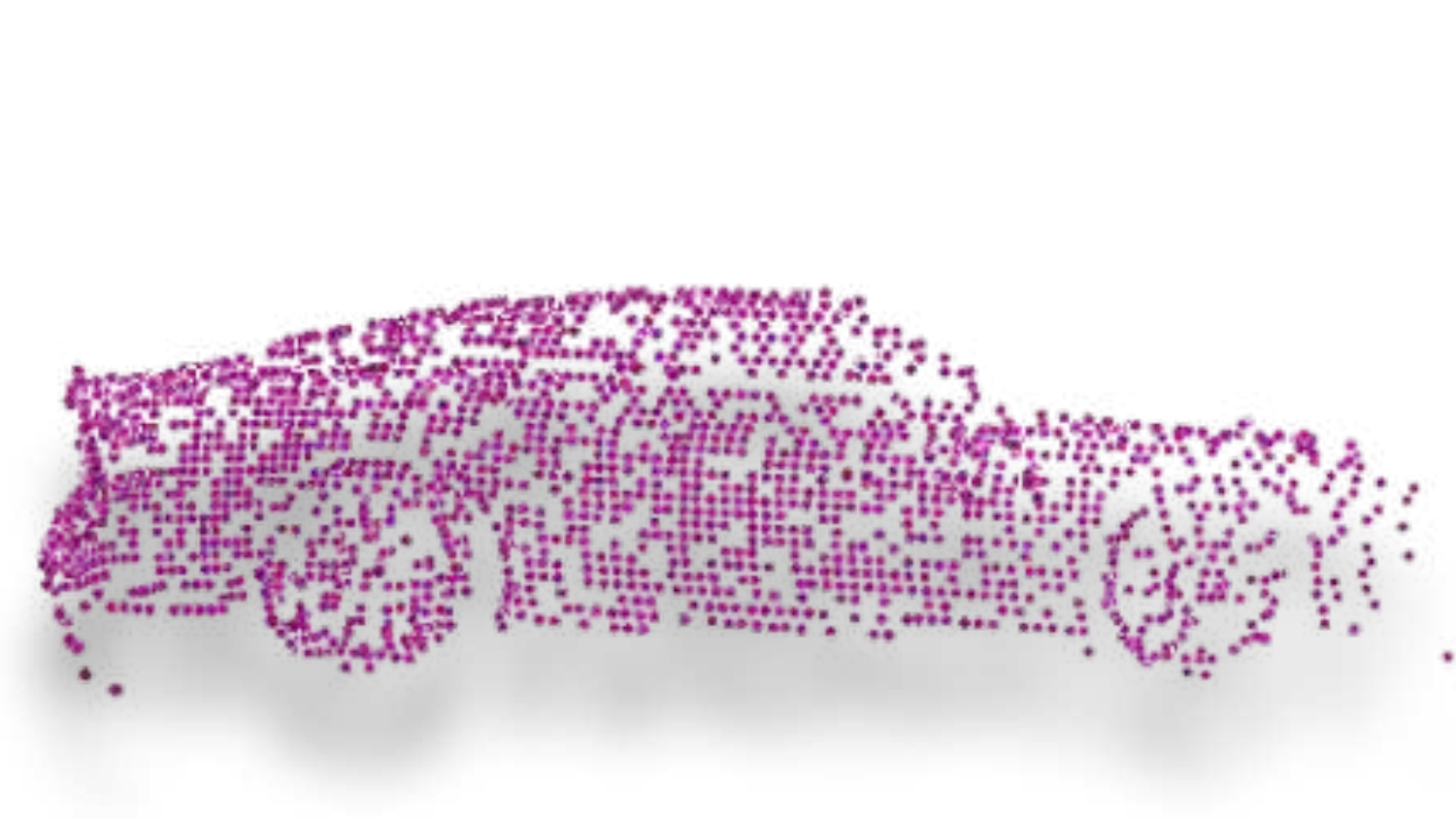}} 
			\subfloat[\centering \tiny Partial View 3  ]{
				\includegraphics[page=15, width = \patchSize]
				{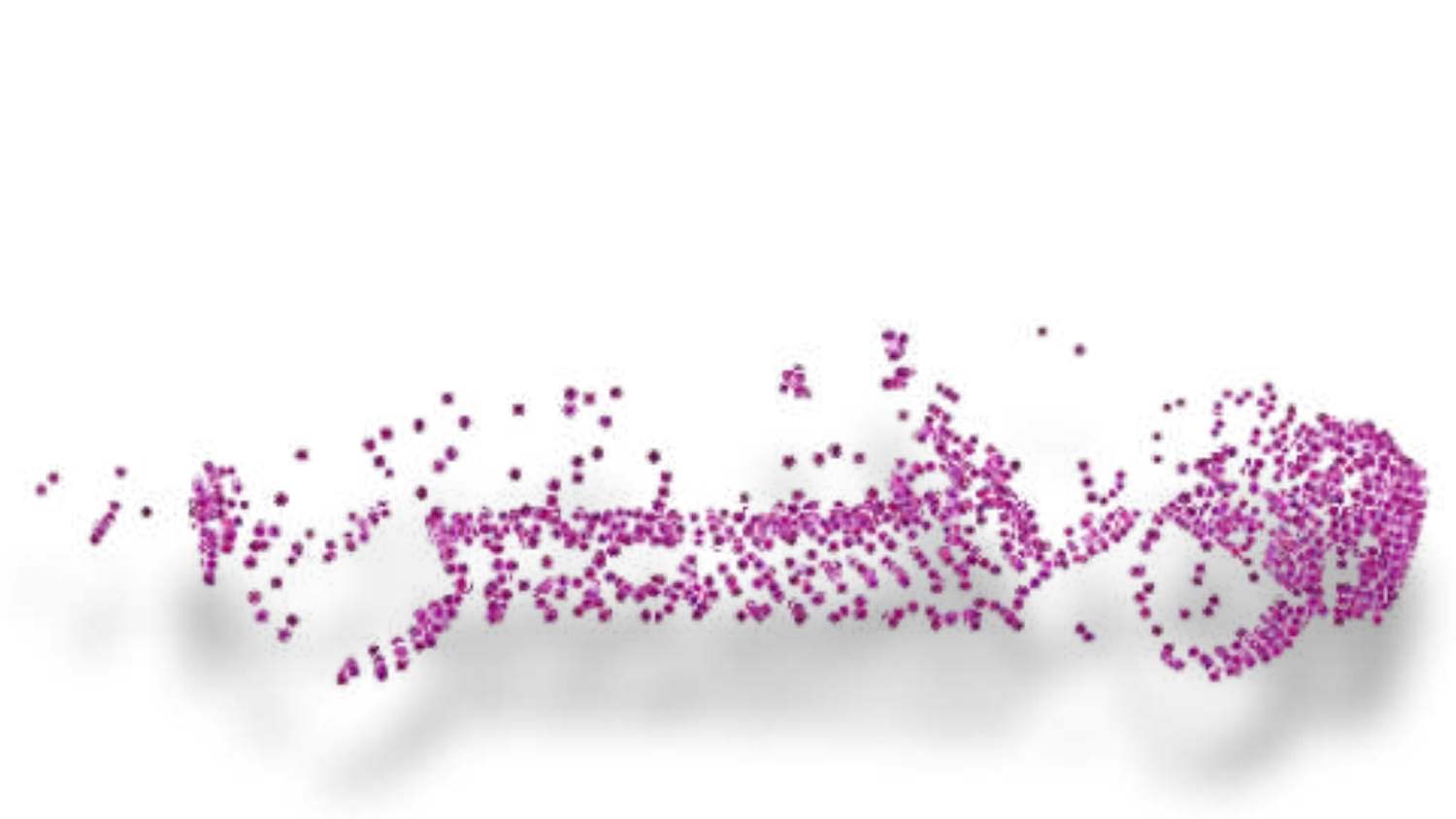}} 
			\subfloat[\centering \tiny Partial View 4  ]{
				\includegraphics[page=16, width = \patchSize]
				{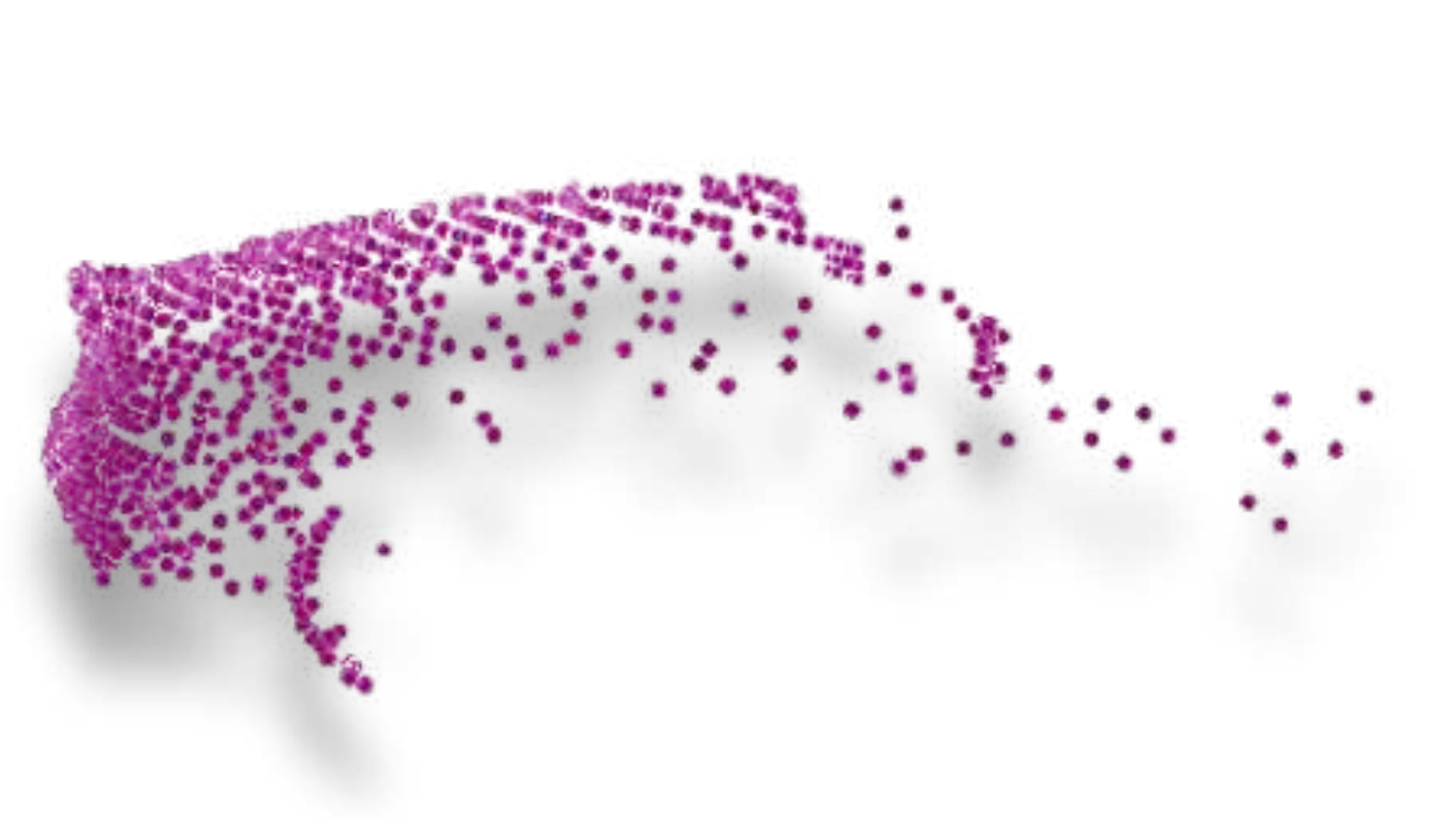}}
			\subfloat[\centering \tiny Partial View 5  ]{
				\includegraphics[page=17, width = \patchSize]
				{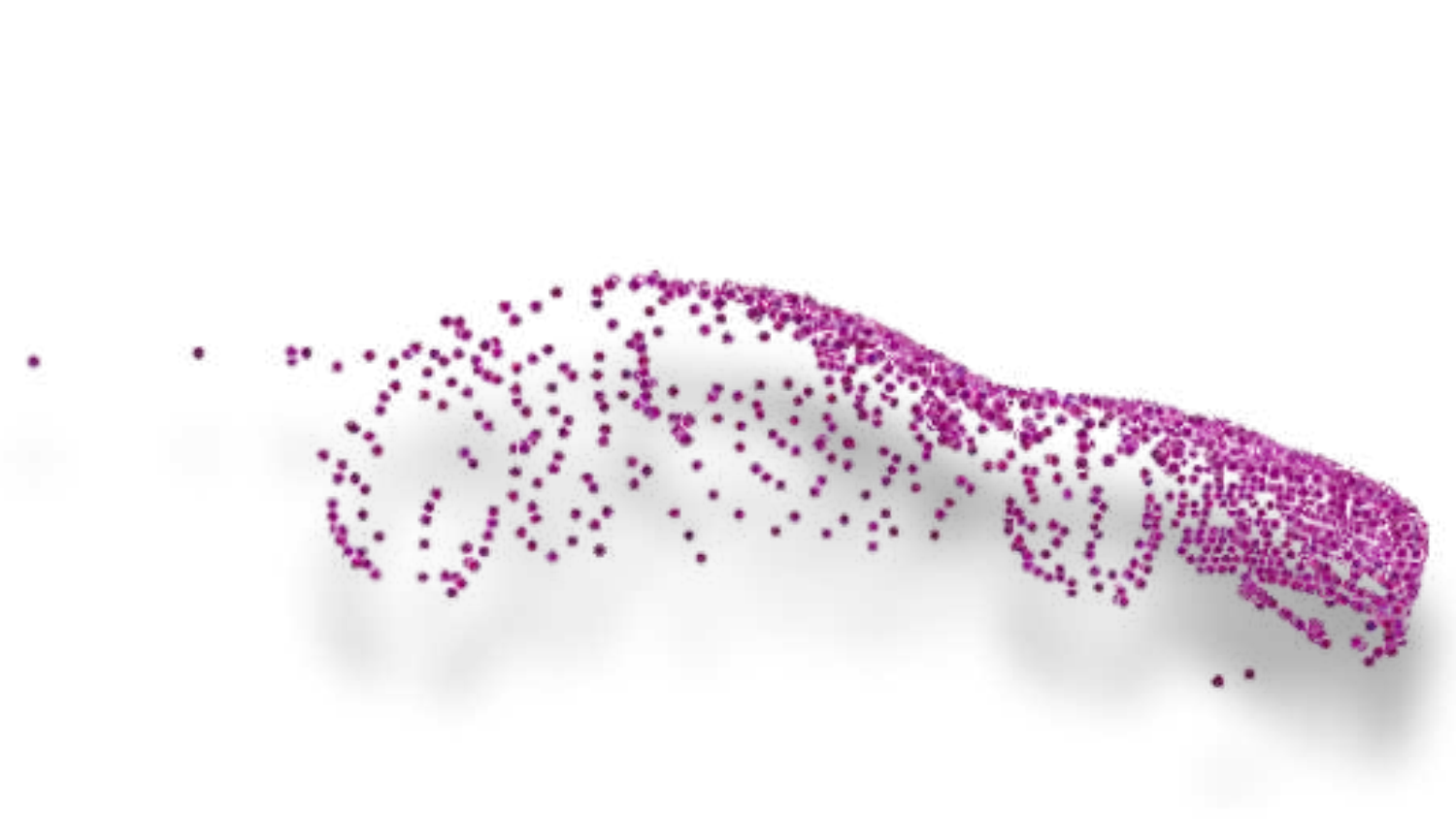}} \\  & &

			\subfloat[\centering  \tiny Complete 1  ]{
				\includegraphics[page=18, width = \patchSize]
				{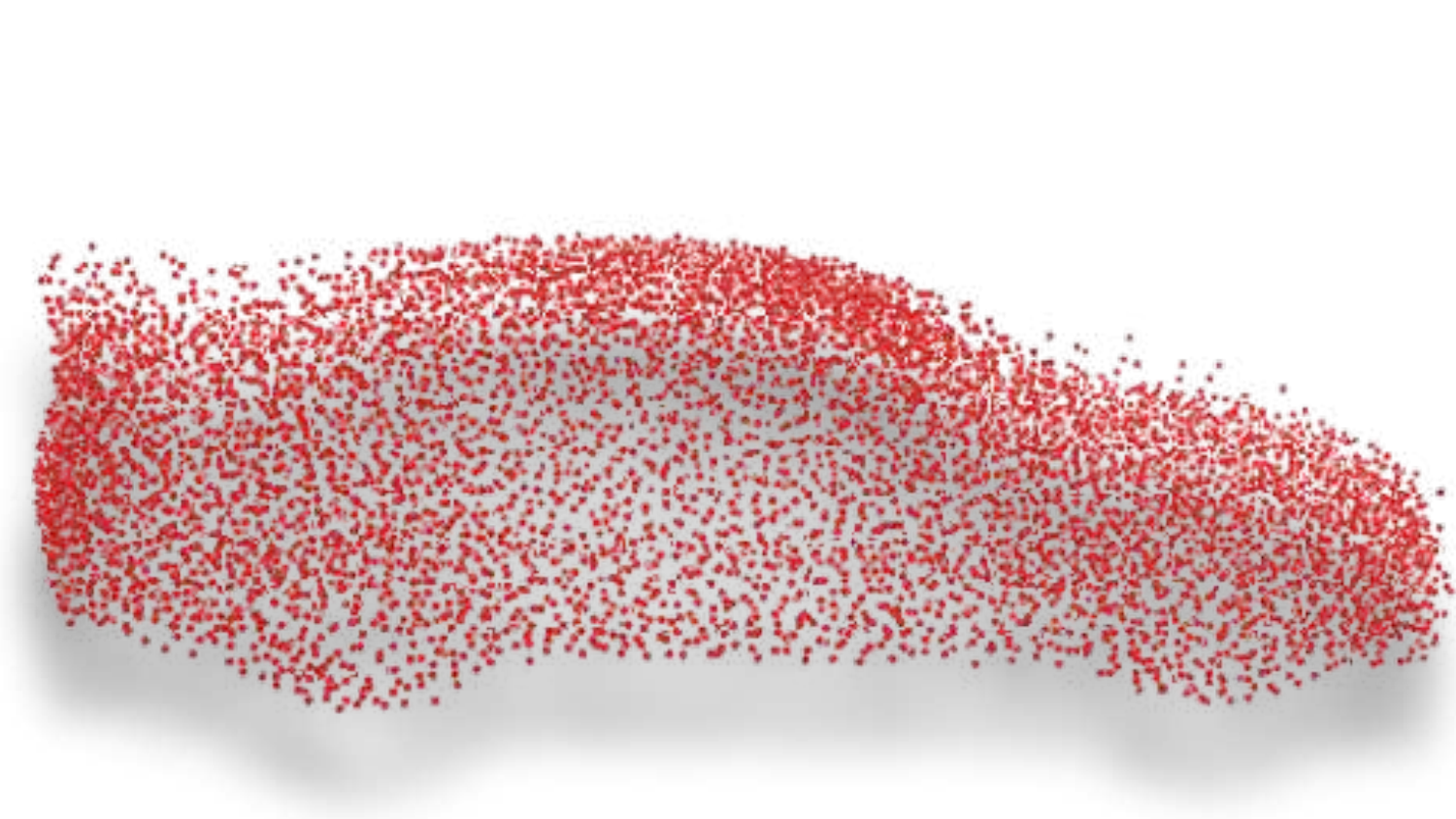}}
			\subfloat[\centering \tiny Complete 2 ]{
				\includegraphics[page=19, width = \patchSize]
				{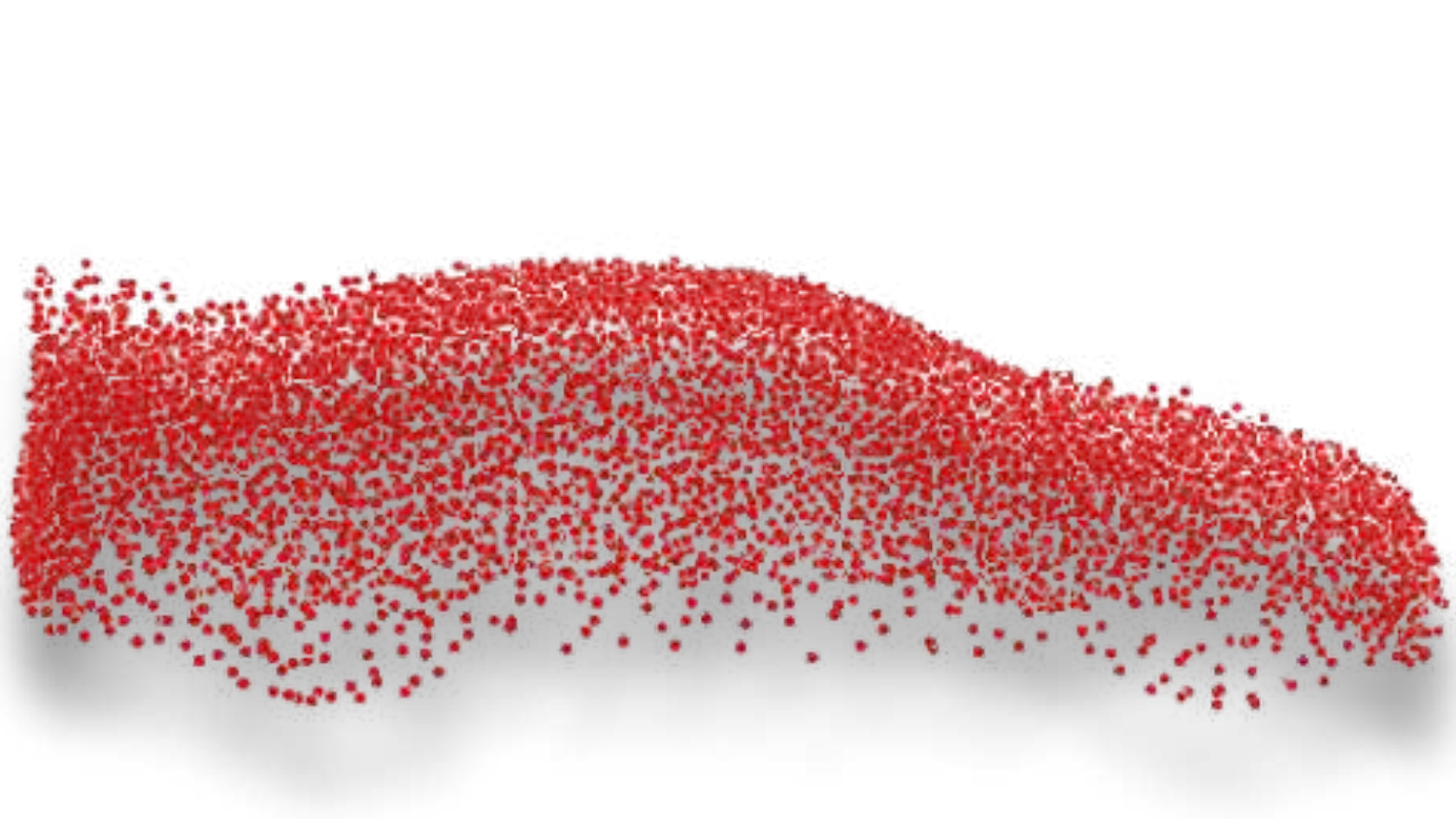}}
			\subfloat[\centering \tiny Complete 3 ]{
				\includegraphics[page=20, width = \patchSize]
				{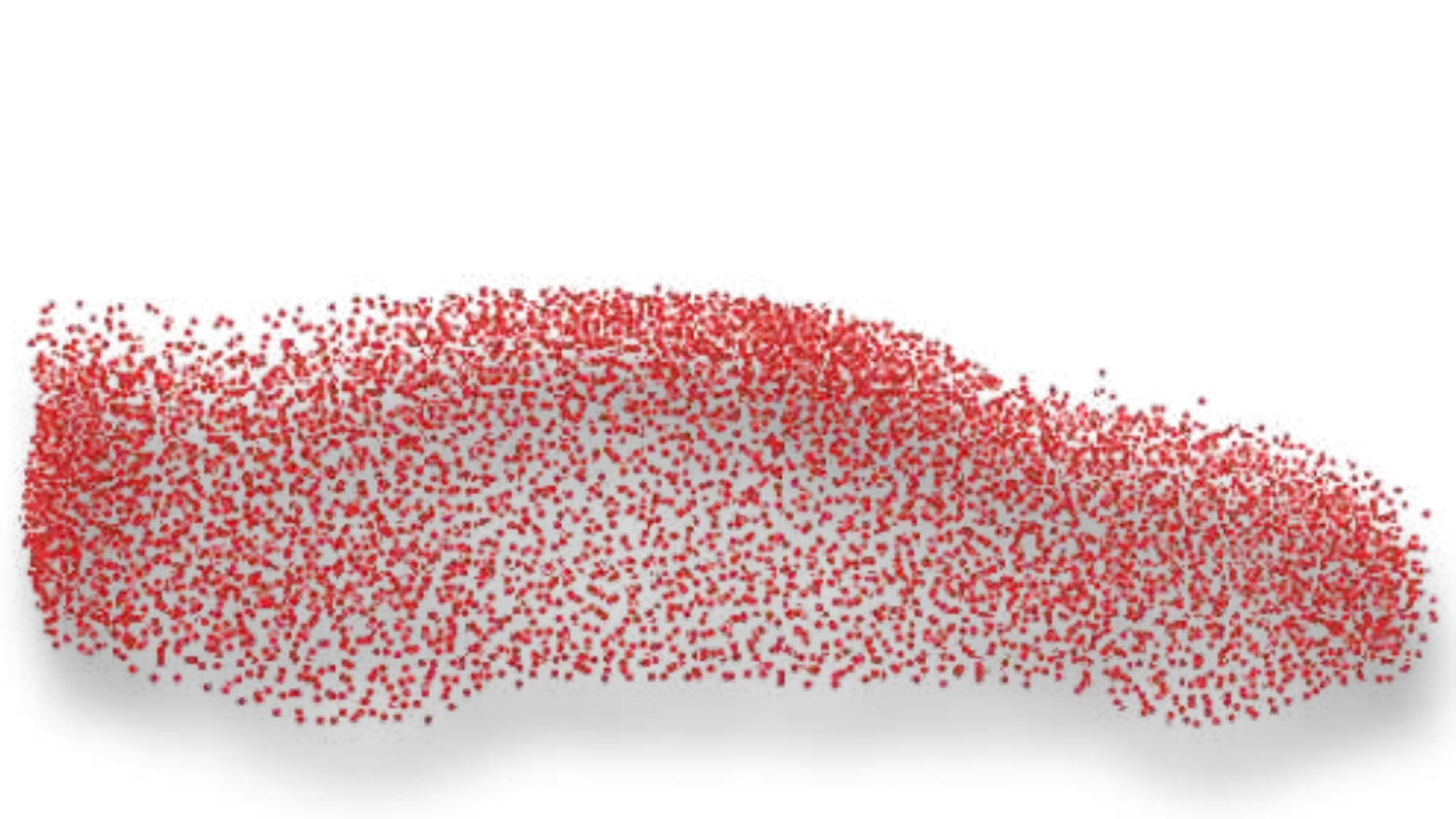}}
			\subfloat[\centering \tiny Complete 4]{
				\includegraphics[page=21, width = \patchSize]
				{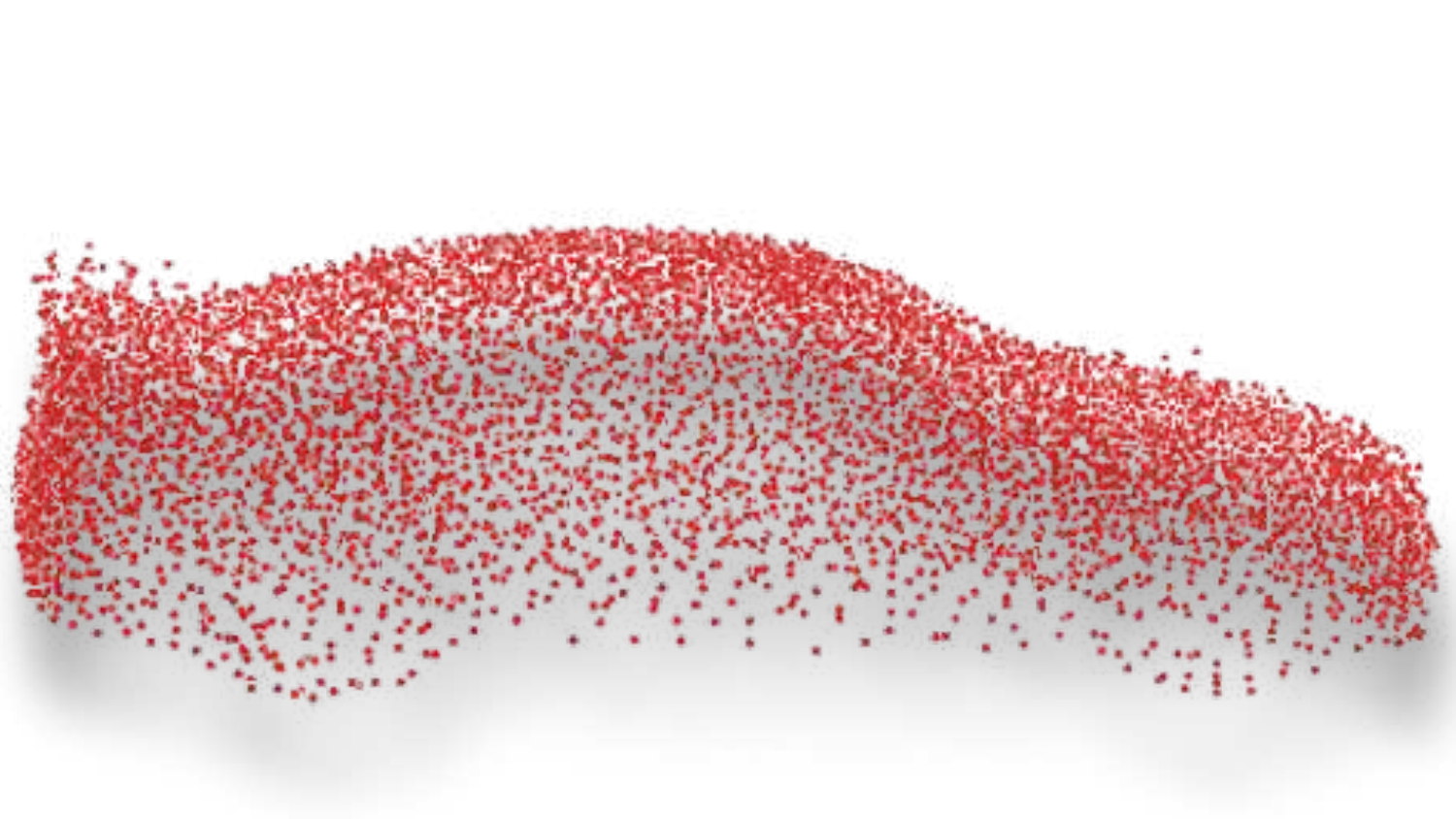}}
			\subfloat[\centering \tiny Complete 5]{
				\includegraphics[page=22, width = \patchSize]
				{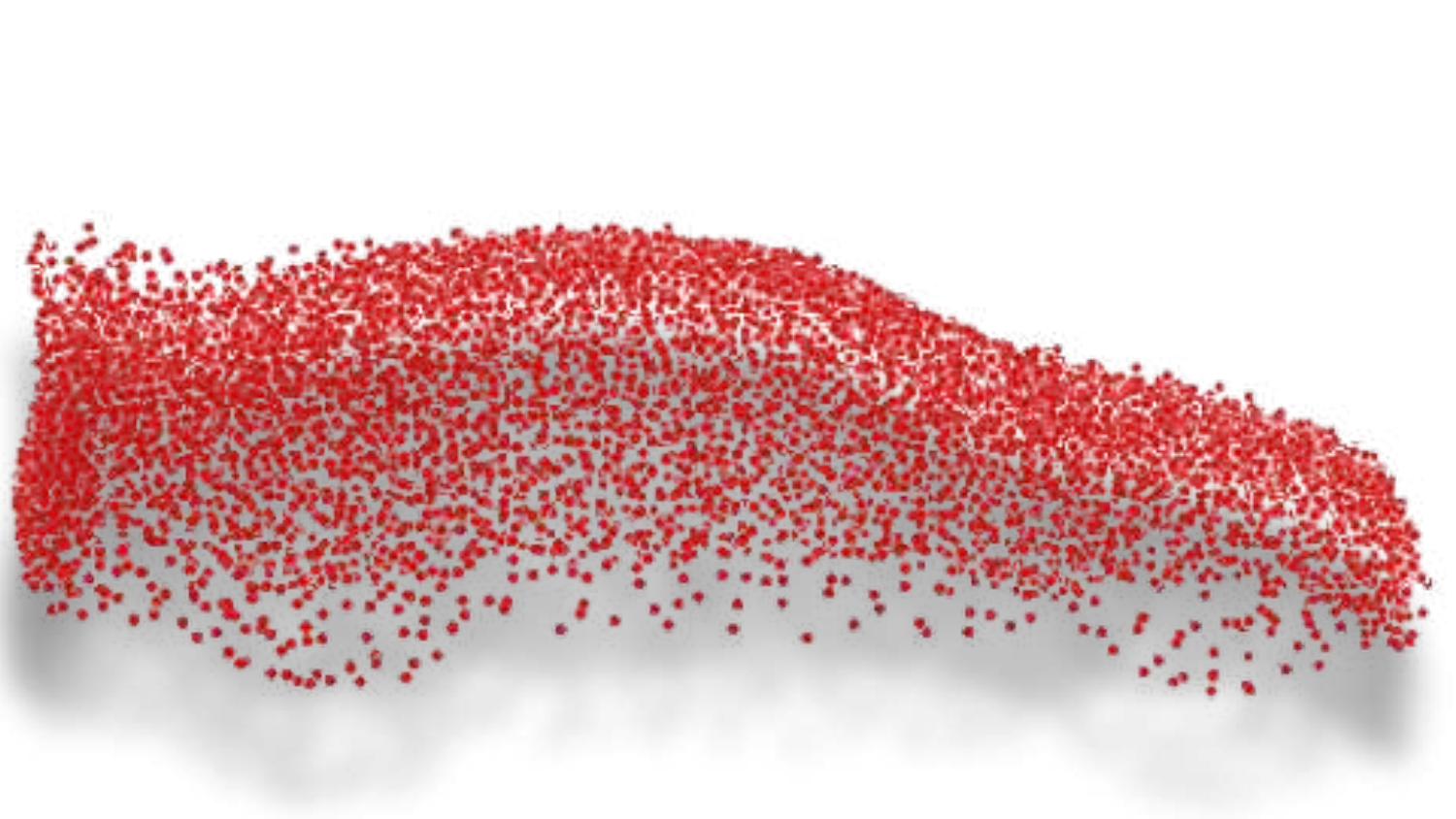}}\\
	    \end{tabular}
	    
\vspace{4mm}	    
\end{figure*}
\setcounter{figure}{3}
\begin{figure*}[ht]
	\captionsetup[subfloat]{labelformat=empty}
		\newcommand{\rowArg}{0.6cm}
		\newcommand{\fullSize}{2.cm}
		\newcommand{\patchSize}{2.6cm}
		\setlength{\tabcolsep}{0.05cm}		
            \begin{tabular}[b]{c c c}
			\multicolumn{2}{c}{\multirow{2}{*}[\rowArg]{
					\subfloat[GT]
					{\includegraphics[page=23, height=\fullSize]
					{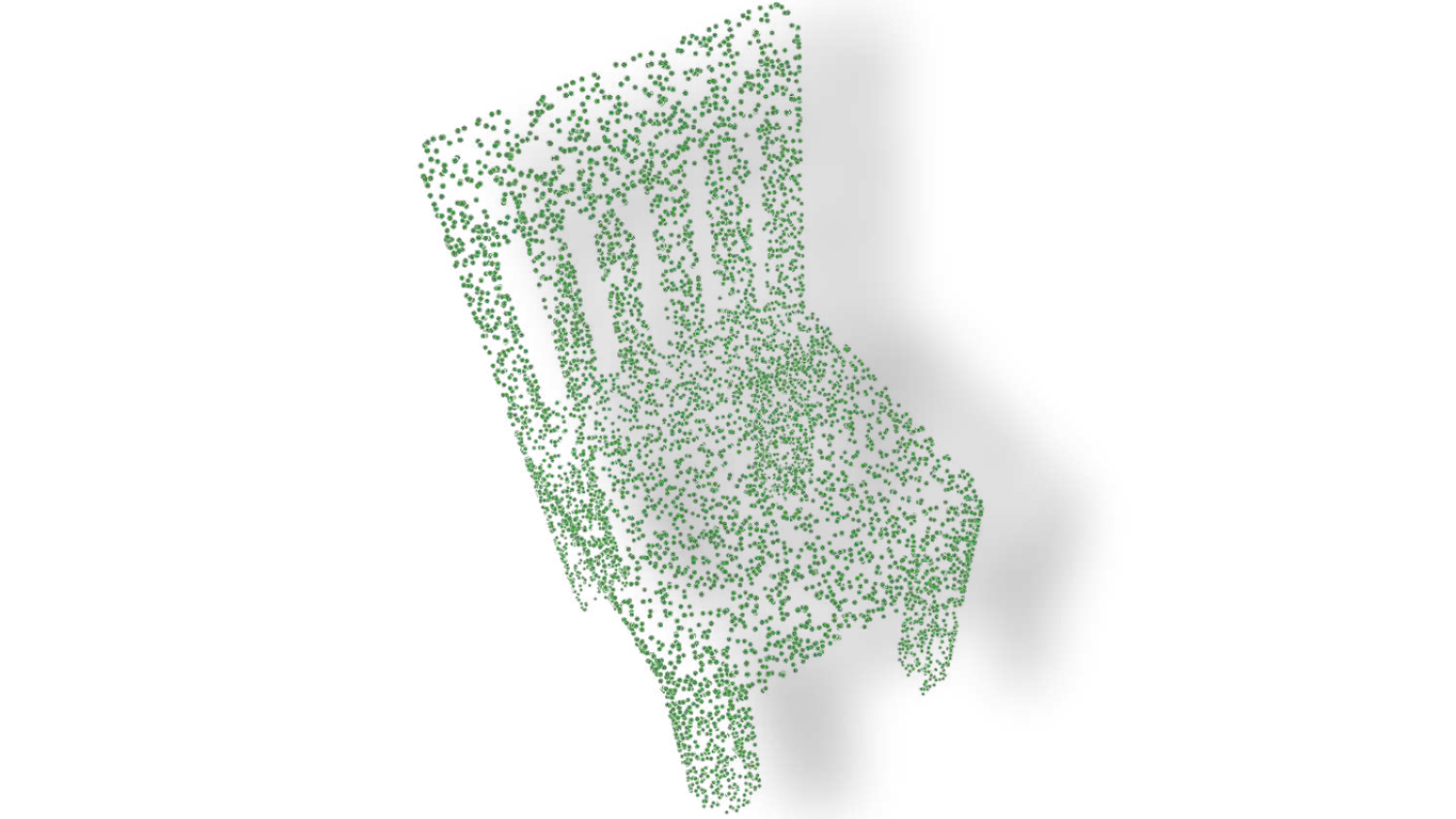}}}} &
			\subfloat[\centering \tiny Partial View 1 ]{
				\includegraphics[page=24, width = \patchSize]
				{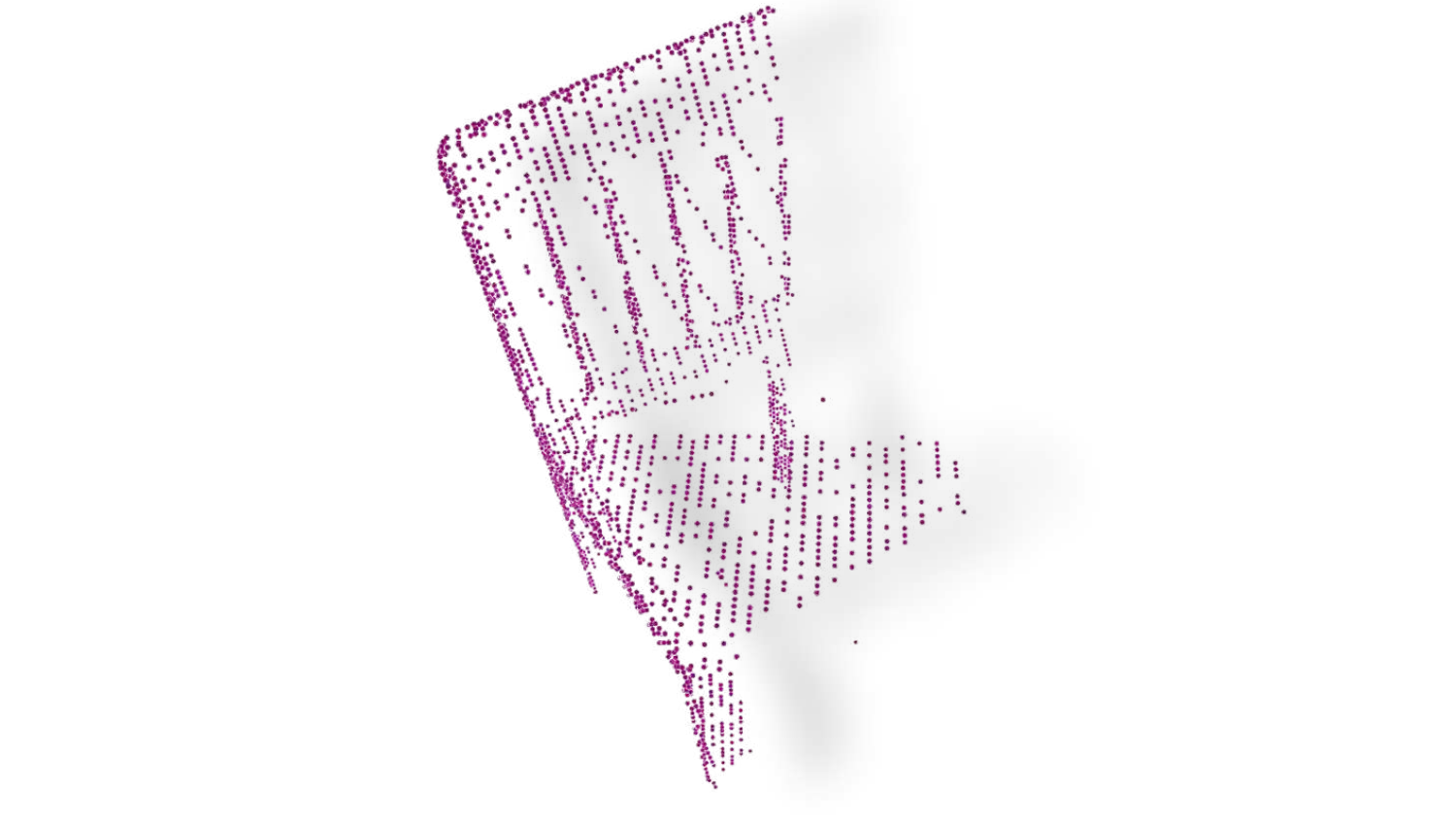}} 
			\subfloat[\centering \tiny Partial View 2  ]{
				\includegraphics[page=25, width = \patchSize]
				{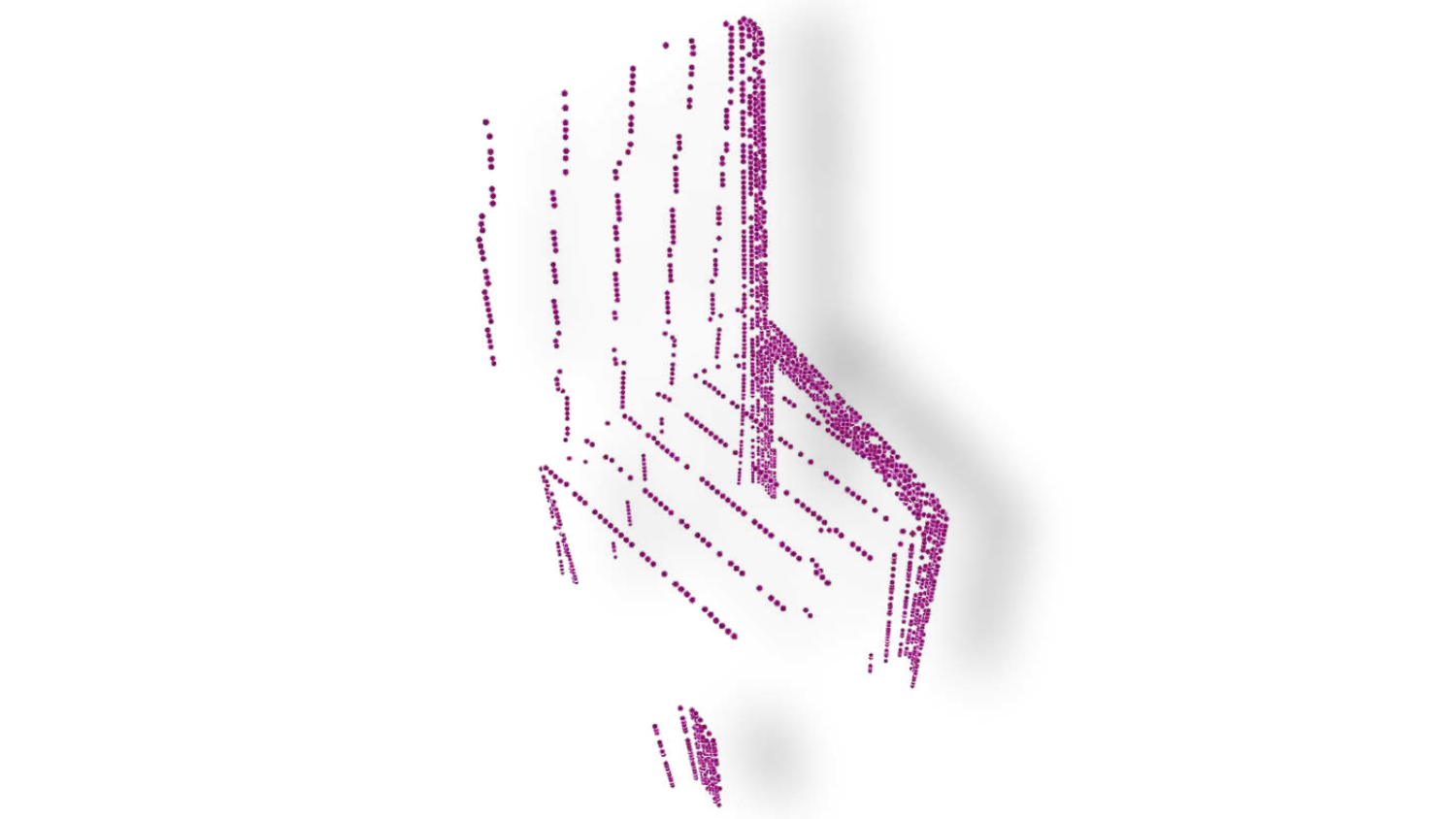}} 
			\subfloat[\centering \tiny Partial View 3  ]{
				\includegraphics[page=26, width = \patchSize]
				{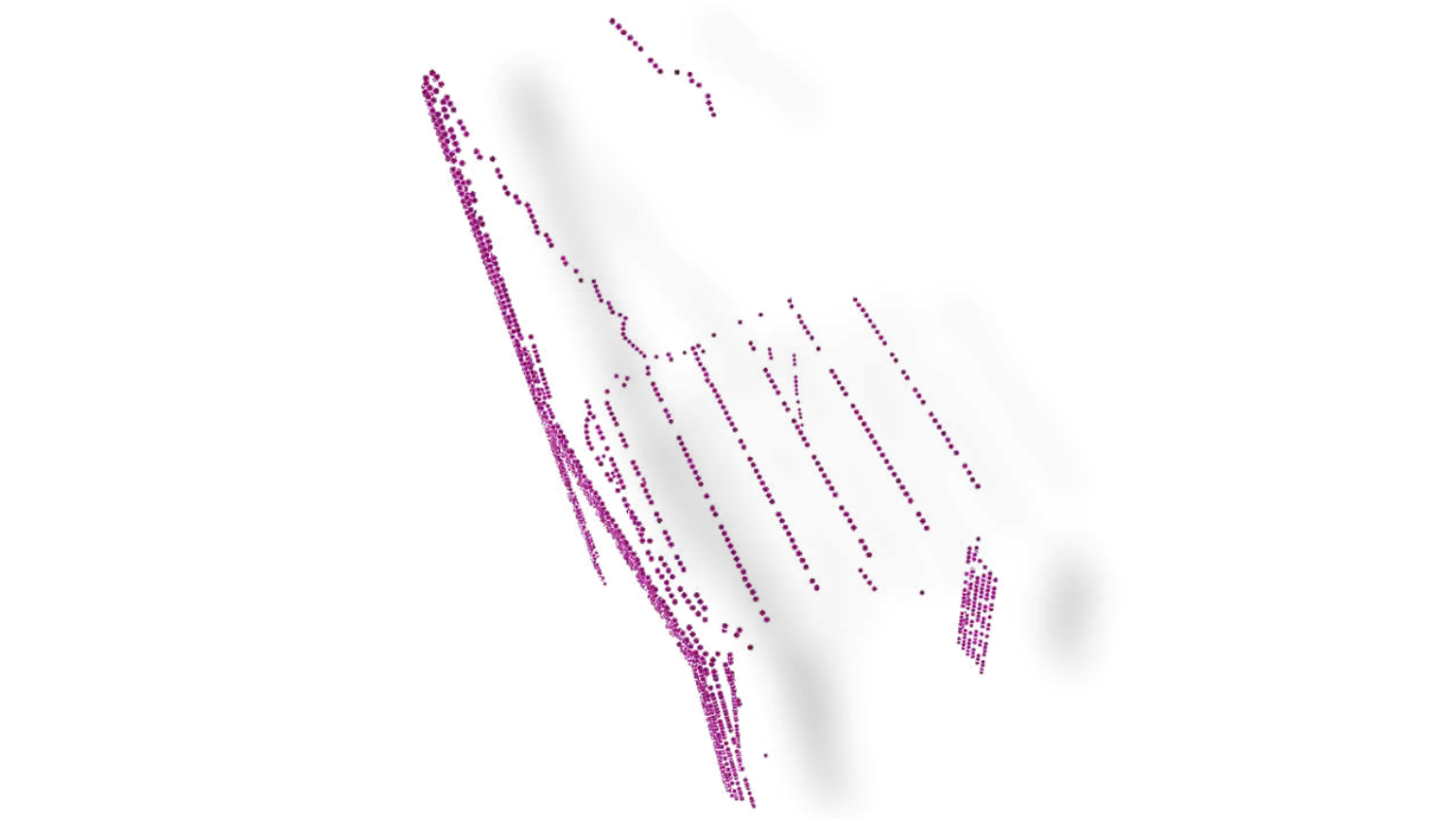}} 
			\subfloat[\centering \tiny Partial View 4  ]{
				\includegraphics[page=27, width = \patchSize]
				{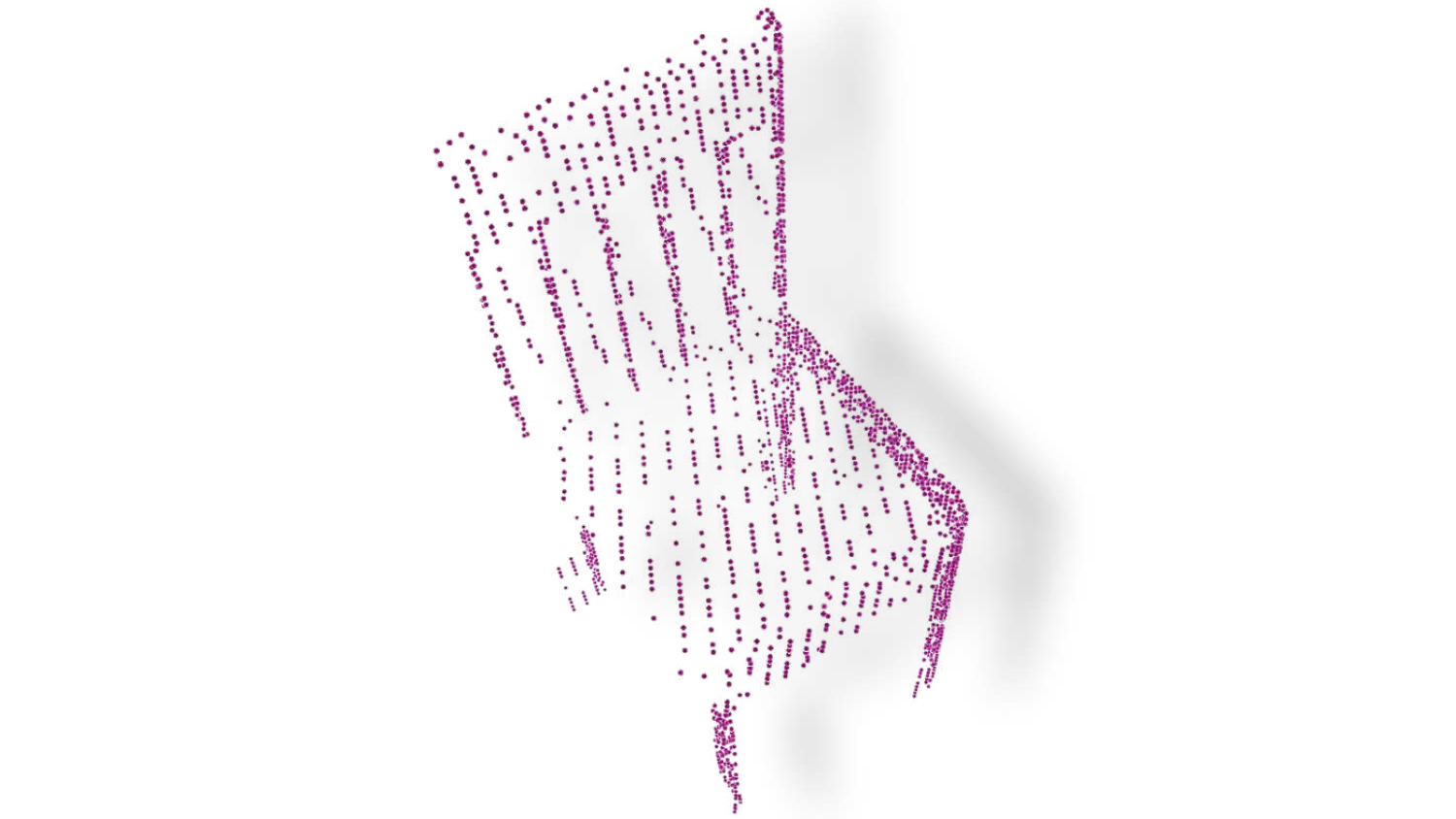}}
			\subfloat[\centering \tiny Partial View 5  ]{
				\includegraphics[page=28, width = \patchSize]
				{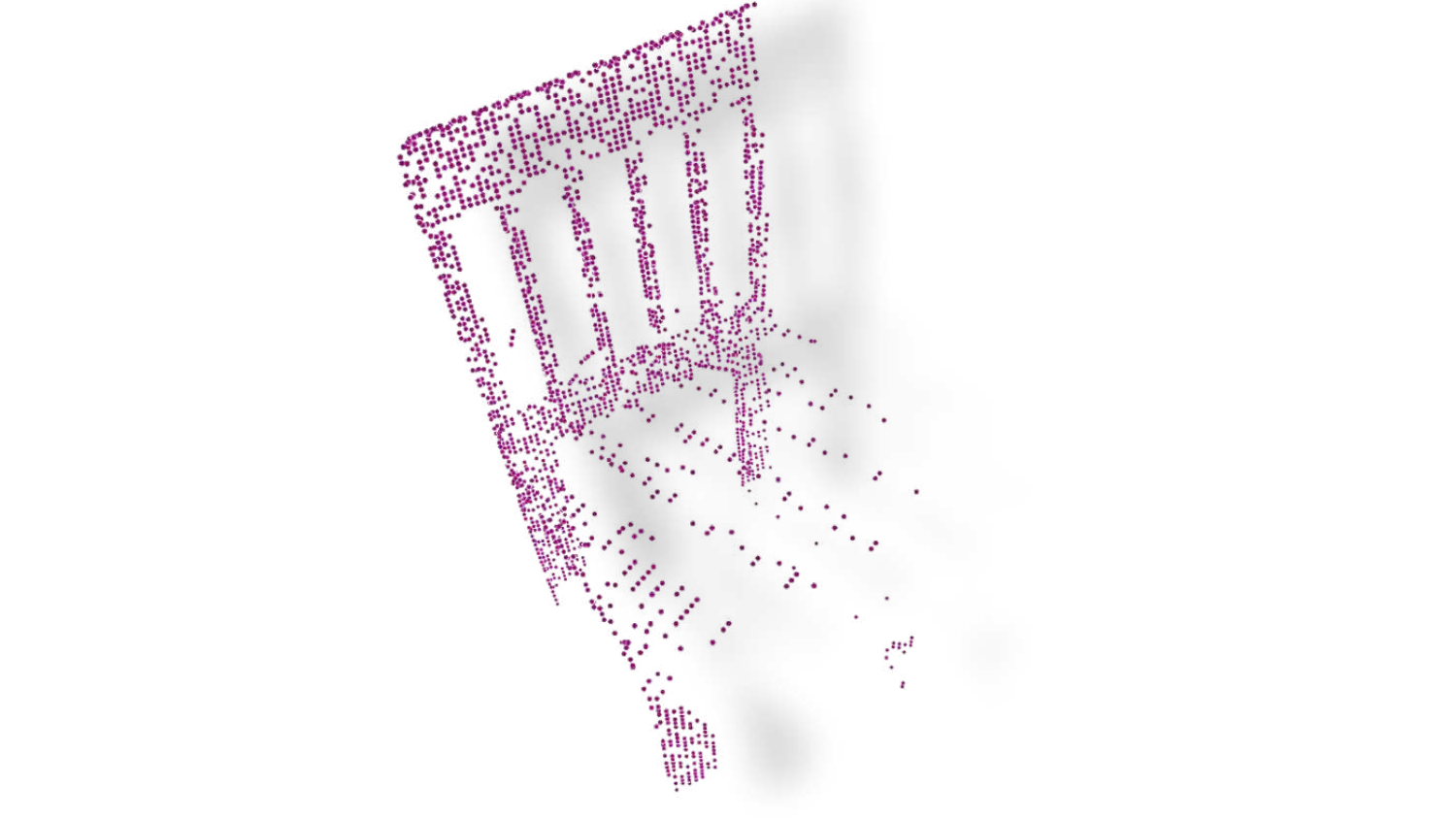}} \\  & &

			\subfloat[\centering  \tiny Complete 1  ]{
				\includegraphics[page=29, width = \patchSize]
				{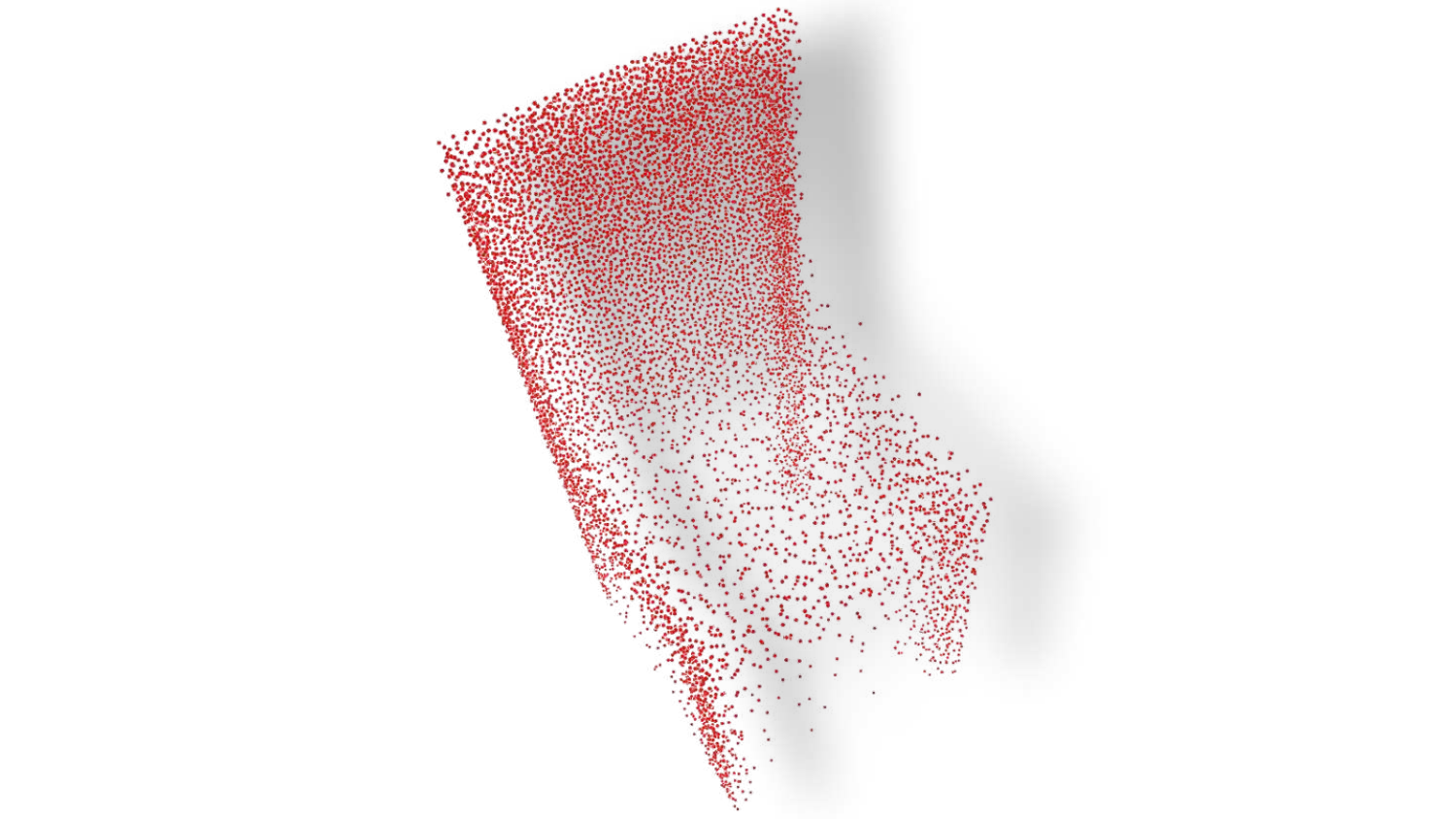}}
			\subfloat[\centering \tiny Complete 2 ]{
				\includegraphics[page=30, width = \patchSize]
				{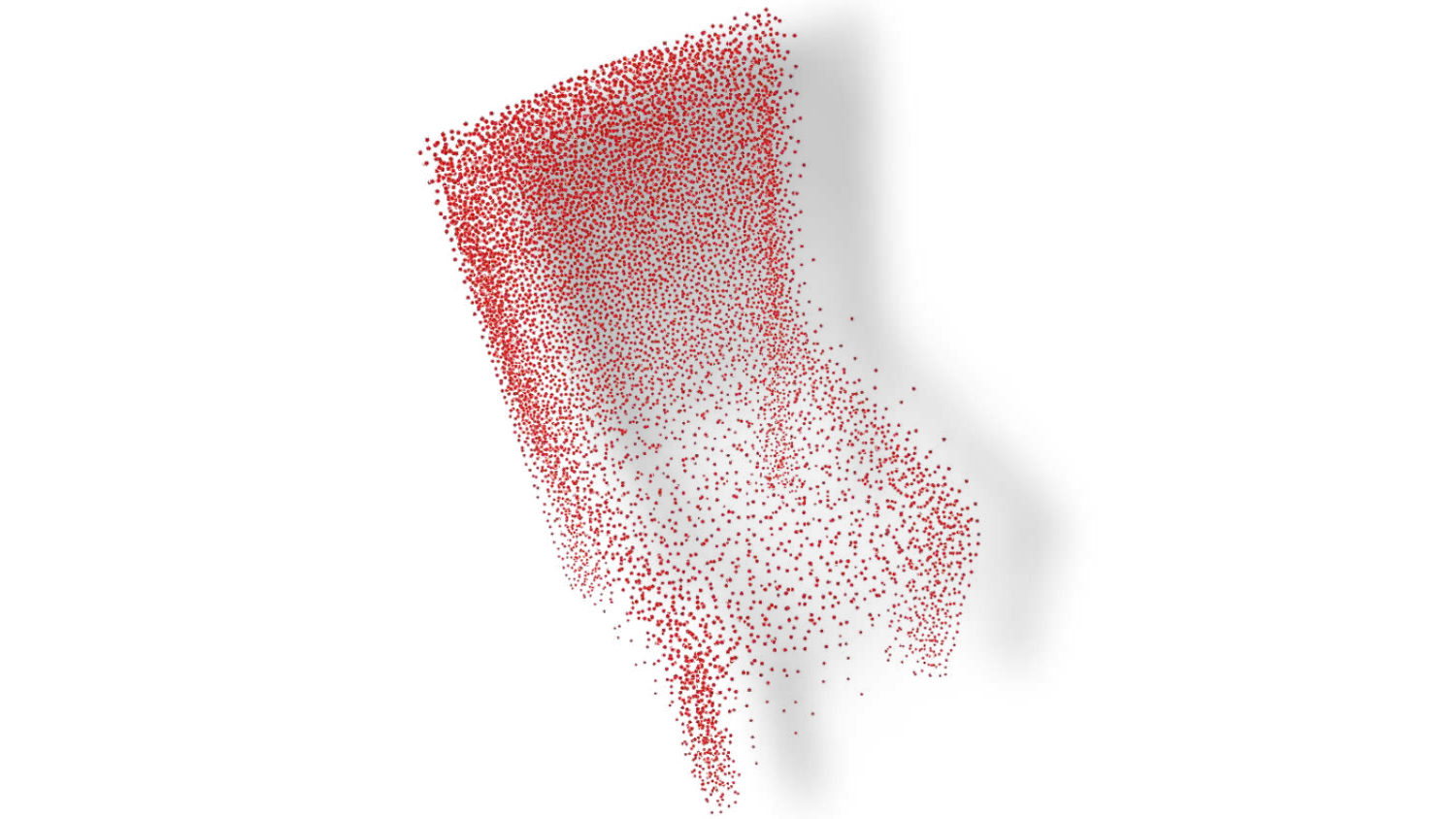}}
			\subfloat[\centering \tiny Complete 3 ]{
				\includegraphics[page=31, width = \patchSize]
				{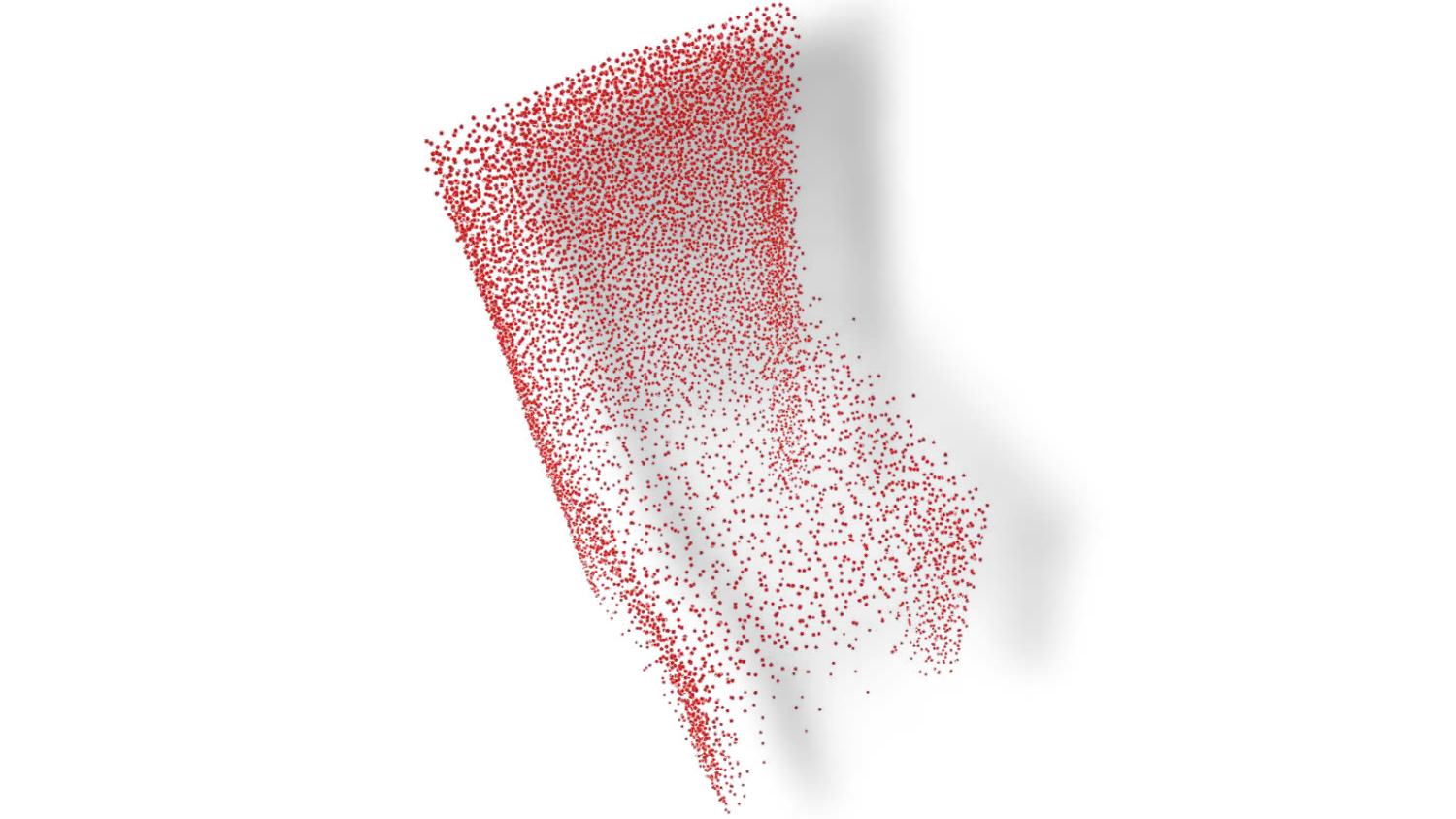}}
			\subfloat[\centering \tiny Complete 4]{
				\includegraphics[page=32, width = \patchSize]
				{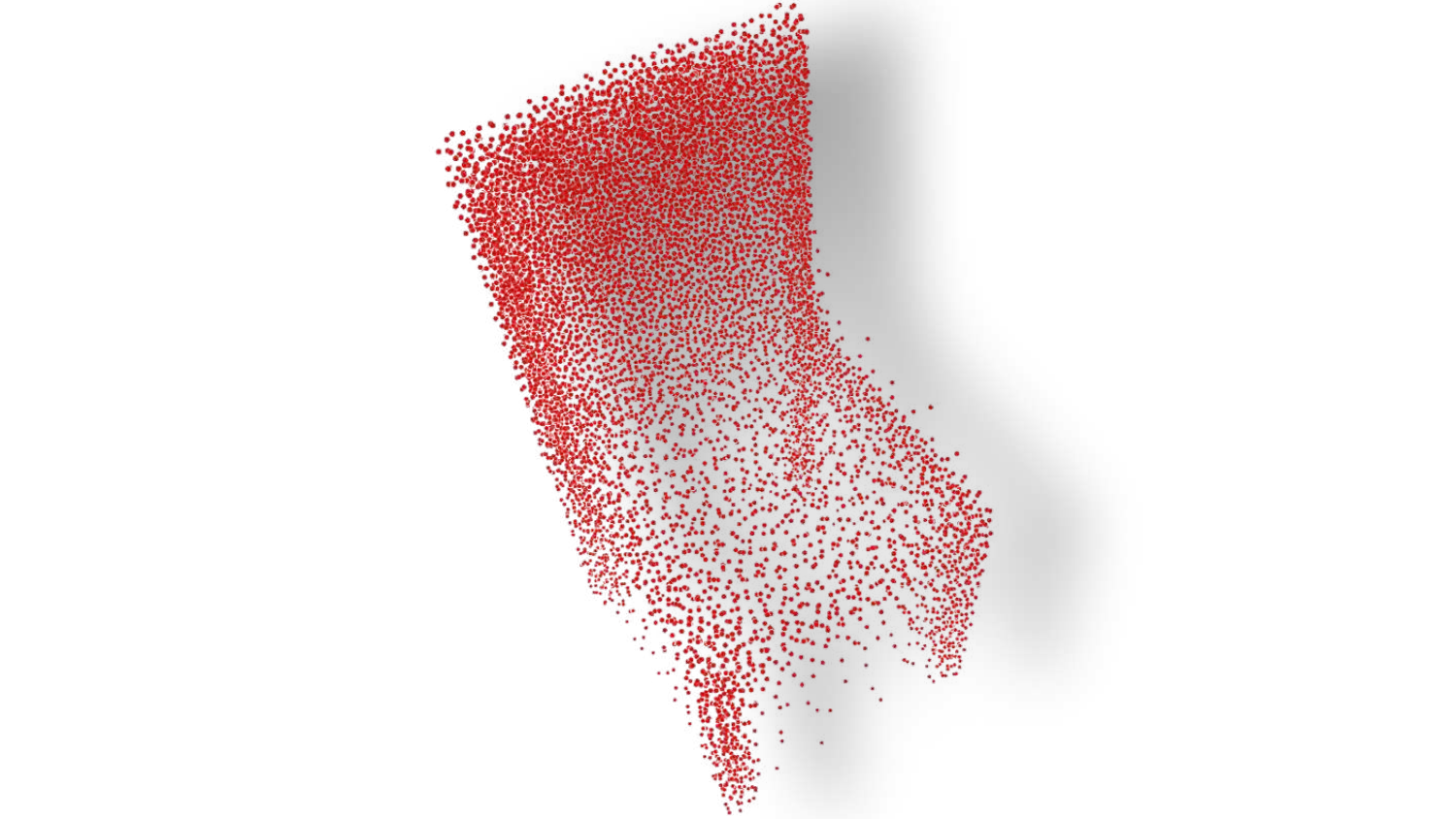}}
			\subfloat[\centering \tiny Complete 5]{
				\includegraphics[page=33, width = \patchSize]
				{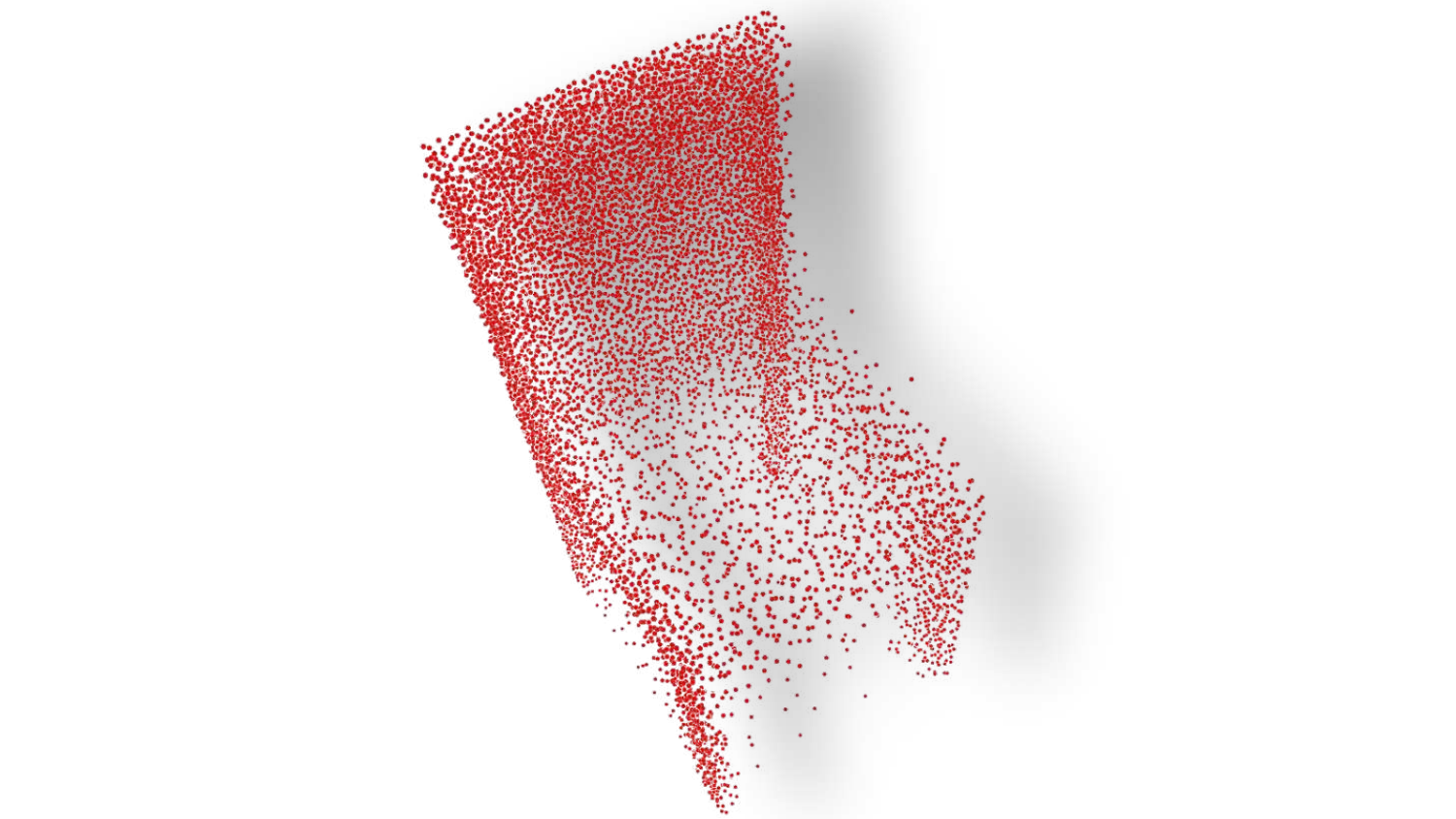}}\\
	    \end{tabular}
    \caption{\textbf{The qualitative effect of viewpoint.} 
    }
    \label{fig:supp_views}	    

\vspace{4mm}     
\end{figure*}
We also illustrate the effect of viewpoint in Figure~\ref{fig:supp_views}.
The results indicate that the quality of the generated complete point clouds is almost invariant to the point of view.
This robustness can be due to the attribute of our designed ACL-SPC framework that receives synthesized partial point clouds from random views during training.

\renewcommand{\thesection}{S7}
\section{Synthesized partial point clouds}
\begin{figure*}
     \centering
     \begin{subfigure}[b]{0.1\textwidth}
         \centering
         \includegraphics[page=1,width=\textwidth]{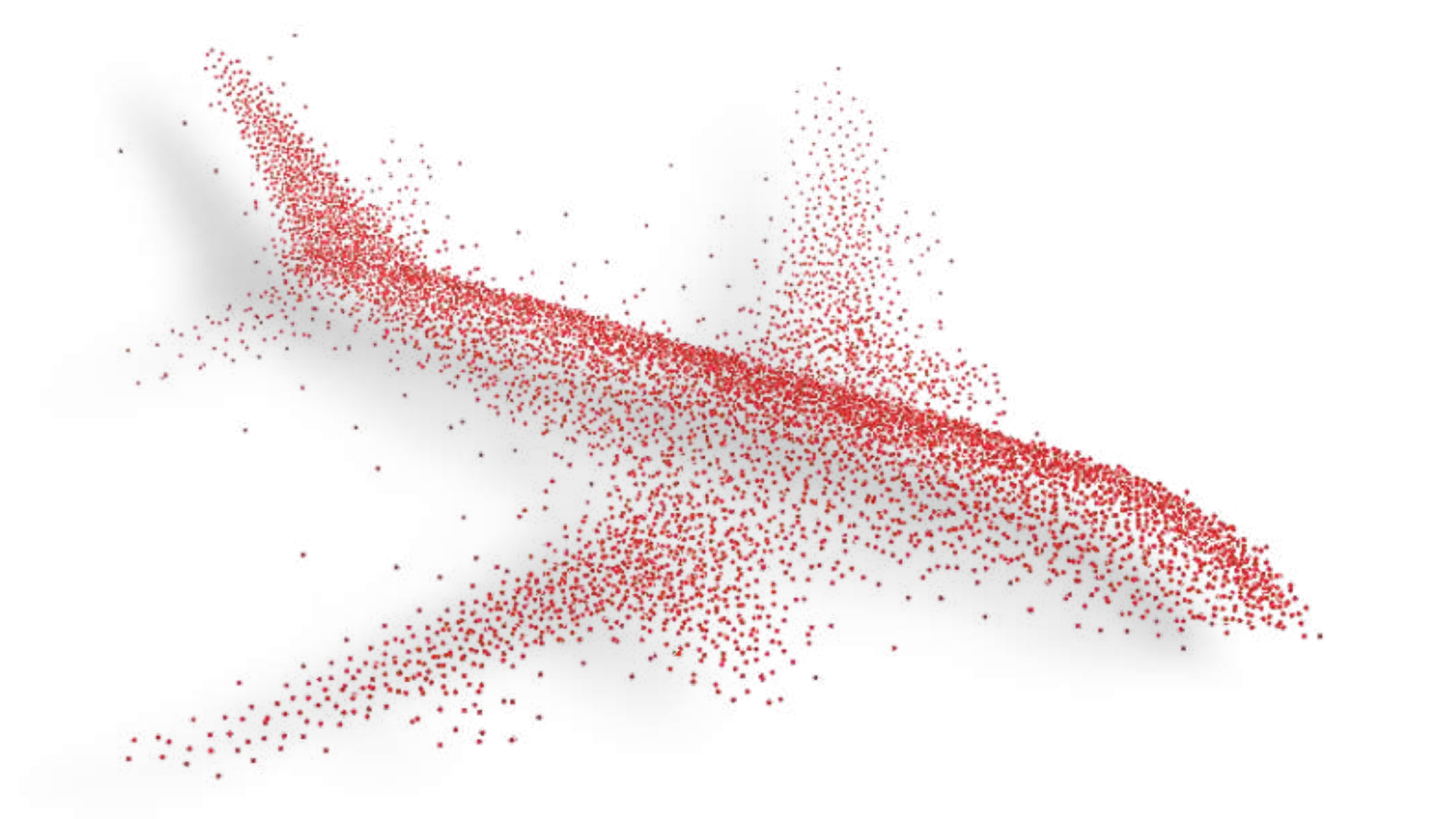}
     \end{subfigure}
     \hfill
     \begin{subfigure}[b]{0.1\textwidth}
         \centering
         \includegraphics[page=2,width=\textwidth]{figures/supp_syn.pdf}
     \end{subfigure}
     \hfill
     \begin{subfigure}[b]{0.1\textwidth}
         \centering
         \includegraphics[page=3,width=\textwidth]{figures/supp_syn.pdf}
     \end{subfigure}
     \hfill
     \begin{subfigure}[b]{0.1\textwidth}
         \centering
         \includegraphics[page=4,width=\textwidth]{figures/supp_syn.pdf}
     \end{subfigure}
     \hfill
     \begin{subfigure}[b]{0.1\textwidth}
         \centering
         \includegraphics[page=5,width=\textwidth]{figures/supp_syn.pdf}
     \end{subfigure}
     \hfill
     \begin{subfigure}[b]{0.1\textwidth}
         \centering
         \includegraphics[page=6,width=\textwidth]{figures/supp_syn.pdf}
     \end{subfigure}
     \hfill
     \begin{subfigure}[b]{0.1\textwidth}
         \centering
         \includegraphics[page=7,width=\textwidth]{figures/supp_syn.pdf}
     \end{subfigure}
     \hfill
     \begin{subfigure}[b]{0.1\textwidth}
         \centering
         \includegraphics[page=8,width=\textwidth]{figures/supp_syn.pdf}
     \end{subfigure}
     \hfill
     \begin{subfigure}[b]{0.1\textwidth}
         \centering
         \includegraphics[page=9,width=\textwidth]{figures/supp_syn.pdf}
     \end{subfigure}
     \\
     \begin{subfigure}[b]{0.1\textwidth}
         \centering
         \includegraphics[page=10,width=\textwidth]{figures/supp_syn.pdf}
     \end{subfigure}
     \hfill
     \begin{subfigure}[b]{0.1\textwidth}
         \centering
         \includegraphics[page=11,width=\textwidth]{figures/supp_syn.pdf}
     \end{subfigure}
     \hfill   
     \begin{subfigure}[b]{0.1\textwidth}
         \centering
         \includegraphics[page=12,width=\textwidth]{figures/supp_syn.pdf}
     \end{subfigure}
     \hfill
     \begin{subfigure}[b]{0.1\textwidth}
         \centering
         \includegraphics[page=13,width=\textwidth]{figures/supp_syn.pdf}
     \end{subfigure}
     \hfill 
     \begin{subfigure}[b]{0.1\textwidth}
         \centering
         \includegraphics[page=14,width=\textwidth]{figures/supp_syn.pdf}
     \end{subfigure}
     \hfill
     \begin{subfigure}[b]{0.1\textwidth}
         \centering
         \includegraphics[page=15,width=\textwidth]{figures/supp_syn.pdf}
     \end{subfigure}
     \hfill
     \begin{subfigure}[b]{0.1\textwidth}
         \centering
         \includegraphics[page=16,width=\textwidth]{figures/supp_syn.pdf}
     \end{subfigure}
     \hfill
     \begin{subfigure}[b]{0.1\textwidth}
         \centering
         \includegraphics[page=17,width=\textwidth]{figures/supp_syn.pdf}
     \end{subfigure}
     \hfill
     \begin{subfigure}[b]{0.1\textwidth}
         \centering
         \includegraphics[page=18,width=\textwidth]{figures/supp_syn.pdf}
     \end{subfigure}
     \\
     \begin{subfigure}[b]{0.1\textwidth}
         \centering
         \includegraphics[page=19,width=\textwidth]{figures/supp_syn.pdf}
         \caption*{$C_0$}
     \end{subfigure}
     \hfill
     \begin{subfigure}[b]{0.1\textwidth}
         \centering
         \includegraphics[page=20,width=\textwidth]{figures/supp_syn.pdf}
         \caption*{$P_{v_1}$}
     \end{subfigure}
     \hfill
     \begin{subfigure}[b]{0.1\textwidth}
         \centering
         \includegraphics[page=21,width=\textwidth]{figures/supp_syn.pdf}
         \caption*{$P_{v_2}$}
     \end{subfigure}
     \hfill
     \begin{subfigure}[b]{0.1\textwidth}
         \centering
         \includegraphics[page=22,width=\textwidth]{figures/supp_syn.pdf}
         \caption*{$P_{v_3}$}
     \end{subfigure}
     \hfill
     \begin{subfigure}[b]{0.1\textwidth}
         \centering
         \includegraphics[page=23,width=\textwidth]{figures/supp_syn.pdf}
         \caption*{$P_{v_4}$}
     \end{subfigure}   
     \hfill
     \begin{subfigure}[b]{0.1\textwidth}
         \centering
         \includegraphics[page=24,width=\textwidth]{figures/supp_syn.pdf}
         \caption*{$P_{v_5}$}
     \end{subfigure}
     \hfill
     \begin{subfigure}[b]{0.1\textwidth}
         \centering
         \includegraphics[page=25,width=\textwidth]{figures/supp_syn.pdf}
         \caption*{$P_{v_6}$}
     \end{subfigure}
     \hfill
     \begin{subfigure}[b]{0.1\textwidth}
         \centering
         \includegraphics[page=26,width=\textwidth]{figures/supp_syn.pdf}
         \caption*{$P_{v_7}$}
     \end{subfigure}
     \hfill
     \begin{subfigure}[b]{0.1\textwidth}
         \centering
         \includegraphics[page=27,width=\textwidth]{figures/supp_syn.pdf}
         \caption*{$P_{v_8}$}
     \end{subfigure}
        \caption{
        \textbf{Visualization of the synthesized partial point clouds.} 
        }
        \label{fig:supp_syns}
\vspace{4mm}        
\end{figure*}     

We visualize the generated synthetic partial point clouds from the predicted complete output in Figure~\ref{fig:supp_syns}.
It shows that our partial point cloud generator can synthesize realistic partial point clouds from a predicted complete output.

\renewcommand{\thesection}{S8}
\section{Training dataset}
\begin{figure*}[t]
     \centering
     \begin{subfigure}[b]{0.1\textwidth}
         \centering
         \includegraphics[page=1,width=\textwidth]{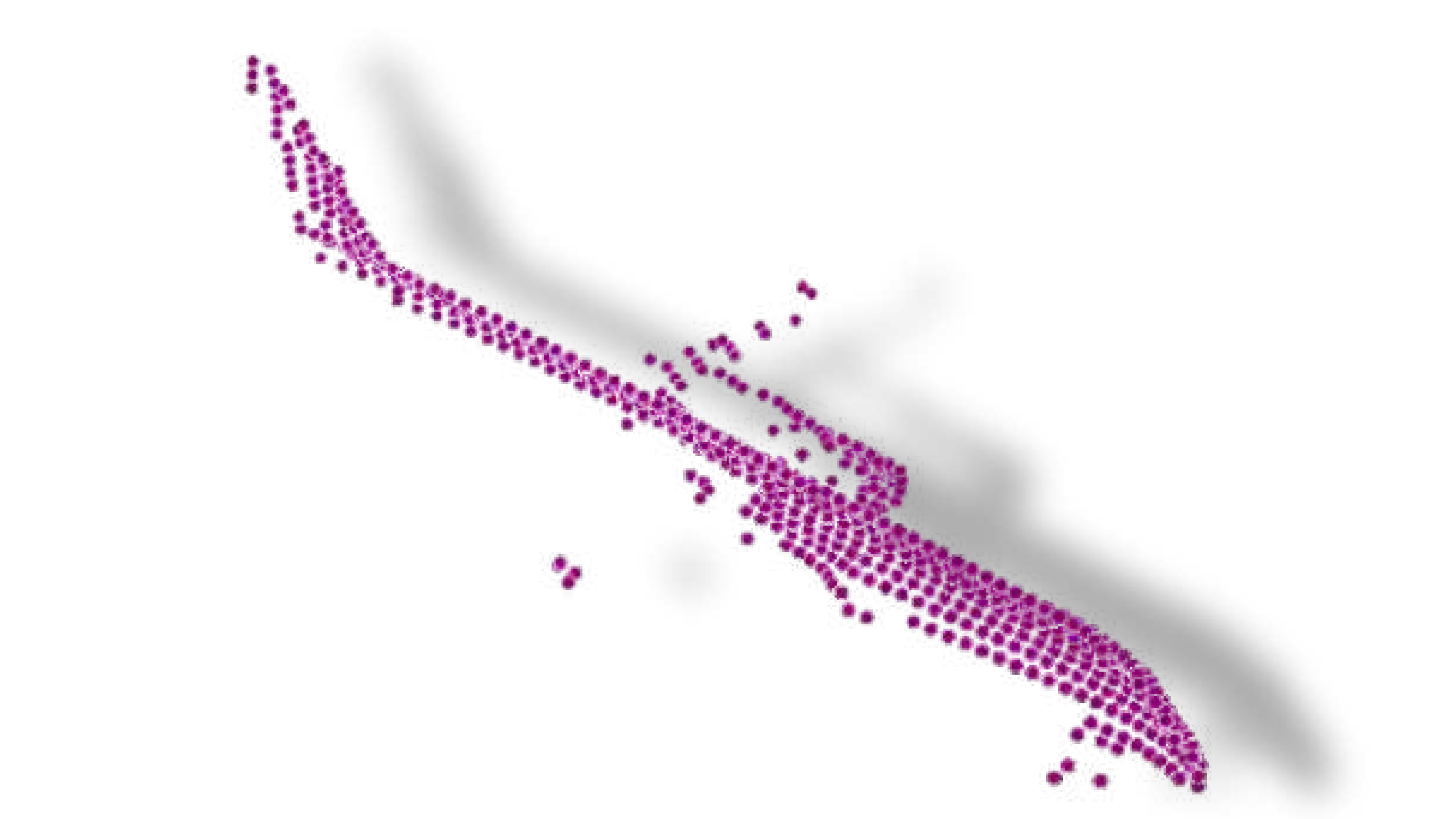}
     \end{subfigure}
     \hfill
     \begin{subfigure}[b]{0.1\textwidth}
         \centering
         \includegraphics[page=2,width=\textwidth]{figures/supp_multi.pdf}
     \end{subfigure}
     \hfill
     \begin{subfigure}[b]{0.1\textwidth}
         \centering
         \includegraphics[page=3,width=\textwidth]{figures/supp_multi.pdf}
     \end{subfigure}
     \hfill
     \begin{subfigure}[b]{0.1\textwidth}
         \centering
         \includegraphics[page=4,width=\textwidth]{figures/supp_multi.pdf}
     \end{subfigure}
     \hfill
     \hfill
     \begin{subfigure}[b]{0.1\textwidth}
         \centering
         \includegraphics[page=5,width=\textwidth]{figures/supp_multi.pdf}
     \end{subfigure}
     \hfill
     \begin{subfigure}[b]{0.1\textwidth}
         \centering
         \includegraphics[page=6,width=\textwidth]{figures/supp_multi.pdf}
     \end{subfigure}
     \hfill
     \begin{subfigure}[b]{0.1\textwidth}
         \centering
         \includegraphics[page=7,width=\textwidth]{figures/supp_multi.pdf}
     \end{subfigure}
     \hfill
     \begin{subfigure}[b]{0.1\textwidth}
         \centering
         \includegraphics[page=8,width=\textwidth]{figures/supp_multi.pdf}
     \end{subfigure}
     \\
     \begin{subfigure}[b]{0.1\textwidth}
         \centering
         \includegraphics[page=9,width=\textwidth]{figures/supp_multi.pdf}
     \end{subfigure}
     \hfill
     \begin{subfigure}[b]{0.1\textwidth}
         \centering
         \includegraphics[page=10,width=\textwidth]{figures/supp_multi.pdf}
     \end{subfigure}
     \hfill   
     \begin{subfigure}[b]{0.1\textwidth}
         \centering
         \includegraphics[page=11,width=\textwidth]{figures/supp_multi.pdf}
     \end{subfigure}
     \hfill
     \begin{subfigure}[b]{0.1\textwidth}
         \centering
         \includegraphics[page=12,width=\textwidth]{figures/supp_multi.pdf}
     \end{subfigure}
     \hfill 
     \hfill
     \begin{subfigure}[b]{0.1\textwidth}
         \centering
         \includegraphics[page=13,width=\textwidth]{figures/supp_multi.pdf}
     \end{subfigure}
     \hfill
     \begin{subfigure}[b]{0.1\textwidth}
         \centering
         \includegraphics[page=14,width=\textwidth]{figures/supp_multi.pdf}
     \end{subfigure}
     \hfill
     \begin{subfigure}[b]{0.1\textwidth}
         \centering
         \includegraphics[page=15,width=\textwidth]{figures/supp_multi.pdf}
     \end{subfigure}
     \hfill
     \begin{subfigure}[b]{0.1\textwidth}
         \centering
         \includegraphics[page=16,width=\textwidth]{figures/supp_multi.pdf}
     \end{subfigure}
     \\
     \begin{subfigure}[b]{0.1\textwidth}
         \centering
         \includegraphics[page=17,width=\textwidth]{figures/supp_multi.pdf}
         \caption*{Input}
     \end{subfigure}
     \hfill
     \begin{subfigure}[b]{0.1\textwidth}
         \centering
         \includegraphics[page=18,width=\textwidth]{figures/supp_multi.pdf}
         \caption*{Single}
     \end{subfigure}
     \hfill
     \begin{subfigure}[b]{0.1\textwidth}
         \centering
         \includegraphics[page=19,width=\textwidth]{figures/supp_multi.pdf}
         \caption*{Multi}
     \end{subfigure}
     \hfill
     \begin{subfigure}[b]{0.1\textwidth}
         \centering
         \includegraphics[page=20,width=\textwidth]{figures/supp_multi.pdf}
         \caption*{GT}
     \end{subfigure}
     \hfill
     \hfill
     \begin{subfigure}[b]{0.1\textwidth}
         \centering
         \includegraphics[page=21,width=\textwidth]{figures/supp_multi.pdf}
         \caption*{Input}
     \end{subfigure}   
     \hfill
     \begin{subfigure}[b]{0.1\textwidth}
         \centering
         \includegraphics[page=22,width=\textwidth]{figures/supp_multi.pdf}
         \caption*{Single}
     \end{subfigure}
     \hfill
     \begin{subfigure}[b]{0.1\textwidth}
         \centering
         \includegraphics[page=23,width=\textwidth]{figures/supp_multi.pdf}
         \caption*{Multi}
     \end{subfigure}
     \hfill
     \begin{subfigure}[b]{0.1\textwidth}
         \centering
         \includegraphics[page=24,width=\textwidth]{figures/supp_multi.pdf}
         \caption*{GT}
     \end{subfigure}
        \caption{
        \textbf{Qualitative comparisons of our method trained on multi-class vs. single-class.} 
        }
        \label{fig:supp_training}

\vspace{4mm}        
\end{figure*}     

In Figure~\ref{fig:supp_training}, we compare our method trained individually on each category to one trained on multiple categories.
Our qualitative comparisons show that our method trained on a training dataset with multiple categories can fill in the missing region similarly to the single case.
As a result, our method is practical in real-world scenarios involving a variety of samples from various categories. 
\renewcommand{\thesection}{S9}
\section{Failure cases}
\begin{figure*}[t]
     \centering
     \begin{subfigure}[b]{0.13\textwidth}
         \centering
         \includegraphics[page=1,width=\textwidth]{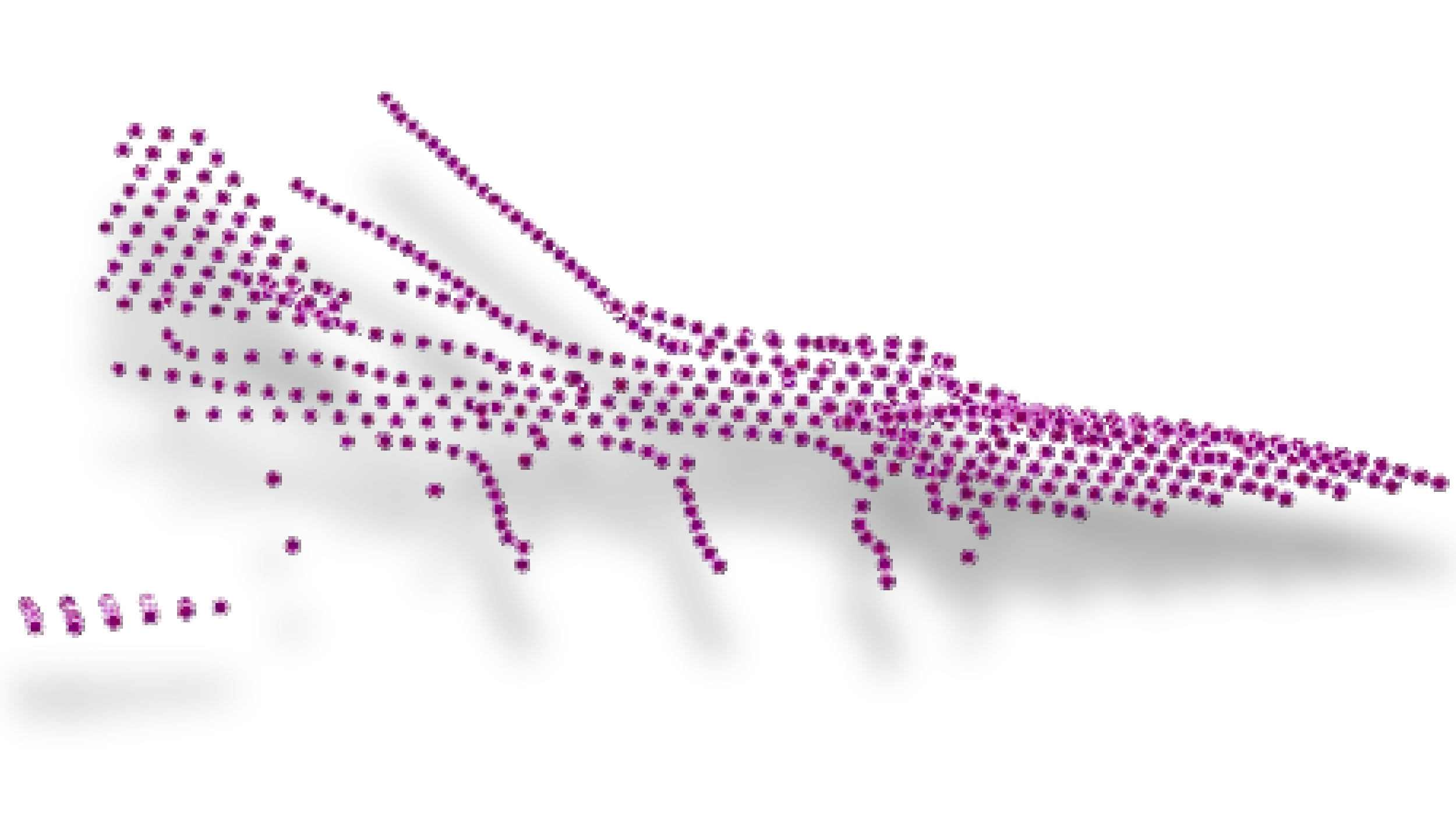}
     \end{subfigure}
     \hfill
     \begin{subfigure}[b]{0.13\textwidth}
         \centering
         \includegraphics[page=2,width=\textwidth]{figures/vis_failure.pdf}
     \end{subfigure}
     \hfill
     \begin{subfigure}[b]{0.13\textwidth}
         \centering
         \includegraphics[page=3,width=\textwidth]{figures/vis_failure.pdf}
     \end{subfigure}
     \hfill
     \hfill
     \begin{subfigure}[b]{0.13\textwidth}
         \centering
         \includegraphics[page=4,width=\textwidth]{figures/vis_failure.pdf}
     \end{subfigure}
     \hfill
     \begin{subfigure}[b]{0.13\textwidth}
         \centering
         \includegraphics[page=5,width=\textwidth]{figures/vis_failure.pdf}
     \end{subfigure}
     \hfill
     \begin{subfigure}[b]{0.13\textwidth}
         \centering
         \includegraphics[page=6,width=\textwidth]{figures/vis_failure.pdf}
     \end{subfigure}
     \\
     \begin{subfigure}[b]{0.13\textwidth}
         \centering
         \includegraphics[page=7,width=\textwidth]{figures/vis_failure.pdf}
     \end{subfigure}
     \hfill
     \begin{subfigure}[b]{0.13\textwidth}
         \centering
         \includegraphics[page=8,width=\textwidth]{figures/vis_failure.pdf}
     \end{subfigure}
     \hfill   
     \begin{subfigure}[b]{0.13\textwidth}
         \centering
         \includegraphics[page=9,width=\textwidth]{figures/vis_failure.pdf}
     \end{subfigure}
     \hfill
     \hfill
     \begin{subfigure}[b]{0.13\textwidth}
         \centering
         \includegraphics[page=10,width=\textwidth]{figures/vis_failure.pdf}
     \end{subfigure}
     \hfill 
     \begin{subfigure}[b]{0.13\textwidth}
         \centering
         \includegraphics[page=11,width=\textwidth]{figures/vis_failure.pdf}
     \end{subfigure}
     \hfill
     \begin{subfigure}[b]{0.13\textwidth}
         \centering
         \includegraphics[page=12,width=\textwidth]{figures/vis_failure.pdf}
     \end{subfigure}
     \\
     \begin{subfigure}[b]{0.13\textwidth}
         \centering
         \includegraphics[page=13,width=\textwidth]{figures/vis_failure.pdf}
         \caption*{Input}
     \end{subfigure}
     \hfill
     \begin{subfigure}[b]{0.13\textwidth}
         \centering
         \includegraphics[page=14,width=\textwidth]{figures/vis_failure.pdf}
         \caption*{Output}
     \end{subfigure}
     \hfill
     \begin{subfigure}[b]{0.13\textwidth}
         \centering
         \includegraphics[page=15,width=\textwidth]{figures/vis_failure.pdf}
         \caption*{GT}
     \end{subfigure}
     \hfill
     \hfill
     \begin{subfigure}[b]{0.13\textwidth}
         \centering
         \includegraphics[page=16,width=\textwidth]{figures/vis_failure.pdf}
         \caption*{Input}
     \end{subfigure}
     \hfill
     \begin{subfigure}[b]{0.13\textwidth}
         \centering
         \includegraphics[page=17,width=\textwidth]{figures/vis_failure.pdf}
         \caption*{Output}
     \end{subfigure}   
     \hfill
     \begin{subfigure}[b]{0.13\textwidth}
         \centering
         \includegraphics[page=18,width=\textwidth]{figures/vis_failure.pdf}
         \caption*{GT}
     \end{subfigure}
        \caption{
        \textbf{Visualization of failure results.} 
        }
        \label{fig:supp_failure}
\end{figure*}     

We also show our failure results in Figure~\ref{fig:supp_failure}.
We show that our method generates redundant points when similar objects are rare in the training dataset.
However, it can still restore the missing parts in the given partial point cloud.
As mentioned in the manuscript, it is due to no restrictions to not generate redundant points which leads to high precision values.
We will develop our self-supervised framework in denoising tasks to improve the results for the failure cases.
\renewcommand{\thesection}{S10}
\section{Ablation on $\alpha$ and $\beta$}
\begin{table*}[t]
    \small
    \centering
    \setlength\tabcolsep{0.005pt}
    \begin{tabularx}{\linewidth}{l >{\centering\arraybackslash}X >{\centering\arraybackslash}X >{\centering\arraybackslash}X  >{\centering\arraybackslash}X
    >{\centering\arraybackslash}X
    >{\centering\arraybackslash}X
    >{\centering\arraybackslash}X
    >{\centering\arraybackslash}X
    >{\centering\arraybackslash}X
    >{\centering\arraybackslash}X
    >{\centering\arraybackslash}X
    >{\centering\arraybackslash}X   
    >{\centering\arraybackslash}X 
    }

    \toprule
    \multirow{2}{*}{\textbf{$\alpha$}}&
    \multirow{2}{*}{\textbf{$\beta$}}&&
    \multicolumn{3}{c}{\textbf{Airplane}} &&
    \multicolumn{3}{c}{\textbf{Car}}&&
    \multicolumn{3}{c}{\textbf{Chair}}\\
    &&& P& C & CD && P & C & CD && P & C & CD\\
    \midrule
    {1}&{0}&
    & {0.72} & {23.42} & {24.14} & 
    & {1.25} & {27.99} & {29.23} &  
    & {1.22} & {37.61} & {38.84} \\
    
    {0.9}&{0.1}&
    & {0.67} & {2.57} & {3.24} & 
    & {1.26} & {4.94} & {6.2} &  
    & {1.05} & {4.21} & {5.26} \\

    {0.5}&{0.5}&
    & {0.74} & {1.71} & {2.45} & 
    & {1.25} & {3.17} & {4.42} &  
    & {1.21} & {2.6} & {3.81} \\

    \textbf{0.1}&\textbf{0.9}&
    & {1.20} & {0.80} & \textbf{2.01} & 
    & {1.65} & {1.28} & {2.93} &  
    & {2.25} & {1.46} & \textbf{3.71} \\ 
    
    {0}&{1}&
    & {3.33} & {0.75} & {4.08} & 
    & {1.79} & {1.09} & \textbf{2.89} &  
    & {4.49} & {1.23} & {5.72}      \\
    \bottomrule
    
    \end{tabularx}

    \caption{
        \textbf{Ablation Studies for $\alpha$ and $\beta$.} 
        We also calculate average values among the categories.
        P, C, and CD refers to precision, coverage, and Chamfer distance, respectively.
        All the values are multiplied by $100$.
        %
        }
    \label{tab:Ablation_weight}
    \vspace{2mm}
\end{table*}

We quantitatively evaluate our method under different settings of $\alpha$ and $\beta$ parameters in equation~\textcolor{red}{5} of the main manuscript.
The results in Table~\ref{tab:Ablation_weight} demonstrate that our method can achieve the best performance with $\alpha=0.1$ and $\beta=0.9$ in average.
%
%
